\documentclass{article}

\usepackage{longtable}
\usepackage{graphicx} %
\usepackage{booktabs}
\usepackage{hyperref}
\usepackage{xcolor}
\usepackage{multirow}
\usepackage{pdflscape}
\usepackage[a4paper, total={6in, 10in}]{geometry}
\usepackage{amssymb}%
\usepackage{pifont}%
\usepackage{caption}
\usepackage{subcaption}

\usepackage{xeCJK} %
\usepackage{multicol}
\usepackage{enumitem}
\usepackage{amsmath} 
\usepackage{cleveref}
\usepackage{tcolorbox}

\tcbuselibrary{breakable, skins}

\usepackage{array}
\usepackage{makecell}
\newcolumntype{C}[1]{>{\centering\arraybackslash}p{#1}}

\newcommand{\para}[1]{\smallskip\noindent\textbf{#1}}

\newcommand{\neutral}{\textit{neutral}}

\title{Bridging the Language Gap in Scholarly Data \textsc{II}: Enhancing Gender Inference for Chinese Names}
\author{Lisette, Ana, Mingrong, Liu}
\date{}

\author{
Lisette Espín-Noboa$^{1,2,\dagger}$,
Mingrong She$^{3}$,
Ana Maria Jaramillo$^{1,2}$,
Liuhuaying Yang$^{2}$
\\
\footnotesize $^1$Graz University of Technology, Graz, Austria\\
\footnotesize $^2$Complexity Science Hub, Vienna, Austria\\
\footnotesize $^3$Maastricht University, Maastricht, The Netherlands\\[0.5em]
\footnotesize $^\dagger$Corresponding author: \href{mailto:espin@csh.ac.at}{espin@csh.ac.at}
}

\date{}

\begin{document}

\maketitle

\begin{abstract}
 Perceived gender inference from names is important for inclusive scientometrics research, yet existing methods perform poorly on non-Western scholars. 
 We address this gap for Chinese names, which appear in two forms, Hanzi (e.g., 张伟, 李娜) and Pinyin, its Romanized form (e.g., Zhang Wei, Li Na). 
 In international bibliographic records, names are often available only in Pinyin. 
 However, a single pinyin name can correspond to multiple distinct hanzi characters with different gender associations (e.g., ``Wei'' may represent 伟 (male), 薇 (female), or 威 (male)), so gender inference from Pinyin alone is intrinsically ambiguous. 
 To address this, we first develop two hanzi models, a regression model that produces a continuous gender score from 0 (male) to 1 (female), and a classification model that outputs one of three labels (\textit{male}, \textit{neutral}, \textit{female}).
 We train them on a stratified sample of hanzi names from Chinese-language physics publications, 
 rated by $11$~native Chinese speakers. 
 On this sample, the models achieve an RMSE of $0.09$ and an R$^2$ of $0.85$ (regression), and macro-AUC of $0.97$ and a Cohen's $\kappa$ of $0.81$ (classification), demonstrating high performance that is robust to class imbalance.
 For pinyin names, we propose a pipeline that aggregates hanzi predictions via transliteration and frequency weighting, together with an ambiguity score that flags names whose gender cannot be resolved from Pinyin alone. 
 Benchmarked against three off-the-shelf tools (\texttt{Namsor}, \texttt{GPT}, and \texttt{Claude}), our models achieve the highest agreement with human annotations for both gender scores and gender labels, in both scripts.
 Applied to an independent sample of Chinese-named authors in American Physical Society journals, our models assign a gender to every name, whereas existing Pinyin-based methods leave between 20\% and 100\% of names unclassified. 
 These findings demonstrate the importance of hanzi-level information for reliable gender inference of Chinese names.
 Code and data are publicly available for replication.
\end{abstract}

\begin{figure}[t]
    \centering
    \includegraphics[width=\textwidth]{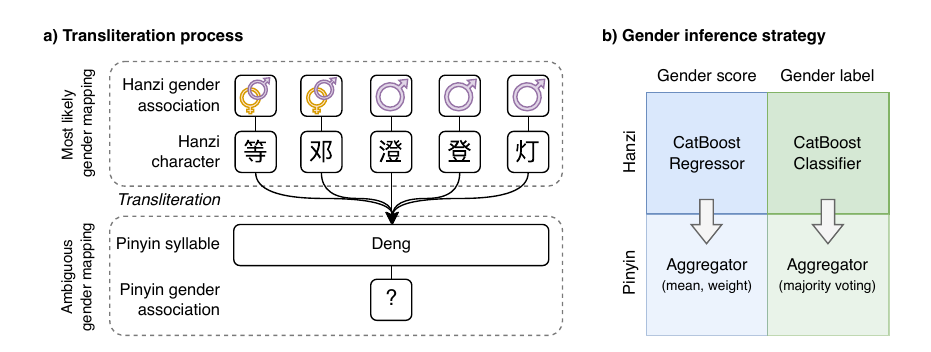}    
    \caption{\textbf{Pinyin transliteration and gender inference pipeline.}
     (a) Transliteration process: 
     A single pinyin syllable (e.g., ``Deng'') maps to multiple hanzi characters, each with its own predominant gender association. 
     (b) Gender inference strategy: 
     First, we train CatBoost models to predict gender from hanzi characters: a regressor to produce continuous gender scores and a classifier to produce categorical labels.
     For pinyin names, we leverage the transliteration process and aggregate the predicted gender from all hanzi characters that map to that syllable.
    }
    \label{fig:overview}
\end{figure}

    \section{Introduction}
    
    \label{sec:intro}
     Scientometrics and science-of-science research often rely on inferring socio-demographics such as gender to understand central dynamics: who collaborates with whom~\cite{iglivc2017whom,bachmann2026cumulative}, who cites whom~\cite{king2017men,wu2024gender}, and how recognition and resources flow across communities~\cite{huang2020historical,squazzoni2021peer,li2026women}. 
    The goal is not only to describe patterns but to detect biases---for example, in hiring, citation, or funding---and to design interventions that make research more equitable~\cite{lariviere2013bibliometrics,li2025undervaluing,barnes2025edge,van2025supporting}. 
    However, for large-scale studies, individual attributes are often unavailable, making other observable but noisy proxies, such as names or bibliometrics, necessary to infer gender~\cite{Karimi2016,lerman2022gendered}, ethnicity~\cite{Chekili2024, Kozlowski2022avoidingbias}, or geographic origin~\cite{Huang2014,Ortu2017}.
    
    Most existing gender-inference tools are developed and evaluated using data from English-speaking or well-resourced populations, where census data and labeled name lists are readily available~\cite{zhou2019examining,chen2021gender,sogaard2022should,vanhelene2024inferring}. 
    As a result, many non-Western naming conventions remain poorly supported~\cite{Karimi2016,santamaria2018comparison,van2023open,vanhelene2024inferring}. 
    Chinese names are a prominent example. 
    Prior work reports substantially lower accuracy for Chinese names, which are often excluded from streamlined analyses~\cite{Karimi2016,van2023open}. 
    This exclusion introduces systematic bias in studies of collaboration, citation, and diversity. 
    Addressing this gap is therefore essential for representative scientometric and science-of-science research.
    
    To address this limitation, we must consider how Chinese given names (hereafter ``names'') are represented. 
    Names may appear in \textbf{Hanzi}, the logographic Chinese script, or in a Romanized form (i.e., Latin alphabet to represent pronunciation). 
    For logographic names, we use Simplified Chinese to ensure consistency with mainland Chinese naming conventions. 
    For Romanized names, we adopt Hanyu \textbf{Pinyin}, the official Romanization system for Standard Mandarin~\cite{cheng2026toward}, the most widely spoken variety of Chinese~\cite{britannica_mandarin_2026}.
    The mapping between the two representations, known as transliteration, is asymmetric. 
    Each \textit{hanzi character} typically corresponds to a single pinyin syllable (with rare exceptions~\cite{Li1988}), but a single pinyin syllable may map to many Hanzi---a well-documented characteristic of Mandarin~\cite{odinye2020mandarin,chinesenames_syllable_rules,chen2000new}.
    In our corpus, 8{,}094~distinct single hanzi characters collapse into only 408~pinyin syllables, and the three most frequent pinyin syllables (yi, ji, yu) each map to over 120~hanzi characters (146, 130, 123, respectively).
    This transliteration ambiguity compounds with a second source of uncertainty.
    Each hanzi character carries both meaning and sound, and parents typically select one or two characters, rarely three, guided by semantics, phonetics, and family tradition rather than strict gendered conventions~\cite{bin1987personal,kaluzynska2016linguistic}.
    As a result, the association between any hanzi character and gender is statistical rather than deterministic.
    Consider the pinyin name ``Deng,'' which may represent 澄, 登, or 灯---typically found in male names therefore perceived as masculine---but also 等 or 邓, can be found in both genders.
    When only ``Deng'' is observed, the underlying hanzi character is unknown, and even if it were known, its gender signal can only be probabilistic.
    Gender inference from Pinyin alone is therefore inherently ambiguous, see~\Cref{fig:overview}a.

    In international bibliographic databases and English-language journals, full names are generally recorded in Pinyin rather than Hanzi. 
    For example, the full name ``张澄'' appears as ``Zhang Deng,'' which readers unfamiliar with Hanzi can pronounce. 
    Because Chinese scholars publish extensively in international venues~\cite{zheng2016chinese}, pinyin representations are pervasive in global scholarly datasets. 
    As a result, gender inference frequently relies on pinyin syllables alone, an imprecise proxy for the underlying hanzi characters, whose association with gender is itself probabilistic. Gender inference from pinyin names must therefore account for the multiple hanzi characters the syllable represents.

    We leverage this many-to-one mapping between Hanzi and Pinyin as the basis of our framework, see~\Cref{fig:overview}b. We first train two independent models that infer gender directly from hanzi character names: (i) a regression model that predicts a continuous gender score between $0$ (mostly male) and $1$ (mostly female), and (ii) a three-class classification model that infers a categorical label (\textit{male}, \textit{neutral}, \textit{female}). The two formulations serve complementary purposes---categorical labels suit applications requiring discrete decisions, while continuous scores preserve the inherent ambiguity of name-gender associations and offer finer-grained distinctions. 
    Each hanzi name is represented by features derived from three existing gender-inference tools, which provide complementary gender signals at the character level.
    Inference for pinyin names does not require additional model training. Instead, we transliterate a pinyin name into all plausible hanzi candidates and reuse the predictions from hanzi-based models. For regression outputs, we aggregate hanzi scores using four strategies that vary in their weighting schemes (uniform vs. frequency-weighted) and reference populations (our corpus vs. national statistics). For classification outputs, we aggregate predictions via majority voting, assigning \textit{male}~or \textit{female}~when one label dominates, and \textit{neutral}~when the labels are tied. %
    
    To evaluate our framework in a real-world scholarly setting, we construct a large-scale corpus from physics publications indexed in the China National Knowledge Infrastructure (CNKI)~\cite{cnki}, spanning 1953--2024. The corpus comprises $106{,}411$~authors, yielding $81{,}177$~unique full names and 39{,}712~unique given names in Hanzi~\cite{she2026bridging} (see~\Cref{app:sec:cnki}). Since single-character given names are under-represented in this corpus (4\%), we augment the dataset with an additional 6{,}432~single-character names~\cite{hanzidb} to ensure sufficient coverage of these rarer name forms.
    A subset of names is annotated for perceived gender by Chinese annotators on a five-point scale ranging from 1 (strongly male) to 5 (strongly female). 
    The averaged scores serve as the reference for training and evaluating our hanzi models. %
    Evaluating pinyin-based inference poses an additional challenge, since annotations exist only for hanzi names. We therefore transliterate the annotated hanzi names into Pinyin and assess how well each aggregation strategy recovers their annotated gender scores. Beyond a single estimate, we argue that pinyin-based inference should also report its ambiguity. We quantify this as the spread of gender scores across all hanzi variants of a pinyin name, flagging names whose inferred gender should be interpreted with caution. Finally, we benchmark our models against off-the-shelf methods, including two commercial large language models (LLMs), \texttt{GPT 5.4 Nano}~and \texttt{Claude Opus 4.8}~(see~\Cref{sec:models:comparison}), on both hanzi and pinyin names.

    Our contributions are threefold. First, we introduce a framework for gender inference grounded in the linguistic structure of Chinese names. By integrating signals from three existing hanzi-based gender-inference tools, we train both regression and classification models that capture complementary aspects of the name–gender relationship.
    Second, we extend our hanzi-level models to Pinyin by aggregating their predictions across all hanzi characters that map to a given pinyin syllable. In doing so, we demonstrate that transliteration ambiguity is not merely a technical limitation but a structural source of bias: commonly used pinyin-based approaches frequently default to predominantly male or unknown predictions, distorting the inferred gender composition of scholarly communities~\cite{Karimi2016,van2023open}, whereas our aggregation strategy preserves the gender signals encoded in the underlying hanzi characters.
    Third, we release code~\cite{chinese_names_github} and data~\cite{chinese_names_data} to support reproducible and inclusive scientometrics and science-of-science research, and provide an interactive tool to facilitate exploration of the linguistic complexity of hanzi- and pinyin-based gender inference~\cite{notmygender}. 
    By addressing the systematic mishandling of Chinese names in existing gender-inference pipelines, our framework enables a more accurate and globally representative assessment of gender in academia.

\section{Related work}
\label{sec:related_work}

We conducted a systematic search on the Clarivate Web of Science database for name-to-gender inference methods targeting Chinese names (see~\Cref{app:sec:litrev} for the full protocol). Of the 14 retrieved papers, 10 are directly relevant and fall into three groups. A first group uses existing gender inference tools to study gender composition in medical or academic authorship contexts, treating name-to-gender assignment as a pipeline step rather than a research problem in its own~\cite{WOS:000879846700001,WOS:000948666700002,WOS:001077484000001,WOS:000468045500004}. One work proposes improvements to \texttt{genderize.io},\footnote{\url{https://genderize.io}} but is developed on a predominantly Western dataset (roughly 90\% French or English names from a single Swiss hospital) and relies on heuristics such as accent removal and compound-name pruning that are inapplicable to non-Western names~\cite{WOS:000721567300011}. A second group examines the perceived gender of Chinese names through survey experiments. In one such study, native English and German speakers reliably distinguished male from female Min Chinese names based on phonological cues such as syllable ending, with high cross-linguistic agreement~\cite{WOS:000671253500004}. More directly relevant to our Mandarin setting, 39 volunteers annotated a pool of Mandarin given names as male or female to construct a balanced sample, which was then used in social perception experiments with Chinese college students. These experiments found that feminine names were consistently associated with warmth and masculine names with competence, confirming that name-gender associations carry real social meaning~\cite{WOS:000635500200005}. However, the annotation step reports neither annotator demographics, nor per-name annotation counts, nor inter-rater agreement---limitations that our own crowd-sourced annotation protocol is designed to address. 
A third group benchmarks multiple tools---\texttt{gender-guesser},\footnote{\url{https://pypi.org/project/gender-guesser/}} \texttt{Gender API},\footnote{\url{https://gender-api.com}} \texttt{Namsor},\footnote{\url{https://namsor.app}} 
\texttt{genderize.io}, 
\texttt{NameAPI},\footnote{\url{https://www.nameapi.org/}}
and 
\texttt{Wiki-Gendersort}~\cite{berube2020wikigendersort}\footnote{\url{https://github.com/nicolasberube/Wiki-Gendersort}}---either on general name samples~\cite{santamaria2018comparison,WOS:000705202000009} or specifically on Chinese names~\cite{WOS:000792126300005}.
The latter found that all evaluated tools inaccurately predicted gender from pinyin given names, cautioning against their use for this population.
None of these previous works develops or evaluates a method that operates directly on hanzi characters, which led us to examine three additional works not captured by our systematic search.

These three additional works share the core idea of combining Hanzi and Pinyin information for gender inference. 
The earliest approach concatenates hanzi and pinyin embeddings and fine-tunes BERT~\cite{devlin2019bert} for gender classification~\cite{jia2019gender}.
A subsequent method combines BERT embeddings of both the hanzi name and its pinyin with a graph-based representation of the hanzi characters~\cite{pan2023underrepresented}. This graph, the Chinese Heterogeneous Graph Attention Network (CHGAT), links each character to its semantic and phonetic components and its pronunciation.
The most recent work also uses both scripts, but only during training, so that inference requires Pinyin alone: a shared pinyin encoder jointly learns pinyin-to-character mapping (as an auxiliary task) and binary gender classification, while a hanzi-based gender classifier supervises it through knowledge distillation; this method achieves lower error rates than commercial tools (\texttt{Namsor}, \texttt{Gender API}, \texttt{genderize.io}) and higher accuracy than \texttt{GPT 3.5}~\cite{du2024misgendered}.
Trained on male-skewed data, this model misclassifies female names more often, as the authors note.
Like these works, we leverage a character-level hanzi-gender mapping. 
Our work differs from these in three ways: 
\textbf{i)} rather than training a dedicated pinyin model, we reuse hanzi-based predictions by mapping each pinyin name to its Hanzi candidates; 
\textbf{ii)} instead of a binary label, we offer either a continuous gender score or a three-class label, each accompanied by an ambiguity score for pinyin names; and 
\textbf{iii)} we validate against perceived-gender ratings from Chinese-speaking annotators and benchmark on both hanzi and pinyin names against off-the-shelf tools, including recent LLMs (\texttt{GPT}~and \texttt{Claude}).

\section{Methodology}
\label{sec:methodology}

We first describe our data, comprising hanzi given names paired with their pinyin transliterations. We detail name-gender associations obtained from external tools across the full corpus and from human annotators for a smaller sample serving as ground-truth. We then present the regression models for both hanzi- and pinyin-based inference, each with its own evaluation metrics and baselines. 
For the classification models, we refer the reader to~\Cref{app:sec:class-models}.

\subsection{Data}
\label{sec:data}

\subsubsection{Hanzi given names (CNKI corpus)}
\label{sec:chinese-given-names}

We use $65{,}241$~publications from the China National Knowledge Infrastructure (CNKI)~\cite{cnki},
focusing on journals published by the Chinese Physical Society (CPS), covering the years $1953$--$2024$~\cite{she2026bridging} (more details in~\Cref{app:sec:cnki:data}).
For each article, we extracted metadata (title, authors, affiliations, publication year, journal, abstract, keywords). 
From these, we extracted a set of unique full names and split each into surname and given name, following standard Chinese naming conventions, with validation against an existing name database~\cite{bao-2023}. 
We discarded $20$~full names consisting of a single character, for which no given name can be separated from the surname.
We used only given names, since surnames are inherited and uninformative for gender, and restricted to those of one or two characters, which account for over 83\% of given names in the population~\cite{bao2021novel}.
This yielded 39{,}712~unique hanzi given names, of which 4\%~are single-character names and the remainder two-character names. 

To improve coverage of the single-character names, we augmented them with HanziDB~\cite{hanzidb}, which lists $8{,}105$~commonly used hanzi characters. We excluded $9$ characters whose pronunciations are not provided in both HanziDB and Hanzi Quanxi.\footnote{\url{https://qxk.bnu.edu.cn/}} After removing characters already present in the CNKI corpus, this added 
6{,}432~new single-character hanzi names. 
Together, these steps yielded a final CNKI hanzi corpus of 46{,}144~unique names (82.45\% two-character and 17.55\% single-character), which we used to train and evaluate the hanzi models and to build hanzi statistics for the pinyin aggregation strategies.

\subsubsection{Transliteration (Hanzi--Pinyin mapping)}
\label{sec:transliteration}

Pinyin is the standard romanization of Mandarin Chinese, mapping each hanzi character to a Latin-script syllable plus a diacritical tone mark (e.g., \textit{zhōng}, \textit{zhòng}). 
The mapping between hanzi characters and pinyin syllables is many-to-many.
In particular, a single pinyin syllable can correspond to dozens of distinct hanzi characters.
The reverse ambiguity also exists but is rarer: roughly 10\% of hanzi characters are polyphonic, i.e., they carry more than one pronunciation and thus more than one pinyin representation~\cite{Li1988}.
Because standard pinyin libraries often ignore multiple readings and name-usage conventions, we built a dedicated Hanzi--Pinyin table in two steps.
First, we transliterated the 8{,}094~single characters from HanziDB (\Cref{sec:chinese-given-names}) using pronunciations from Hanzi Quanxi,\footnote{\url{https://qxk.bnu.edu.cn/}} prioritizing the reading most commonly used in given names.
Second, we used this character-level table to transliterate all 46{,}144~hanzi given names, yielding 18{,}739~unique pinyin names (of which 91~come from the single-character names added from HanziDB; see~\Cref{sec:chinese-given-names}).
We work with tone-stripped Pinyin throughout, since Chinese authors routinely drop tone marks when romanizing their names internationally (e.g., ``中城'' $\rightarrow$ \textit{zhōngchéng} $\rightarrow$ ``zhongcheng''). 
The resulting table thus supports both Hanzi-to-Pinyin lookup (for CNKI name statistics) and Pinyin-to-Hanzi lookup (for aggregating hanzi gender predictions).

\begin{figure}[!tbp]
    \centering
    \includegraphics[width=1.\linewidth]{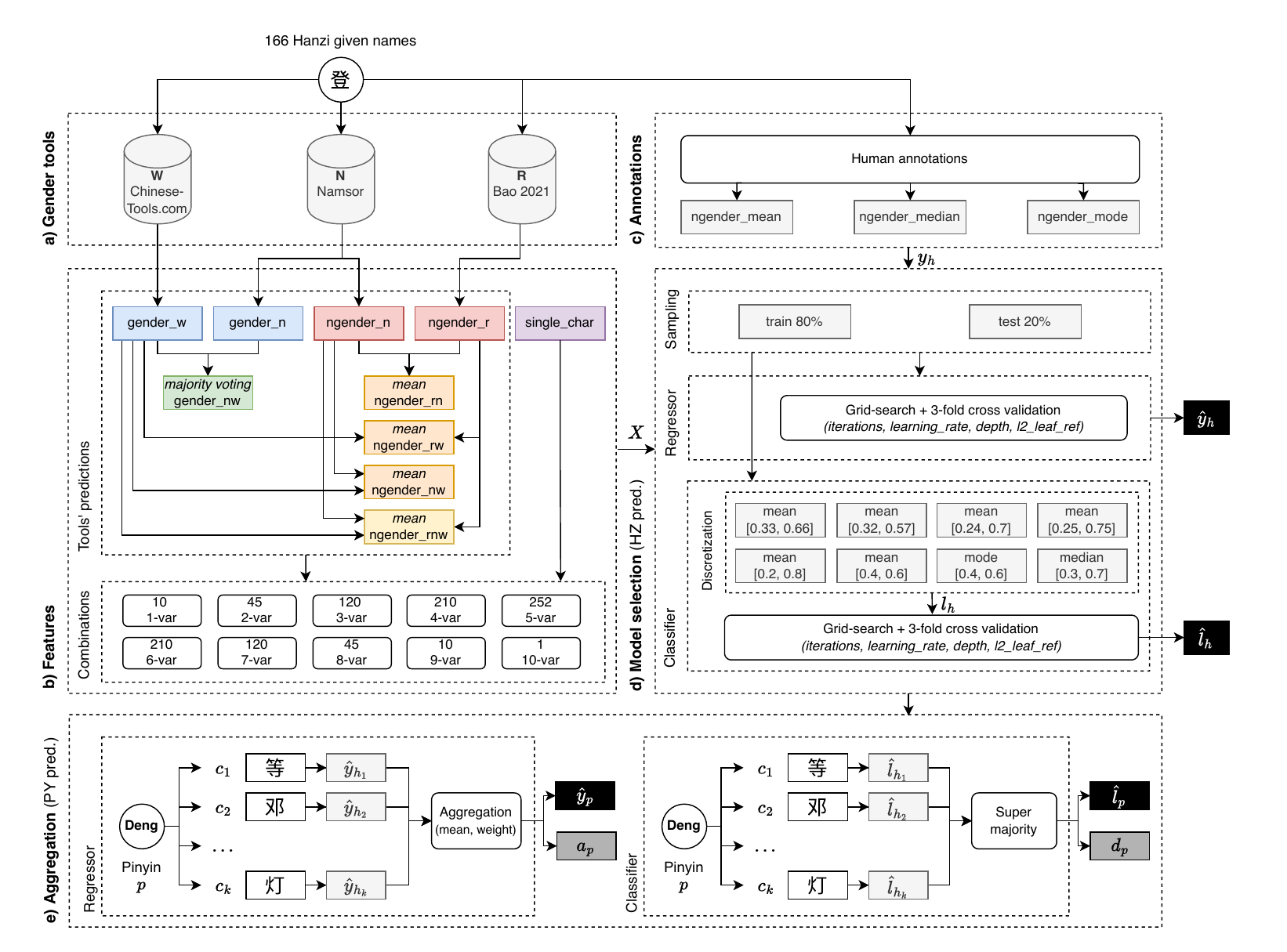}
    \caption{
    \textbf{Proposed gender inference pipeline.}
The ground-truth dataset consists of $166$~hanzi given names.
(a) Three external gender tools produce predictions for each hanzi name as 
a label (\texttt{gender\_*}) or score (\texttt{ngender\_*}).
(b) Ten features are derived from these predictions and individual variables, then combined through majority voting or pairwise means. All $1{,}023$~non-empty feature combinations are used individually to train models.
(c) Eleven human annotators label each hanzi name on a 5-point Likert scale (1 = male to 5 = female). Aggregated labels are generated using \texttt{mean}, \texttt{median}, and \texttt{mode}, yielding three alternative ground-truth values $y_h$.
(d) For each $y_h$, CatBoost regressors and classifiers are trained on the features $X$ from (b). Regressors predict the numeric annotation score $\hat{y}_h$. 
Scores $y_h$ are discretized into categorical labels $l_h \in \{\textit{male}, \textit{neutral}, \textit{female}\}$, which classifiers are trained to predict ($\hat{l}_h$).
The best models are selected based on performance on a held-out set.
(e) For pinyin-based inference, each Pinyin $p$ is transliterated back to all possible hanzi character mappings $c_i$, and their corresponding inferred hanzi gender scores $\hat{y}_{h_i}$ from (d) are aggregated into $\hat{y}_p$ using simple or weighted averages, with an ambiguity score $a_p$ capturing their spread. For labels, super majority determines the most common gender $\hat{l}_p$ among the hanzi representations $\hat{l}_{h_i}$, with a disagreement score $d_p$ capturing their divergence. Ties default to neutral.
}
    \label{fig:model_pipeline}
\end{figure}

\subsubsection{Hanzi features (X)}
\label{sec:features}
Our models require features that capture the gender association of each hanzi name in the corpus.
Rather than deriving features from publication metadata (e.g., co-authorship networks~\cite{lerman2022gendered}) or training deep-learning models that require extensive gender-labeled hanzi corpora to achieve sufficient coverage~\cite{jia2019gender,pan2023underrepresented}, we opt for a lightweight approach. 
We rely on external name-gender tools that operate solely on the name string and return gender associations derived from large historical census data (see~\Cref{fig:model_pipeline}a).
This keeps the features independent of any bibliometric context and makes the model applicable to names encountered outside a publication setting.

Among such tools, general-purpose name-gender APIs such as \texttt{Namsor}~\cite{namsor} have been shown to outperform other off-the-shelf methods in comparative evaluations~\cite{vanhelene2024inferring}, but they remain less reliable for Asian names, especially for Chinese names represented in Pinyin~\cite{WOS:000705202000009,WOS:000792126300005,du2024misgendered}.
However, our preliminary analyses against the gold-standard annotations show that \texttt{Namsor}~aligns well with perceived gender for Chinese names when hanzi characters are available (Pearson $\rho=0.73$, $RMSE=0.19$; see~\Cref{app:sec:tools_performance}). 
Specifically, we query \texttt{Namsor}~(\textbf{\texttt{N}}) through its \texttt{Genderize Name Geo API}~\cite{namsor}, which returns, among other metrics, \texttt{genderScale}, a numeric score in $[-1,1]$ (male/female), and \texttt{likelyGender}, its binary label representation, based on an onomastic model.
To improve on this baseline, we supplement \texttt{Namsor}~with two additional tools, yielding three complementary sources of gender signals for hanzi names. %
(i)~\texttt{Chinese-Tools gender guesser} (\textbf{\texttt{W}})~\cite{chinesetools_gender_guesser} returns a binary label (male/female); 
(ii)~\texttt{Chinese Name Database} 1930--2008 (\textbf{\texttt{R}})~\cite{bao-2023} provides a continuous gender score (\texttt{NG}), ranging within $[-1,1]$ (female/male), based on difference in proportions of a character used by male versus female, and a hanzi name uniqueness score (\texttt{NU}), a 1–6 log-scaled measure based on character usage in names, which we later use as population weights in the pinyin-based aggregation strategies (see~\Cref{sec:pinyin-model});

Note that \texttt{R}\ and \texttt{N}\ use opposite polarities; we standardize both to $[0,1]$ (with $0$ = male, $1$ = female).
From these three tools, we derive ten features per hanzi name (see~\Cref{fig:model_pipeline}b), including numeric scores, categorical labels, majority-vote labels, a boolean single-character indicator, and averaged score combinations.
As detailed in~\Cref{app:sec:tools_performance}, among the three tools, \texttt{R}~shows the highest agreement with the mean gold-standard ratings, and the combined score \texttt{mean\_ngender\_rn}~(the average of \texttt{R}\ and \texttt{N}) is the best-performing numeric feature across all ten. %
These features are used in combination (from single-feature to all ten) for model training, and for inference on the full corpus. %

\subsubsection{Gold-standard annotations}
\label{sec:gold-standards}

Chinese names do not map deterministically to a single gender. Even when a hanzi character is predominantly associated with one gender, it may carry secondary associations with another, and different native speakers may perceive the same name differently. 
We therefore rely on multiple human annotators to capture the perceived gender of hanzi names (not the gender identity of people). 
We annotate Hanzi rather than Pinyin because Pinyin omits the characters that carry most gender associations.
Rather than imposing binary labels, we adopt a five-point Likert scale to reflect the continuous spectrum of gender associations a single name can carry, including an explicit neutral midpoint for names with weak or mixed gender associations.

We sampled $166$~names from the CNKI corpus to serve as the sole ground truth for training and evaluating our hanzi models. 
These names ($50$~single-character and $116$~two-character names) were stratified by gender agreement among the three gender-inference tools (\Cref{sec:features}).
Eleven native Chinese speakers\footnote{27\% female, 73\% male; 55\% aged 20--30, 45\% aged 30--40.} rated each name as follows: \texttt{1} = definitely male, \texttt{2} = more likely male, \texttt{3} = neutral, \texttt{4} = more likely female, \texttt{5} = definitely female. 
We rescaled the annotators' ratings to $[0,1]$ and aggregated them in three ways: mean, median, and mode (\Cref{fig:model_pipeline}c). 
Each served separately as a target gender score for the regression models (\Cref{sec:regression-models}).
Further, we discretized the gender scores into three-class labels under eight different threshold configurations (\Cref{fig:model_pipeline}d) and evaluated each configuration separately as target labels for the classification models (\Cref{app:sec:class-models}).
We assessed shared structure among annotators with a metric cultural consensus model based on the inter-annotator correlation matrix~\cite{romney1986culture}. We quantified the inter-rater reliability of the averaged ratings using the intraclass correlation coefficient (ICC(2,k)~\cite{shrout1979intraclass,koo2016guideline}), appropriate for a fully crossed two-way random-effects design with average measures.

\subsection{Regression models}
\label{sec:regression-models}

We train a regression model that maps each hanzi given name to a continuous gender score, capturing the graded nature of name--gender associations rather than forcing a binary decision.

\subsubsection{Hanzi-based regression (gender score \texorpdfstring{$\hat{y}_h$}{ŷₕ})}
\label{sec:hanzi-model}

We train regression models to predict the continuous perceived gender score $y_h$, defined as the mean, median, or mode of annotators' ratings rescaled to $[0,1]$ (see~\Cref{sec:gold-standards}). %
We use an 80\%-20\% stratified train-test split of the $166$~gold-standard names.
We fit \texttt{CatBoost} regressors with grid search over hyperparameters (iterations, learning rate, depth, L2 regularization), 3-fold cross-validation on the training set, and early stopping.
This is repeated for each feature combination and each target aggregation (mean, median, mode) under three different random seeds.
We let CatBoost handle missingness natively via minimum-value substitution.\footnote{\url{https://catboost.ai/docs/en/concepts/algorithm-missing-values-processing}}

\para{Evaluation.}
We select the model with the lowest root mean squared error (\texttt{RMSE}) on the held-out test set and report both RMSE and $R^2$.
This selection implicitly determines the best feature set $X$ and target aggregation (mean, median or mode).

\subsubsection{Pinyin-based regression (gender score \texorpdfstring{$\hat{y}_p$}{ŷₚ})}
\label{sec:pinyin-model}

For pinyin gender-score inference, we do not train a new model.
Instead, for each Pinyin $p$, we retrieve all its matching hanzi names $h$ (\Cref{sec:transliteration}) along with their predicted gender scores $\hat{y}_h$ (\Cref{sec:hanzi-model}).
These are then aggregated into a single pinyin-level score (\Cref{fig:model_pipeline}e, left).

We consider four aggregation strategies that vary along two axes.
The first is the \emph{candidate set} of hanzi names, either $\mathcal{H}_{\text{cnki}}(p)$ from the original CNKI corpus or the augmented $\mathcal{H}_{\text{cnki+}}(p) \supseteq \mathcal{H}_{\text{cnki}}(p)$, which includes single-character names from HanziDB (\Cref{sec:chinese-given-names}).
The second is the \emph{weighting scheme}, either uniform or proportional to name frequency.
All four strategies are instances of a single weighted mean:
\begin{equation}
\hat{y}_p^{(s)} \;=\; \frac{\sum_{h \in \mathcal{H}_s(p)} w_s(h)\,\hat{y}_h}{\sum_{h \in \mathcal{H}_s(p)} w_s(h)},
\end{equation}
differing only in $\mathcal{H}_s(p)$ and $w_s(h)$:

  \begin{enumerate}[label=(\alph*)]

  \item \textbf{Mean CNKI:} Uniform weights over CNKI-observed names.
  \begin{equation*}
    \mathcal{H}_s(p) = \mathcal{H}_{\text{cnki}}(p),
    \quad w_s(h) \equiv 1.
  \end{equation*}

  \item \textbf{Mean CNKI$+$:} Uniform weights over the augmented set
    including single-character names from HanziDB.
  \begin{equation*}
    \mathcal{H}_s(p) = \mathcal{H}_{\text{cnki+}}(p),
    \quad w_s(h) \equiv 1.
  \end{equation*}

    \item \textbf{Weighted CNKI:} Weighted by the number of distinct CNKI authors with name $h$, obtained after author name disambiguation~\cite{she2026bridging} (see~\Cref{app:sec:cnki:weights}).\begin{equation*}
    \mathcal{H}_s(p) = \mathcal{H}_{\text{cnki}}(p),
    \quad w_s(h) = w_{\text{cnki}}(h).
  \end{equation*}

  \item \textbf{Weighted Pop.:} Weighted by a smooth proxy for
    national name frequency, derived from the
    Bao~\cite{bao-2023} uniqueness index
    $\mathrm{NU}_{\text{pop}}(h) \in [1, 6]$.
  \begin{equation*}
    \mathcal{H}_s(p) = \mathcal{H}_{\text{cnki+}}(p),
    \quad w_s(h) = w_{\text{pop}}(h)
      = \left(10^{-\mathrm{NU}_{\text{pop}}(h)}
        - 10^{-6}\right) \times 10^{6}.
  \end{equation*}
  Lower \texttt{NU} values denote more common characters, so $w_{\text{pop}}$ approximates relative frequency (e.g., \texttt{NU}=2 corresponds to roughly $1$ in $100$ individuals, while \texttt{NU}=6 yields a floor weight near zero). 
  Because $w_{\text{pop}}$ is defined nationally, it can weight every hanzi name mapping to $p$, including those absent from CNKI (\Cref{sec:chinese-given-names}). 
  
\end{enumerate}

\para{Evaluation.}
When the underlying Hanzi is known, gender can be inferred from pinyin with high accuracy. The ground-truth set, where both Hanzi and their transliterations are available, therefore provides an ideal benchmark for evaluating pinyin-level aggregation strategies. We transliterate each hanzi name to Pinyin and compare its mean annotator score $y_h$ with the aggregated pinyin-based prediction $\hat{y}_p$, effectively measuring how much information is lost through romanization and how well each strategy recovers it. 
We report $R^2$, MAE, MSE, and RMSE, and select the strategy with the lowest RMSE.

\para{Ambiguity.}
Performance of the pinyin inference is inherently limited by transliteration, as multiple hanzi names with divergent genders may share the same pinyin form. 
As a result, an aggregated score alone might not fully capture the heterogeneity of the underlying predictions. 
To quantify this heterogeneity, we define the \textit{ambiguity} of a pinyin-level score as the difference between the 90th and 10th percentiles ($q_{90} - q_{10}$) of the hanzi-based predictions associated with a given pinyin name. 
A high ambiguity indicates conflicting gender associations across hanzi variants.

\subsubsection{Comparison with off-the-shelf tools.}
\label{sec:models:comparison}

To contextualize our models, we compare their predictions on the held-out gold standard against three off-the-shelf tools: \texttt{Namsor}, \texttt{Claude Opus 4.8}~\cite{anthropic2026opus48}, and \texttt{GPT 5.4 Nano}~\cite{openai2026gpt54nano}. 
Each tool is queried separately with hanzi and pinyin input.
For \texttt{Namsor}, we use the same standardized \texttt{genderScale} as in \Cref{sec:features}.
Since \texttt{Namsor}'s hanzi output also serves as one of our model features (\Cref{sec:features}), we re-queried the API for this comparison.
\texttt{Claude}\ and \texttt{GPT}\ are prompted as described in \Cref{app:sec:prompt}.
We evaluate gender scores with $R^2$, MAE, MSE, and RMSE. %

\section{Results}
\label{sec:result_experiments}

\begin{figure}[t]
    \centering
    \begin{subfigure}[t]{0.48\textwidth}
        \centering
        \includegraphics[width=\textwidth]{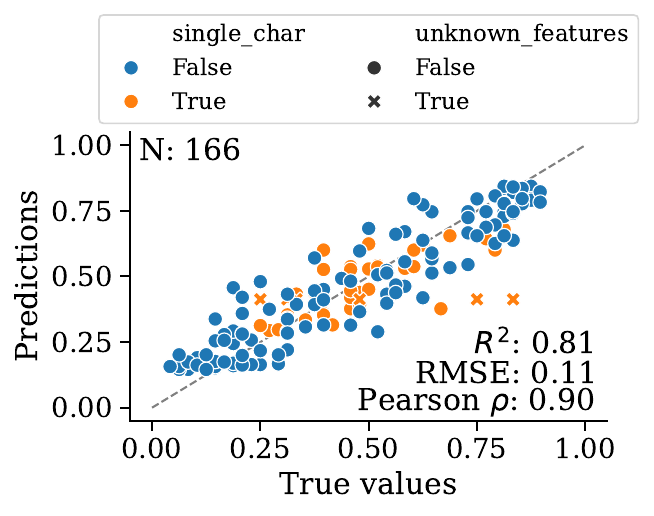}
        \caption{Ground-truth}
        \label{fig:hz-reg-all}
    \end{subfigure}
    \hfill
    \begin{subfigure}[t]{0.48\textwidth}
        \centering
        \includegraphics[width=\textwidth]{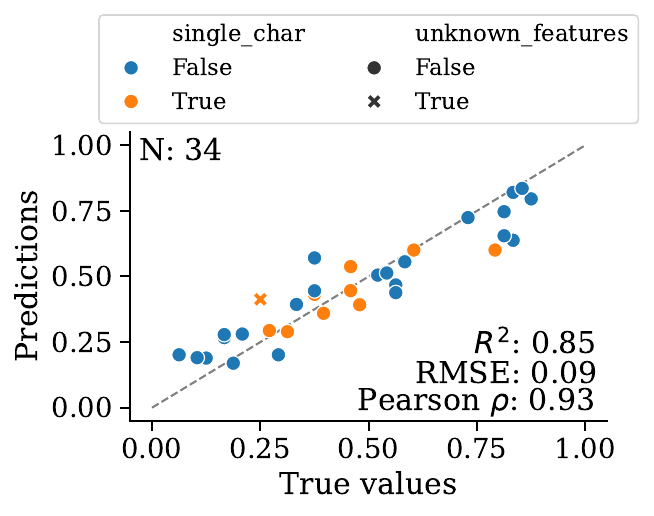}
        \caption{Test set}
        \label{fig:hz-reg-val}
    \end{subfigure}
    
    \begin{subfigure}[t]{0.48\textwidth}
        \centering
        \includegraphics[width=\textwidth]{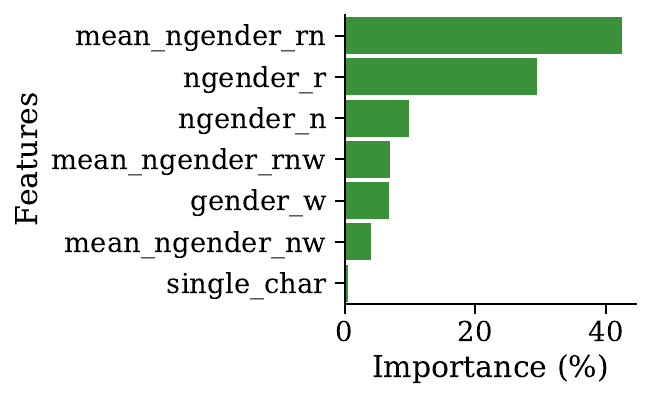}
        \caption{CatBoost feature importance}
        \label{fig:hz-reg-fi}
    \end{subfigure}
    \hfill
    \begin{subfigure}[t]{0.48\textwidth}
        \centering
        \includegraphics[width=\textwidth]{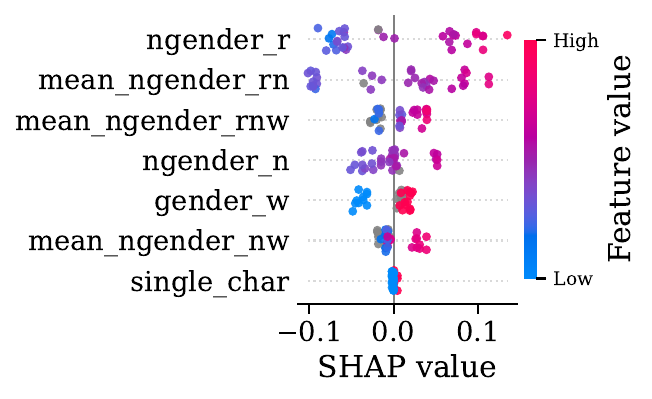}
        \caption{Impact on model output}
        \label{fig:hz-reg-shap}
    \end{subfigure}
    \caption{\textbf{Performance of hanzi-based gender score inference ($\hat{y}_h$).}
    Predicted versus true scores on (a) the full ground-truth set ($N=166$, train and test combined) and (b) the held-out test set ($N=34$).
    Points represent hanzi given names. Color indicates whether the name is single-character, and shape indicates whether features from all tools were available.
    Fit is summarized by $R^2$, root mean squared error (RMSE), and Pearson correlation ($\rho$).
    (c) CatBoost feature importance and (d) SHAP values both rank the aggregated score \texttt{mean\_ngender\_rn}\ and \texttt{ngender\_r}\ as the dominant predictors. The binary indicator \texttt{single\_char}\ and the categorical indicator \texttt{gender\_w}\ contribute minimally.
    }
    \label{fig:hz_reg}
\end{figure}

We report results for both regression models, hanzi- and pinyin-based gender-score inference. 
Regression preserves the continuous nature of annotator perception, avoids arbitrary cutoffs, and yields a single interpretable score aligned with our gold-standard aggregation (\Cref{sec:gold-standards}). 
We present classification results in \Cref{app:sec:class-models}, evaluating multiple thresholding schemes as performance is sensitive to how gender scores are discretized into labels (\textit{male}, \textit{neutral}, \textit{female}). %

\para{Annotator agreement.}
Annotators labeling the hanzi names sample (\Cref{sec:gold-standards}) exhibited strong shared perception of gender.
The cultural consensus analysis indicated a clear uni-dimensional structure ($\lambda_1/\lambda_2 = 12.76$), with the first factor explaining $67.2\%$ of inter-annotator variance.
Bootstrap resampling over names confirmed robustness (median $\lambda_1/\lambda_2 = 11.27$; $95\%$ CI $[8.41, 14.74]$).
Consistent with this, single-annotator agreement was moderate (ICC(2,1) = 0.62, 95\% CI [0.57, 0.68]), as expected for a subjective judgment, while the averaged ratings were highly reliable (ICC(2,k) = 0.95, 95\% CI [0.94, 0.96]). Nearly identical absolute- and consistency-based ICCs further indicate negligible systematic differences between annotators. Thus, most variance reflects differences between names rather than disagreement among annotators.

\begin{figure}[t]
    \centering
    \begin{subfigure}[t]{1.0\textwidth}
        \centering
        \includegraphics[width=1.0\textwidth]{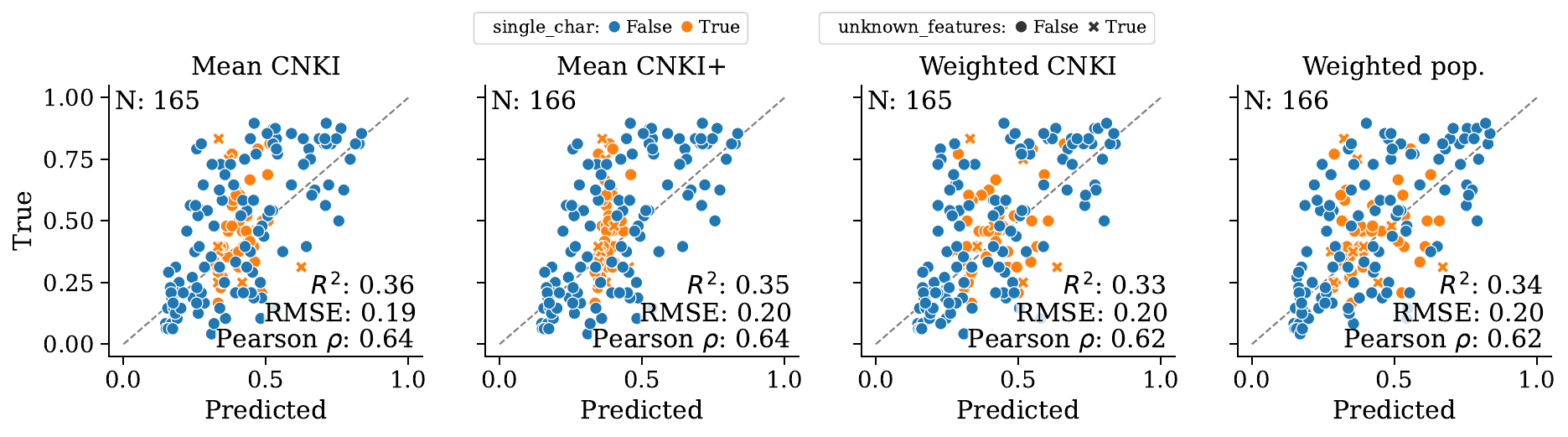}
        \caption{Ground-truth} %
        \label{fig:py-reg:all}
    \end{subfigure}
    \hfill
    \begin{subfigure}[t]{1.0\textwidth}
        \centering
        \includegraphics[width=1.0\textwidth]{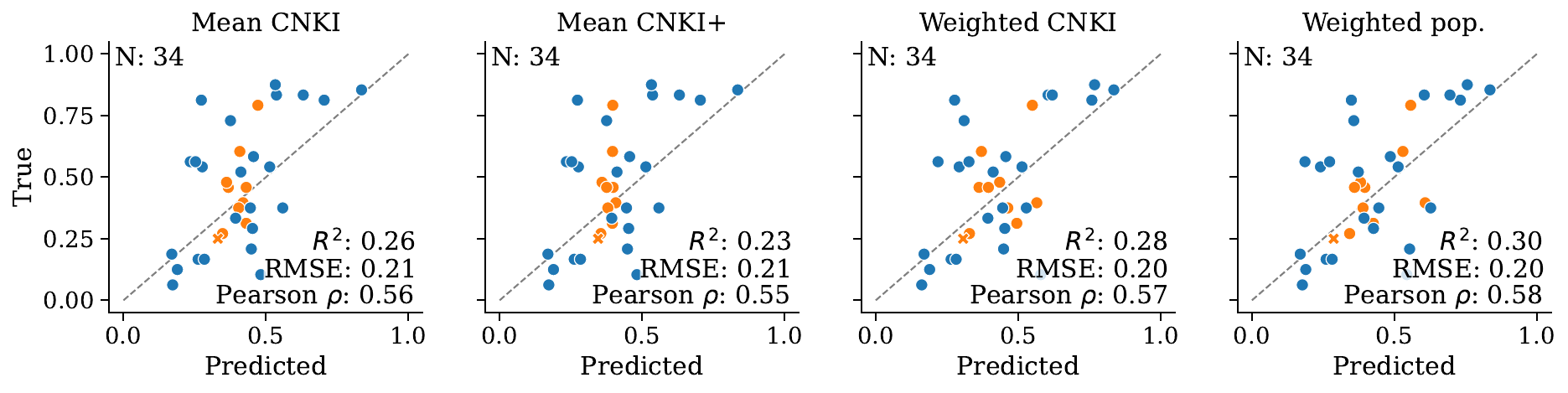}
        \caption{Test set}
        \label{fig:py-reg:test}
    \end{subfigure}
    \caption{\textbf{Performance of pinyin-based gender score inference ($\hat{y}_p$).}
    True scores (y-axis) versus predicted values (x-axis) on (a) the full ground-truth set (train and test combined) and (b) the held-out test set.
    Points represent pinyin names transliterated from the hanzi ground-truth dataset. 
    True values are the mean score from the human annotations on the hanzi version of each name.
    Color indicates whether the name is single-character, and shape indicates whether features from all tools were available.
    Each column corresponds to one aggregation strategy (\Cref{sec:pinyin-model}). 
    Some panels show fewer name pairs because of our single-character augmentation (\Cref{sec:chinese-given-names}).
    Fit is summarized by $R^2$, root mean squared error (RMSE), and Pearson correlation ($\rho$).
    }
    \label{fig:py-reg}
\end{figure}

\subsection{Hanzi-based inference (gender score \texorpdfstring{$\hat{y}_h$}{ŷₕ})}
\label{sec:result-ch-regression}

\Cref{fig:hz_reg} shows the performance of our best hanzi regression model, which predicts the \textbf{mean} of the gold-standard annotations and uses 7 of the 10 derived features.

\para{Predicted vs.\ true scores.}
\Cref{fig:hz-reg-all,fig:hz-reg-val} show predicted versus true gender scores. 
On the full ground-truth set (train and test combined, $N=166$), the model achieves $R^2 = 0.81$, RMSE $= 0.11$, and Pearson $\rho = 0.90$ (\Cref{fig:hz-reg-all}).
On the held-out test set ($N=34$), the fit is slightly stronger, with $R^2 = 0.85$, RMSE $= 0.09$, and $\rho = 0.93$ (\Cref{fig:hz-reg-val}).
The small drop on the full set reflects a minority of names with one or more missing features.
Most points lie close to the diagonal, indicating that the model recovers the gold-standard mean well on both datasets. %

\para{Feature importance and interpretability.}
\Cref{fig:hz-reg-fi,fig:hz-reg-shap} show which features drive the model.
CatBoost feature importance (\Cref{fig:hz-reg-fi}) ranks the aggregated numeric score from \texttt{R}\ and \texttt{N}\ (\texttt{mean\_ngender\_rn}) highest ($\sim$40\%), followed by \texttt{ngender\_r}\ alone ($\sim$30\%) and \texttt{ngender\_n}\ ($\sim$10\%).
The next contributions come from \texttt{mean\_ngender\_rnw}\ and \texttt{gender\_w}\ (each $\sim$7\%) and \texttt{mean\_ngender\_nw}\ ($\sim$3\%). The feature \texttt{single\_char}\ contributes less than 1\%.
The main predictive power thus comes from the numeric gender signals of \texttt{R}\ and \texttt{N}, while the binary and categorical indicators \texttt{gender\_w}\ and \texttt{single\_char}\ contribute minimally.
The SHAP values (\Cref{fig:hz-reg-shap}) corroborate this ranking. High values of \texttt{ngender\_r}\ and \texttt{mean\_ngender\_rn}\ push the prediction toward female (positive SHAP, high feature value), and low values push it toward male (negative SHAP, low feature value). The SHAP values for \texttt{single\_char}\ cluster near zero.

\subsection{Pinyin-based inference (gender score \texorpdfstring{$\hat{y}_p$}{ŷₚ})}
\label{sec:result-py-regression}
We assess pinyin-based inference by comparing each aggregation
strategy (\Cref{sec:pinyin-model}) against the hanzi gold-standard annotations, on both the full ground-truth set
(top, \Cref{fig:py-reg:all})
and the held-out test set (bottom, \Cref{fig:py-reg:test}).
Across all four strategies, single-character names (orange points) pose the greatest challenge, with aggregated scores clustering near $0.4$ while the corresponding hanzi-based true values span $0.2$ to $0.7$.
Further investigation reveals that these single-character Pinyin forms map to many distinct hanzi names (19--101) with
heterogeneous gender scores, compressing the aggregated score
toward the mean with high variance (\Cref{app:tbl:pinyin_stats_test_set}, top).
The weighted strategies attempt to mitigate this compression by
prioritizing more frequent or representative hanzi names, resulting in more spread predictions for single-character names under Weighted CNKI and Weighted Pop (right most columns, \Cref{fig:py-reg}).
Two-character pinyin names, on the other hand, are more constrained, mapping to fewer Hanzi with more homogeneous gender scores (\Cref{app:tbl:pinyin_stats_test_set}, bottom), explaining their closer alignment with the Hanzi-based predictions.
Overall, the best-performing strategies are Mean CNKI on the full ground-truth set ($R^2=0.36$, RMSE$=0.19$; \Cref{fig:py-reg:all}, top-left) and Weighted Pop.\ on the held-out test set ($R^2=0.30$, RMSE$=0.20$; \Cref{fig:py-reg:test}, bottom-right).

\begin{figure}
    \centering
    \includegraphics[width=1.0\linewidth]{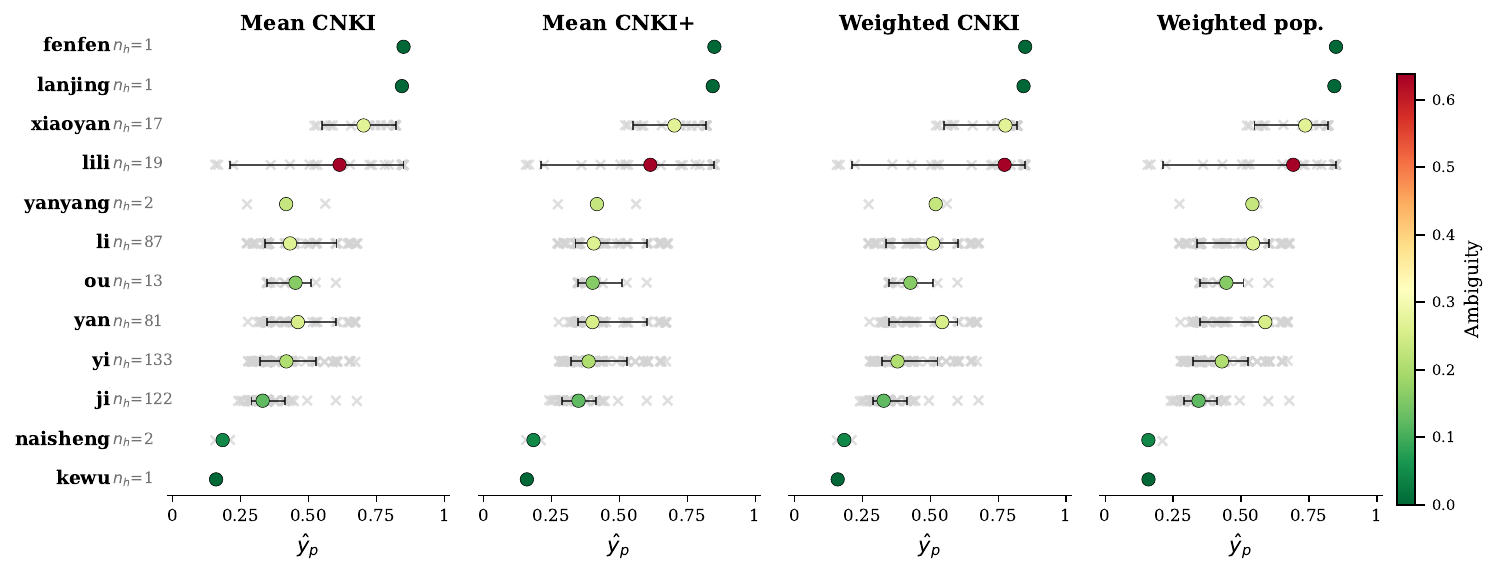}
    \caption{\textbf{Ambiguity of pinyin-level gender score aggregation.}
    A sample of pinyin names with varying gender associations, shown across the four aggregation strategies. 
    Circles denote the aggregated pinyin-level score ($\hat{y}_p$), with error bars spanning the 10th to 90th percentile ($q_{10}$, $q_{90}$) of the hanzi-level predictions. 
    Color encodes the ambiguity of the aggregation, defined as the width of this interval ($q_{90} - q_{10}$). 
    Grey crosses mark the individual gender scores of all hanzi variants associated with each pinyin name.
    The number of hanzi variants ($n_h$) is reported next to each name. 
    Names with high ambiguity (warm colors, wide error bars) indicate conflicting gender associations across hanzi variants, warranting caution when interpreting the aggregated score.}
    \label{fig:sample_py_ambiguity}
\end{figure}

\Cref{fig:sample_py_ambiguity} illustrates the effect of ambiguity on pinyin-level gender scores across the full corpus. 
When the aggregated score falls near the extremes of the scale (e.g., \textit{fenfen}, \textit{kewu}, $n_h = 1$), predictions are stable across all four strategies, as these names typically map to a single hanzi with an unambiguous gender association. 
However, the number of hanzi variants alone does not determine ambiguity. 
\textit{Ji} ($n_h = 122$) exhibits low ambiguity despite its many variants, as the underlying hanzi names cluster around similar gender scores. 
In contrast, \textit{lili} ($n_h = 19$) shows considerably higher ambiguity with far fewer variants, confirming that ambiguity captures not only the spread but also the concentration of gender associations. 
The case of \textit{lili} also highlights how aggregation strategies diverge in practice. 
The weighted approaches assign a more female score than the unweighted means, reflecting that the hanzi names most frequently associated with \textit{lili} in CNKI tend to be female, a pattern not captured by treating all variants equally. 
A similar effect appears for \textit{xiaoyan} ($n_h = 17$), where the high ambiguity and wide interdecile range result in markedly different predictions depending on whether frequency information is incorporated. 
These patterns motivate the inclusion of ambiguity as a complementary measure in our released dataset.

\begin{figure}[t]
    \centering
    \begin{subfigure}[t]{1.0\textwidth}
        \centering
        \includegraphics[width=1.0\textwidth]{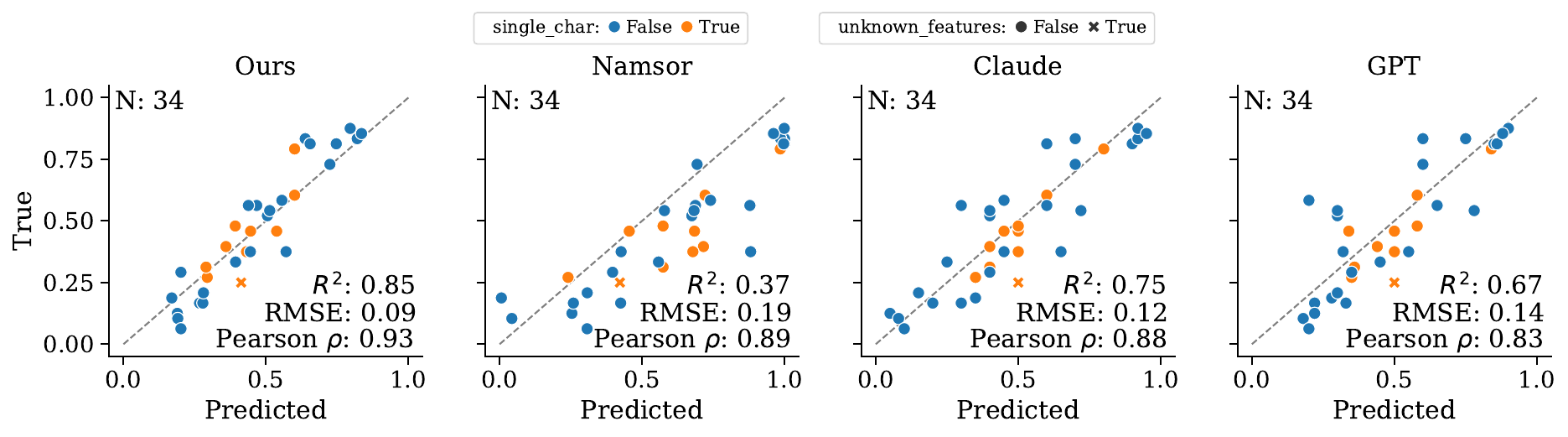}
        \caption{Hanzi-based inference}
        \label{fig:comparison_test_set:hanzi}
    \end{subfigure}
    \hfill
    \begin{subfigure}[t]{1.0\textwidth}
        \centering
        \includegraphics[width=1.0\textwidth]{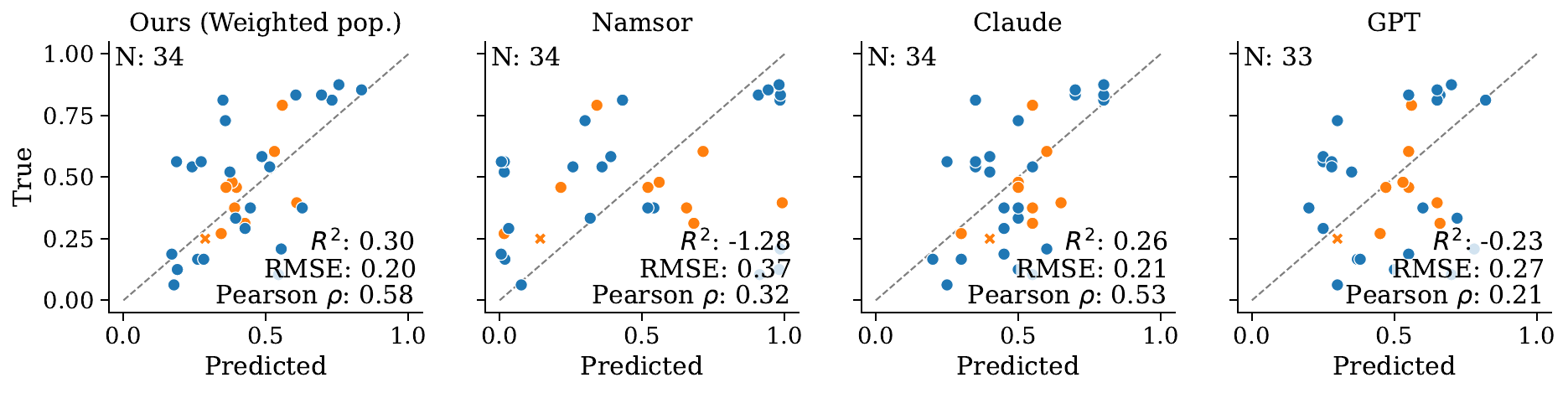}
        \caption{Pinyin-based inference}
        \label{fig:comparison_test_set:pinyin}
    \end{subfigure}
    \caption{\textbf{Comparison of gender score predictions across methods.}
    Mean annotator scores (y-axis) versus predicted gender scores (x-axis) for our model and three off-the-shelf methods (\texttt{Namsor}, \texttt{GPT 5.4 Nano}, \texttt{Claude Opus 4.8}), using (a)~hanzi-based and (b)~pinyin-based inference.
    Each point represents one of 34 held-out names; the dashed line indicates perfect agreement.
    Our model achieves the highest $R^2$ and lowest RMSE in both settings.
    }
    \label{fig:comparison_test_set}
\end{figure}

\subsection{Comparison with off-the-shelf methods}
\label{sec:result-comparison}

For hanzi-based inference (\Cref{fig:comparison_test_set:hanzi}), our model aligns closely with the ground-truth scores on the held-out test set ($R^2 = 0.85$, RMSE $= 0.09$), outperforming all three off-the-shelf methods, followed by \texttt{Claude}~($R^2 = 0.75$, RMSE $= 0.12$).
\texttt{Namsor}, despite being one of our input features (\Cref{sec:features}), yields the lowest $R^2$ ($0.37$) and highest RMSE ($0.19$) among hanzi-based methods. Two factors may explain this. First, \texttt{Namsor}'s outputs changed between the query used to build our features and the one used for this comparison (\Cref{sec:limitations}). Second, \texttt{Namsor}~alone is a weaker signal than its combination with complementary features, as reflected in its ranking third in feature importance (\Cref{fig:hz_reg}).
Standalone performance of each tool is reported in \Cref{app:sec:tools_performance}.

The transition to Pinyin reveals a sharp decline across all methods (\Cref{fig:comparison_test_set:pinyin}), consistent with the transliteration ambiguity discussed in~\Cref{sec:intro}.
Collapsing multiple Hanzi into a single pinyin syllable discards gender-discriminative information, most visibly for single-character names (orange), whose gender is well captured in Hanzi yet largely unrecoverable in Pinyin. 
\texttt{Namsor}\ and \texttt{GPT}\ yield negative $R^2$ on Pinyin ($-1.28$ and $-0.23$, respectively), with \texttt{Namsor}\ showing the steepest decline from $R^2=0.37$ on Hanzi. \texttt{Claude}\ remains positive but drops sharply ($R^2=0.26$, from $0.75$ on Hanzi). Our best pinyin aggregation strategy, Weighted pop., outperforms all three methods ($R^2=0.30$, RMSE $=0.20$), though it explains substantially less variance than hanzi-based inference ($R^2=0.85$). This gap reflects the inherent information loss of transliteration.

\section{Limitations and future work}
\label{sec:limitations}
We discuss the limitations of our work and the directions for future work they suggest.
First, our gold-standard annotations are drawn from author names in Chinese physics publications (CNKI). 
Although we stratified the sample by gender agreement among the tools' predictions, yielding a balanced range of gender scores, our models may perform differently on names from other populations, such as other disciplines or diaspora communities. 
Name-gender associations also change over time, following regional trends, historical events, and naming practices~\cite{gao2011shall}, so our models may perform differently on earlier or later cohorts. 
Our population weights partly account for such differences in the pinyin aggregation, but they do not affect the hanzi-level predictions themselves. 
Enlarging the annotated sample and extending it to names from other populations and birth cohorts would yield more precise performance estimates and help assess how well our models generalize. 
Where birth year or career start is available, it could refine predictions, but supplying it as an input alone is not enough.
Classifiers fine-tuned on names paired with birth years improve slightly, whereas adding birth year to LLM prompts does not help and often reduces accuracy, especially for names whose gender associations changed over time~\cite{you2024beyond}.

Second, each pinyin name in our evaluation is compared against the annotated score of a single hanzi variant, whereas our pinyin models aggregate over all plausible variants. 
For ambiguous names, part of the error therefore reflects ambiguity that cannot be resolved from pinyin alone, rather than model failure. 
Our ambiguity score makes this uncertainty explicit, flagging names whose pinyin-based predictions should be interpreted with caution. 
Since it is computed from the hanzi variants observed in our corpus, future work can recalibrate it for other populations by re-estimating the set of variants from their own data.

Third, our pinyin aggregation depends on corpus-specific choices. 
The candidate hanzi variants for each pinyin name are those observed in our corpus, and we evaluated four aggregation strategies, two unweighted and two weighted by either CNKI character frequencies or population-level name statistics. 
Weighting by population statistics performed best on our test set, although differences between strategies were small. 
For other populations, different weights or a more complete set of hanzi variants, for example from national name records, may be more appropriate. 
We did not evaluate such adaptations.
However, the aggregation step is modular, so candidate sets and weights can be re-estimated from any target corpus without retraining the hanzi models.

Finally, commercial tools may change their outputs over time without notice, which limits reproducibility.
When we re-queried \texttt{Namsor}~for the comparison in \Cref{sec:models:comparison}, its outputs differed from those used as model features (RMSE $= 0.28$, Pearson $\rho = 0.64$).
Of the 156 names labeled in both queries, 63 (40\%) changed label, 55 of them from male to female.
Our models were trained and evaluated consistently with the first query, but re-running our pipeline with current API outputs may yield different features.
The same concern applies to LLMs, which are increasingly used for name-to-gender inference~\cite{goyanes2024automatic,sebo2024performance,dominguez2024comparative}.
These studies report high accuracy on mostly Western names, but also sensitivity to prompt wording and occasional changes in output for the same input~\cite{goyanes2024automatic,dominguez2024comparative}, and LLMs remain largely untested on Asian names~\cite{sebo2024performance,dominguez2024comparative}.
In our analyses, the \texttt{Claude}~versions used during prototyping were no longer available when we scaled up, so all inferences had to be repeated with a newer version.
Requests to classify the full pinyin set through \texttt{Claude Code} were repeatedly blocked by a content filtering policy, which Anthropic attributes to safeguards against reproducing existing material~\cite{antropic_statement}. 
Splitting the names into smaller batches did not help, and a workaround through the \texttt{Claude.ai} chat interface soon stopped functioning. 
Similar refusals have been reported for ChatGPT~\cite{goyanes2024automatic}. 
Researchers relying on commercial tools or LLMs for name-to-gender inference should therefore validate their outputs against annotated data before use.
Future work should also push providers to document model changes and the provenance of their outputs, beyond version releases.\section{Discussion}
\label{sec:discussion}

Our results show that hanzi-level information is central to inferring the perceived gender of Chinese names.
Here, we discuss what perceived gender means for name-based inference, how to choose between gender scores and labels, the consequences of names left unclassified, the role of LLMs, and how our approach extends beyond Chinese names.

\para{Perceived gender and ambiguity.}
Our models infer the \textit{perceived} gender of a name, not the gender identity of the person who holds it.
Like any name-based tool, they rely only on the name as text, and can therefore capture only the associations that people attach to names~\cite{lockhart2023name,WOS:000635500200005}.
For this reason, their predictions are suited to aggregate analyses rather than to assigning gender to individuals~\cite{Gonzalez-Salmon2024-hz}.
Since pinyin encodes only pronunciation and carries little gender information on its own, inferring the gender of a pinyin name requires relying on its possible hanzi forms (\Cref{fig:overview}a).
Existing approaches do so but assume binary gender and return a single estimate per pinyin name, whether by learning pinyin representations from hanzi~\cite{du2024misgendered}, linking pinyin names to a closed pool of known hanzi names~\cite{li2025gender}, or aggregating registry records of recorded sex~\cite{shi2025open}.
None of them quantifies the uncertainty that arises when these hanzi forms carry divergent gender associations.
Our \textit{ambiguity} and \textit{disagreement} scores fill this gap by quantifying how consistent the gender associations of a name's hanzi candidates are, for predicted scores and labels, respectively (\Cref{sec:pinyin-model,fig:sample_py_ambiguity}).
We argue that pinyin-based inference should report such an index alongside the most likely gender, since the same prediction can be reliable for one name and arbitrary for another.

\para{Choosing between scores and labels.}
Our regression and classification models serve different purposes.
The continuous gender score suits analyses of aggregate trends and avoids fixing class boundaries, whereas the three-class labels suit studies that require discrete groups.
Annotators provided graded ratings rather than class labels, since asking for labels would require fixing the number of classes in advance and could create artificial disagreement for near-neutral names.
Labels must therefore be derived from the ratings, and we compared eight ways of doing so, varying the aggregation of annotator scores and the thresholds that separate male, neutral, and female (\Cref{app:sec:class:hz:discretization}).
Since these choices affect the resulting labels, studies using discrete classes should report the thresholds they apply~\cite{Gonzalez-Salmon2024-hz}.
For both outputs, allowing for neutral perception matters, because collapsing names perceived as neutral into male or female can skew aggregate gender ratios~\cite{lockhart2023name,you2024beyond}.

\para{Unknowns and downstream consequences.}
Studies of gender disparities in science that use name-based inference as a pre-processing step commonly discard names for which no gender is returned~\cite{huang2020historical,jadidi2018gender,jaramillo2025systematic}.
As a result, many Asian names are disproportionately excluded from these analyses~\cite{lockhart2023name,Gonzalez-Salmon2024-hz}.
To quantify this loss, we used a second independent dataset of physics publications from the American Physical Society (APS)~\cite{APS2026Datasets}, restricted to the subset of Chinese-named authors who wrote their names in both scripts, typically in pinyin followed by hanzi in parentheses (\Cref{app:sec:implications}).
For these authors, the hanzi form reveals the exact characters behind each pinyin name, so pinyin-based predictions can be compared against predictions from the actual hanzi rather than from all plausible candidates.
Among authors with fully spelled-out given names, existing pinyin-based methods leave between 20\% and 100\% of names unclassified, whereas our models assign a score and a label to every name in both scripts (\Cref{fig:py_baselines:dist}).
Discarding these names therefore reduces coverage substantially and may bias estimates if the excluded researchers differ in seniority, subfield, or productivity.
Since Chinese researchers produce a large and growing share of global scientific output~\cite{Adams2023,Dhand2023}, such exclusions can distort findings on authorship gaps, citation disparities, and funding inequalities~\cite{ni2021gendered,lerman2022gendered,li2025gender}.
Downstream estimates can also be corrected for misclassification using error rates from a labeled sample~\cite{marshall1990validation,barron1977effects}, such as those we report for hanzi and pinyin names (\Cref{app:fig:class-val,app:fig:py_class_performance:cmtest}).

\para{The role of LLMs.}
Our models performed best in all four comparisons against human annotations, namely gender scores for hanzi and pinyin names (\Cref{fig:comparison_test_set}) and gender labels for hanzi (\Cref{app:fig:hz_class_comparison}) and pinyin names (\Cref{app:fig:py_class_comparison}), with \texttt{Claude}\ ranking second in each.
On hanzi names, \texttt{Claude}\ reached $R^2 = 0.75$ for gender scores (\Cref{fig:comparison_test_set}), and per-class precision and recall between 0.80 and 0.88 for labels, but it performed considerably worse on pinyin names (\Cref{app:fig:hz_pr,app:fig:py_pr}).
This suggests that LLMs can capture the gender associations people perceive in Chinese characters, although this does not hold for all models, as \texttt{GPT}\ performed worse in both scripts.
Consistently, \texttt{Claude}\ also agreed most with our predictions on APS (80\% agreement and $\kappa = 0.62$ for hanzi, 66\% and $\kappa = 0.41$ for pinyin, \Cref{fig:py_baselines:agreement,fig:py_baselines:kappa}).
However, LLM outputs are difficult to reproduce over time~\cite{goyanes2024automatic,dominguez2024comparative} (\Cref{sec:limitations}), which currently limits their use as a standalone tool for research.

\para{Beyond Chinese names.}
The problem we address is not specific to Chinese names.
Other naming systems in which romanization discards meaning-bearing characters, such as Japanese names written in kanji, face the same many-to-one ambiguity~\cite{du2024misgendered}.
Our pipeline is modular, combining character-level predictions, a mapping from romanized forms to candidate characters, and population weights, so it can be adapted to such systems given suitable character-level resources.

\section{Conclusion}

We presented a framework for inferring the perceived gender of Chinese names written in Hanzi or Pinyin.
Our hanzi models, trained on human annotations of names from Chinese physics publications, combine signals from existing name-gender tools into either a continuous gender score or a three-class label, and agree more closely with human perception than off-the-shelf tools, including commercial LLMs.
For pinyin names, which existing tools often misclassifies or leave unclassified, our pipeline reuses hanzi predictions through transliteration and reports an ambiguity score that flags names whose gender cannot be resolved from Pinyin alone.
Our results show that hanzi-level information is key to reliable gender inference for Chinese names, and that inference from pinyin should be accompanied by a measure of its uncertainty.
We release our code~\cite{chinese_names_github}, predictions, annotated data~\cite{chinese_names_data}, and an interactive tool~\cite{notmygender}, so that our models can be reproduced, applied to other corpora, and used as a benchmark for future tools.
We hope these resources help science-of-science research include Chinese scholars, who are often left out of gender analyses, and thereby provide a more accurate picture of gender in global science.

\section*{Acknowledgements}
We thank Jana Diesner and Jinseok Kim for their feedback on earlier versions of this manuscript.\section*{Funding}
LEN was supported by the Vienna Science and Technology Fund WWTF under project no. ICT20-07, and the Austrian Science Promotion Agency FFG project no. 873927 ESSENCSE. 
MS acknowledges the financial support from the China Scholarship Council under grant number 202108080241. 
AMJ acknowledges the support of the MAMMOTH project Horizon 2020 HORIZON-CL4-2021-HUMAN-01-24 (RIA). 
LY was supported by the Austrian Ministry for Innovation, Mobility and Infrastructure as part of CSH Postdoc Program for Complexity Science and Data Competence.
\section*{Author contributions}
The study was conceived by LEN and LY. 
MS and LY provided domain expertise in the Chinese language and Chinese naming conventions. 
Data curation was performed by MS (collection of CNKI data), LEN (features from W's tool), AMJ (features from N's tool) and LY (features from R's tool). 
The methodology was developed by LEN (gender inference), MS (name disambiguation) and LY (pinyin-based regression). 
Formal analysis and investigation were performed by LEN (gender inference) and MS (name disambiguation). 
LY developed the visualization tool.
The first draft of the manuscript was written by LEN, and all authors reviewed and edited the manuscript. 
The work was supervised by LEN and LY. 
All authors read and approved the final manuscript.
\bibliographystyle{abbrv}

\newpage

\appendix
\renewcommand{\thefigure}{A.\arabic{figure}}
\renewcommand{\thetable}{A.\arabic{table}}
\setcounter{figure}{0}
\setcounter{table}{0}

\section*{Supplementary information}

\section{CNKI corpus and author name disambiguation}
\label{app:sec:cnki}

The CNKI corpus introduced in~\Cref{sec:chinese-given-names} was assembled and disambiguated in a companion study~\cite{she2026bridging}.
Because one of our pinyin aggregation strategies weights hanzi names by their number of distinct CNKI authors (\Cref{sec:pinyin-model}), the reliability of that weight depends on the disambiguation step.
This section therefore summarizes the corpus, the disambiguation algorithm, and its validation, so that the weighting scheme can be assessed without consulting the companion study; the full treatment, including the threshold grid search, the annotation interface, and per-method confusion matrices, is reported there.

\subsection{Data source}
\label{app:sec:cnki:data}

We collected $65{,}243$~publications from the China National Knowledge Infrastructure (CNKI), covering the $20$~journals published by the Chinese Physical Society up to 2024-02-14~(\Cref{app:tbl:cps_journals}).
Physics was chosen for its long publication record, large author community, and comparability with international corpora such as APS.
For each article we extracted the title, author list, author affiliations, publication year, journal, abstract, and keywords.
$2$~records were discarded because their author field contains no parseable Chinese personal name---one lists a transliterated foreign name followed by a stray character, the other a laboratory group rather than an individual---leaving the $65{,}241$~publications used throughout the paper.
The corpus records author names exactly as they appear on each paper and carries no author identifiers, which is what makes the disambiguation step necessary.

\begin{table}[htbp]
\centering
\caption{\textbf{Journals published by the Chinese Physical Society included in the CNKI corpus.}
For each journal we report the language of its articles (\textsc{hz}~=~hanzi, \textsc{en}~=~English) and the number of papers indexed up to 2024-02-14.
Journals marked in both columns publish in both languages, and counts are pooled across former titles of the same journal.
The column totals $65{,}243$~records, of which $65{,}241$~carry a parseable Chinese personal name and enter the analysis.}
\label{app:tbl:cps_journals}
\small
\begin{tabular}{l c c r}
\toprule
Journal & \textsc{hz} & \textsc{en} & Papers \\
\midrule
Acta Physica Sinica                             & \checkmark &        & 19{,}730 \\
Chinese Physics B                               &        & \checkmark &  8{,}813 \\
College Physics                                 & \checkmark &        &  6{,}081 \\
Chinese Journal of Atomic and Molecular Physics  & \checkmark & \checkmark &  5{,}743 \\
Wuli (Physics)                                  & \checkmark &        &  5{,}379 \\
Chinese Journal of Luminescence                 & \checkmark &        &  2{,}719 \\
Chinese Journal of Chemical Physics             &        & \checkmark &  2{,}559 \\
Chinese Physics C                               &        & \checkmark &  2{,}515 \\
Journal of Chinese Mass Spectrometry Society    & \checkmark &        &  2{,}059 \\
Nuclear Physics Review                          & \checkmark &        &  1{,}741 \\
Chinese Physics Letters                         &        & \checkmark &  1{,}513 \\
Journal of Quantum Optics                       & \checkmark &        &  1{,}402 \\
Communications in Theoretical Physics           &        & \checkmark &  1{,}039 \\
Physics Teaching                                & \checkmark &        &  1{,}001 \\
Chinese Journal of High Pressure Physics        & \checkmark &        &    852 \\
Chinese Journal of Light Scattering             & \checkmark &        &    595 \\
Progress in Physics                             & \checkmark &        &    514 \\
Journal of Chinese Electron Microscopy Society  & \checkmark &        &    499 \\
Chinese Journal of Liquid Crystals and Displays & \checkmark &        &    482 \\
Chinese Journal of Magnetic Resonance           & \checkmark & \checkmark &      7 \\
\bottomrule
\end{tabular}
\end{table}

\subsection{Why disambiguation is required}
\label{app:sec:cnki:why}

The \textbf{Weighted CNKI} strategy (\Cref{sec:pinyin-model}) weights each hanzi name $h$ by how common it is among CNKI physicists.
Measuring that frequency by counting names (text) would conflate popularity of a name with productivity of the people who bear it: a single prolific author publishing forty papers would outweigh forty authors sharing a different name, and the resulting weights would describe publication behaviour rather than naming practice.
We therefore define the weight over \emph{people} rather than names, which requires resolving which names belong to the same individual.
Disambiguation reduces the $81{,}177$~distinct hanzi full names in the corpus to $106{,}411$~\emph{identities}, each a set of records judged to belong to one person; these are the $106{,}411$~authors reported in~\Cref{sec:data}.
Mapping identities back to names yields the weights defined in~\Cref{app:sec:cnki:weights}.

\subsection{Disambiguation algorithm}
\label{app:sec:cnki:method}

\begin{figure}[htbp]
    \centering
    \includegraphics[width=0.95\linewidth]{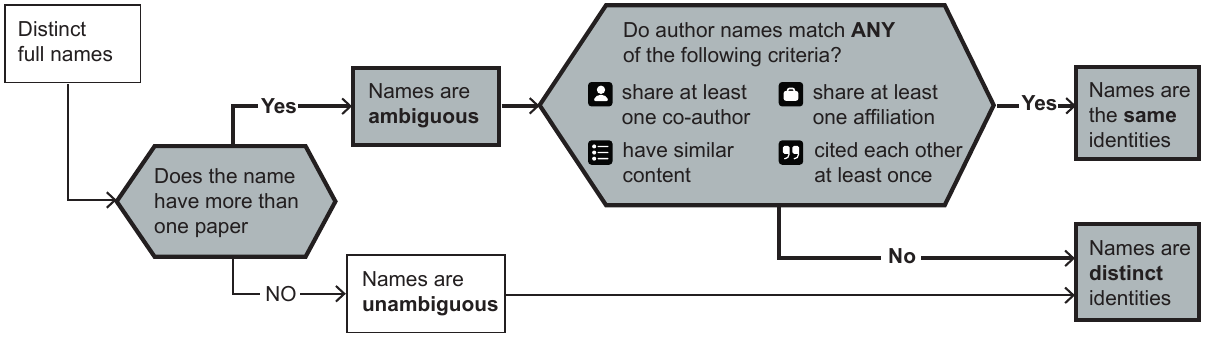}
    \caption{\textbf{Author name disambiguation pipeline.}
    A name associated with more than one paper is treated as ambiguous.
    Each pair of papers sharing that name is tested against co-authorship, affiliation, content, and citation criteria; a match on any criterion merges the pair into one identity, and pairs matching none are separated into distinct identities.
    The disambiguation stage is shaded in dark gray.
    Adapted from~\cite{she2026bridging}.}
    \label{app:fig:disambiguation}
\end{figure}

A name appearing on a single paper is trivially resolved to one identity.
Ambiguity arises only for names appearing on at least two papers, and only these enter the pipeline (\Cref{app:fig:disambiguation}).
For every pair of papers sharing such a name, the algorithm evaluates four criteria in sequence and assigns the pair to the same identity as soon as any one of them is satisfied; pairs surviving all four are assigned to distinct identities.

\begin{enumerate}[label=(\roman*),topsep=2pt,itemsep=1pt]
  \item \textbf{Shared co-authorship.} The two papers share at least one co-author.
  \item \textbf{Affiliation similarity.} The two affiliation strings reach a normalized Levenshtein similarity~\cite{yujian2007normalized} of at least $0.6$, or one string contains the other after stripping non-alphanumeric characters.
  \item \textbf{Content similarity.} The cosine similarity between Word2Vec~\cite{mikolov2013efficient} document vectors, built from title, abstract, and keywords, exceeds a script-specific threshold: $0.9207$~for hanzi and $0.9621$~for pinyin.
  \item \textbf{Citation link.} One paper cites the other, indicating a self-citation.
\end{enumerate}

The first, second, and fourth criteria capture social and institutional ties; the third adds thematic proximity for authors whose records show no such ties.
Separate content thresholds are required because the hanzi and pinyin corpora occupy different embedding spaces and their similarity scores are not on a common scale.
Both thresholds were selected by grid search to maximize agreement between the two scripts on which pairs count as similar, rather than to equalize the numeric cutoffs~\cite{she2026bridging}.

\subsection{Resolved identities}
\label{app:sec:cnki:results}

The algorithm was run twice on the same publications: once on the hanzi records, and once on a parallel version in which names were transliterated to pinyin and titles, abstracts, and keywords were machine-translated into English.
The second run mirrors how these authors appear in international bibliographic databases.

Of the $81{,}177$~distinct hanzi full names, $35{,}873$~($44\%$) appear on at least two papers and enter the pipeline; the pinyin corpus contains $68{,}601$~distinct pinyin full names, of which $33{,}441$~($49\%$) are ambiguous.
Shared co-authorship resolves by far the largest share of ambiguous names in both scripts, followed by affiliation similarity, while content and citation criteria each contribute a few hundred additional merges.
After all four criteria, the pipeline yields $106{,}411$~identities from the hanzi records and $99{,}222$~from the pinyin records.
The two counts move in opposite directions for the two reasons one would expect.
Transliteration collapses distinct hanzi names onto shared spellings, so the pinyin corpus starts from fewer distinct names, and each colliding spelling now covers more people: $14{,}467$~pinyin names resolve into more than one identity, against $13{,}580$~hanzi names, yielding $45{,}088$~identities where hanzi yields $38{,}814$.
That extra splitting, however, does not offset the collapse, and the pinyin run ends with fewer identities overall.

\subsection{Validation}
\label{app:sec:cnki:eval}

Because the corpus carries no author identifiers, the algorithm was evaluated against a manually annotated sample.
We randomly drew $80$~pairs of same-name hanzi records, balanced by construction between pairs the algorithm had merged and pairs it had separated, and asked two native Chinese speakers to judge independently whether each pair referred to the same person, given the two names and the two paper titles.
Inter-annotator agreement was Cohen's $\kappa$ = $0.75$~(Krippendorff's $\alpha$ = $0.76$); the ten pairs on which the annotators initially disagreed or could not decide were resolved by discussion.
Because the two runs describe the same records, the resulting labels transfer directly to the corresponding pinyin pairs, which lets us score both scripts against one ground truth rather than two independently annotated ones.

Against this ground truth the algorithm reaches an F$_1$ of $0.89$~on hanzi names (precision $1.00$, recall $0.80$) and $0.88$~on pinyin names (precision $0.95$, recall $0.82$).
Precision is thus near-perfect in both scripts: the algorithm almost never merges records belonging to different people, which is the error mode that would most distort the name weights.
Recall is the weaker of the two, meaning some records of the same person are left unmerged.
The same evaluation shows the algorithm outperforming two established baselines~\cite{sinatra2016quantifying,waqas2021multilayer}, chiefly through higher recall, with per-method confusion matrices reported in~\cite{she2026bridging}.

\subsection{Consistency across scripts}
\label{app:sec:cnki:jaccard}

Because both runs describe the same publications, the two sets of identities should agree.
To quantify this, each hanzi identity was transliterated to pinyin and compared, through the Jaccard similarity of their paper sets, against every pinyin identity that carries the same spelling and shares at least one of its papers; the resulting scores were averaged per hanzi identity.
Restricting the comparison to overlapping identities is what makes it a measure of agreement between the two runs: identities that share a pinyin spelling but no papers are different people brought together by romanization, not a disagreement about the same person.
Mean Jaccard similarity across the $106{,}411$~hanzi identities is $0.92$, and $87.3\%$~of them are assigned exactly the same papers under both scripts.

The remaining identities disagree, sometimes completely.
These are concentrated where transliteration is many-to-one: distinct hanzi names sharing a pinyin spelling create collisions that exist only in the romanized corpus, and the algorithm partitions the affected records differently.
This is the same structural loss of information that motivates our gender inference framework, observed here on an independent task: what pinyin discards is not recoverable from the romanization itself, whether the goal is identifying a person or inferring a perceived gender.

\subsection{From identities to name weights}
\label{app:sec:cnki:weights}

The weight $w_{\text{cnki}}(h)$ of the Weighted CNKI strategy (\Cref{sec:pinyin-model}) is the number of disambiguated identities whose given name is $h$, obtained by applying the surname-given name split of~\Cref{sec:chinese-given-names} to each identity and counting each person once regardless of how much they published.
Two properties of this definition matter for the aggregation strategies.
First, the weights come from the hanzi run, not the pinyin one: the pinyin identities of~\Cref{app:sec:cnki:results} are never used as weights, since the quantity we need is how many people bear a given \emph{character} sequence.
Second, $w_{\text{cnki}}(h) = 0$ for the 6{,}432~single-character names added from HanziDB (\Cref{sec:chinese-given-names}), which by construction do not occur in the corpus.
These names therefore cannot be weighted by CNKI counts, which is why the Weighted CNKI strategy aggregates over $\mathcal{H}_{\text{cnki}}(p)$ while the population-weighted strategy, whose weights are defined nationally, extends to the larger candidate set $\mathcal{H}_{\text{cnki+}}(p)$.

\section{Chinese name-to-gender literature search} %
\label{app:sec:litrev}

We selected a set of keywords to capture all works that tackle name-to-gender inference for Chinese names. As a query for gender inference, we used: \texttt{ALL=(gender) AND ALL=(infer) AND ALL=(name) AND ALL=(chinese) OR ALL=(name-to-gender)}, over the title, abstract, and full text of all indexed articles. All queries were performed on 2023-10-23.
This search returned 14 published articles, of which we retained 10 that proposed or evaluated methods for Chinese name gender inference. These 10 papers (\Cref{tbl:name_to_gender_literature_search}) are the basis for our related literature in~\Cref{sec:related_work}.

Only one of the 10 papers focused on the name-to-gender inference problem using computational methods for Chinese names~\cite{WOS:000792126300005}; however, it focuses on Pinyin rather than Hanzi. Of the other 9 papers, there are 6 papers using a computational approach for name-to-gender inference with proprietary databases~\cite{santamaria2018comparison,WOS:000468045500004,WOS:000705202000009,WOS:000721567300011,WOS:000948666700002,WOS:001077484000001} and one with their own data~\cite{WOS:000879846700001}, all of them mention the limitation of bad inference for Asian names in general but do not address it. And the remaining 2 papers that address name-to-gender in Chinese names focus on the reasons and meanings behind gender selections for certain names, or the gender stereotypes behind the use of certain genders~\cite{WOS:000671253500004,WOS:000635500200005}. None of the retrieved papers addressed the name-to-gender inference from Hanzi.

\begin{table}[h!]

\caption{
\textbf{Chinese name–to–gender inference literature identified via Web of Science.} 
A structured search in Clarivate Web of Science (WoS) yielded 14 records, of which 10 were retained after excluding studies that did not propose or evaluate methods for Chinese name gender inference. Columns indicate publication year and whether the study includes a computational approach (Comp. appr.), uses an external database (DB), provides code (Code), proposes a new method (New meth.), compares multiple methods (Meth. comp.), or explicitly focuses on Chinese names (Ch. names). %
}
\label{tbl:name_to_gender_literature_search}

\centering
\small
\setlength{\tabcolsep}{3pt}
\renewcommand{\arraystretch}{1.1}

\begin{tabular}
{C{0.25cm} 
p{7.0cm} 
C{0.8cm} 
C{1.0cm} 
C{0.85cm} C{0.85cm} C{0.85cm} C{0.85cm} C{0.9cm}}
\toprule
\# &
Title &
Year &
\makecell{Comp.\\appr.} &
DB &
Code &
\makecell{New\\meth.} &
\makecell{Meth.\\comp.} &
\makecell{Ch.\\names} \\
\midrule

1 & Comparison and benchmark of name-to-gender inference services~\cite{santamaria2018comparison} & 2018 & \checkmark & \checkmark & \checkmark & \textcolor{lightgray}{\ding{53}} & \checkmark & \textcolor{lightgray}{\ding{53}} \\

2 & Augmenting Data Quality through High-Precision Gender Categorization~\cite{WOS:000468045500004} & 2019 & \checkmark & \textcolor{lightgray}{\ding{53}} & \textcolor{lightgray}{\ding{53}} & \checkmark & \textcolor{lightgray}{\ding{53}} & \textcolor{lightgray}{\ding{53}} \\

3 & The sound of gender: inferring the gender of names in a foreign language~\cite{WOS:000671253500004} & 2019 & \textcolor{lightgray}{\ding{53}} & \textcolor{lightgray}{\ding{53}} & \textcolor{lightgray}{\ding{53}} & \textcolor{lightgray}{\ding{53}} & \textcolor{lightgray}{\ding{53}} & \checkmark \\

4 & The impact of gender orientation of names on individuals' evaluation of impressions and interpersonal interaction~\cite{WOS:000635500200005} & 2021 & \textcolor{lightgray}{\ding{53}} & \textcolor{lightgray}{\ding{53}} & \textcolor{lightgray}{\ding{53}} & \textcolor{lightgray}{\ding{53}} & \textcolor{lightgray}{\ding{53}} & \checkmark \\

5 & Performance of gender detection tools: a comparative study of name-to-gender inference services~\cite{WOS:000705202000009} & 2021 & \checkmark & \checkmark & \textcolor{lightgray}{\ding{53}} & \textcolor{lightgray}{\ding{53}} & \checkmark & \textcolor{lightgray}{\ding{53}} \\

6 & Using genderize.io to infer the gender of first names: how to improve the accuracy of the inference~\cite{WOS:000721567300011} & 2021 & \checkmark & \checkmark & \textcolor{lightgray}{\ding{53}} & \checkmark & \textcolor{lightgray}{\ding{53}} & \textcolor{lightgray}{\ding{53}} \\

7 & How accurate are gender detection tools in predicting the gender for Chinese names? A study with 20,000 given names in Pinyin format~\cite{WOS:000792126300005} & 2022 & \checkmark & \checkmark & \textcolor{lightgray}{\ding{53}} & \textcolor{lightgray}{\ding{53}} & \checkmark & \checkmark \\

8 & Gender equity in rheumatology leadership in the Asia-Pacific~\cite{WOS:000879846700001} & 2022 & \checkmark & \textcolor{lightgray}{\ding{53}} & \textcolor{lightgray}{\ding{53}} & \textcolor{lightgray}{\ding{53}} & \textcolor{lightgray}{\ding{53}} & \textcolor{lightgray}{\ding{53}} \\

9 & Gender parity perspectives in Asia Pacific: a descriptive analysis of Philippine rheumatology conferences~\cite{WOS:000948666700002} & 2023 & \checkmark & \checkmark & \textcolor{lightgray}{\ding{53}} & \textcolor{lightgray}{\ding{53}} & \textcolor{lightgray}{\ding{53}} & \textcolor{lightgray}{\ding{53}} \\

10 & Ethnicity and gender trends of UK authors in The British Medical Journal and the Lancet over the past two decades: a comprehensive longitudinal analysis~\cite{WOS:001077484000001} & 2023 & \checkmark & \checkmark & \textcolor{lightgray}{\ding{53}} & \textcolor{lightgray}{\ding{53}} & \textcolor{lightgray}{\ding{53}} & \textcolor{lightgray}{\ding{53}} \\

\bottomrule
\end{tabular}

\end{table}

\section{Gender score regression (extended)}
\label{app:sec:reg}

\begin{figure}[t]
    \centering
    \begin{subfigure}[t]{0.32\textwidth}
        \centering
        \includegraphics[width=\textwidth]{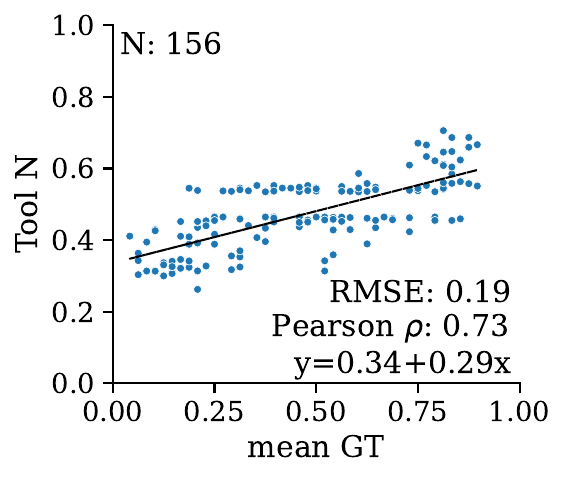}
        \caption{Tool N}
        \label{app:fig:tool:N}
    \end{subfigure}
    \hfill
    \begin{subfigure}[t]{0.32\textwidth}
        \centering
        \includegraphics[width=\textwidth]{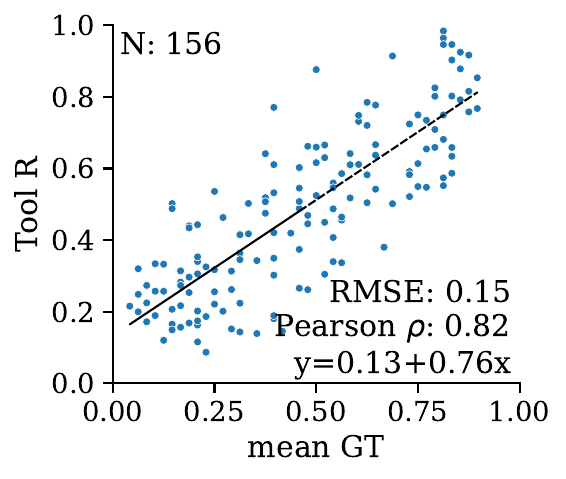}
        \caption{Tool R}
        \label{app:fig:tool:R}
    \end{subfigure}
    \hfill
    \begin{subfigure}[t]{0.32\textwidth}
        \centering
        \includegraphics[width=\textwidth]{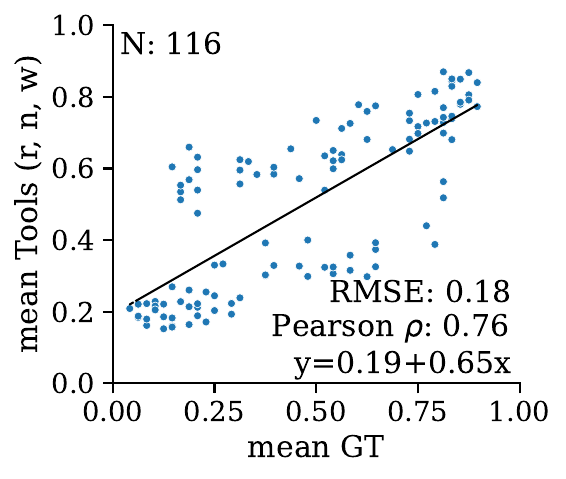}
        \caption{Mean RNW}
        \label{app:fig:tools:NRW}
    \end{subfigure}
    
    \begin{subfigure}[t]{0.32\textwidth}
        \centering
        \includegraphics[width=\textwidth]{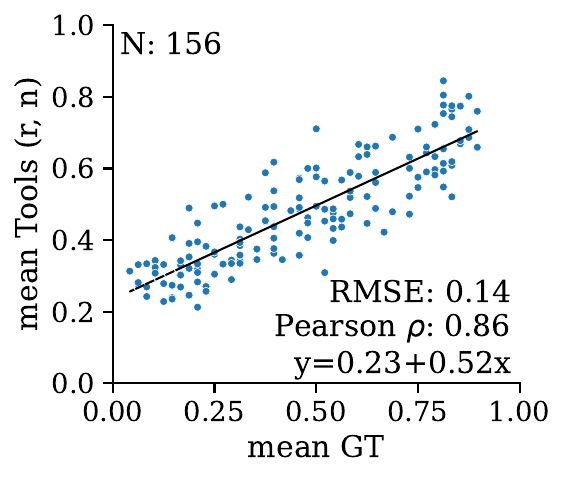}
        \caption{Mean RN}
        \label{app:fig:tools:RN}
    \end{subfigure}
    \hfill
    \begin{subfigure}[t]{0.32\textwidth}
        \centering
        \includegraphics[width=\textwidth]{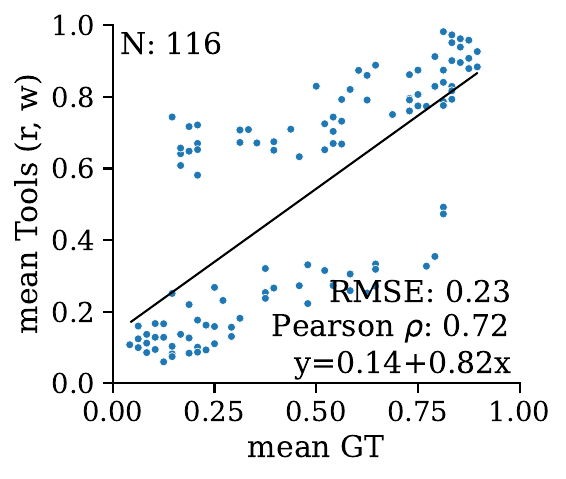}
        \caption{Mean RW}
        \label{app:fig:tools:RW}
    \end{subfigure}
    \hfill
    \begin{subfigure}[t]{0.32\textwidth}
        \centering
        \includegraphics[width=\textwidth]{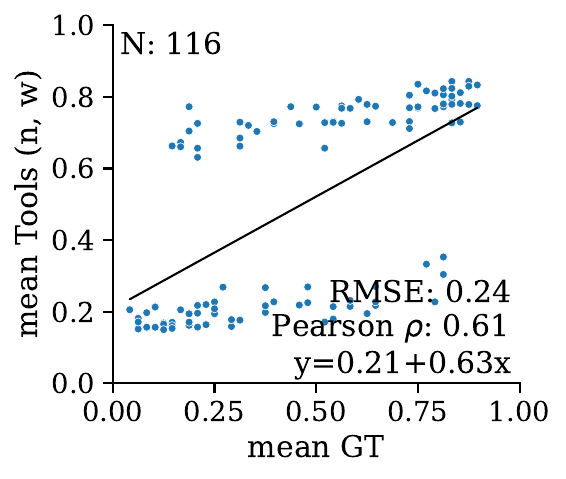}
        \caption{Mean NW}
        \label{app:fig:tools:NW}
    \end{subfigure}
    \caption{\textbf{Agreement between tool-based gender inference and annotator-perceived gender.}
    Comparison between gender scores produced by inference tools and the mean perceived gender assigned by human annotators (gold standard). Each point represents a name; the x-axis shows the average annotator score, and the y-axis the score produced by the tool or tool aggregation. Panels show individual tools (a) N and (b) R, and aggregated estimates obtained by averaging tools: (c) RNW, (d) RN, (e) RW, and (f) NW. Solid lines indicate linear fits; Pearson correlation ($\rho$) and RMSE quantify agreement with the gold standard. Aggregating tools generally improves alignment with annotator judgments, with the RN combination showing the strongest correlation and lowest error.
    Panels involving tool W (c, e, f) show fewer names ($N = 116$) because rare single-character names were excluded, as W defaults to male for all such names.
    }
    \label{app:fig:tools_performance}
\end{figure}

\subsection{Hanzi features' predictive power}
\label{app:sec:tools_performance}

To predict the perceived gender of Hanzi names, we leverage existing gender-inference tools as input features for our models. Each of the three tools captures a different signal. \texttt{Namsor}~(\texttt{N}), a widely adopted commercial API, has shown strong performance in bibliometric gender inference tasks \cite{vanhelene2024inferring}. Tools \texttt{R}~\cite{bao-2023} and \texttt{W}~\cite{chinesetools_gender_guesser} appear to derive their estimates from population-level name–gender distributions. Since each tool individually achieves reasonable agreement with annotator judgments \Cref{app:fig:tool:N,app:fig:tool:R}, we hypothesized that combining them could yield stronger predictive features. We therefore constructed pairwise averages \Cref{app:fig:tools:RN,app:fig:tools:RW,app:fig:tools:NW} and a three-tool aggregate \Cref{app:fig:tools:NRW}, evaluating each against the ground-truth annotations. The RN combination achieved the lowest RMSE and highest correlation with the gold standard, confirming its predictive value.

However, rather than selecting a single best combination for downstream analysis, we trained multiple models, each using a distinct subset of tool-derived features. This design allowed us to assess not only numeric scores (means of two or more tools) but also categorical features: \texttt{N}~returns both a numeric score and a categorical label, \texttt{R}~provides only a numeric score, and \texttt{W}~outputs a binary prediction. When combining tools, we applied majority voting for labels and arithmetic means for numeric scores. We additionally included a single-character flag after observing that tool \texttt{W}~defaults to male for rare single-character names \Cref{sec:features}.

Across all model configurations, \texttt{mean\_ngender\_rn}---the average of the numeric scores from tools \texttt{R}~and \texttt{N}---emerged as the dominant predictor in both regression and classification tasks. For regression on the mean of annotator scores (\Cref{fig:model_pipeline}c), it ranked first in feature importance, followed by the individual numeric scores from \texttt{R}~and \texttt{N}~(\Cref{fig:hz_reg}); the best model achieved $R^2=0.85$ and $\text{RMSE}=0.09$ on the held-out set. For classification across multiple discretization schemes (\Cref{app:fig:discretization}), it ranked first, this time followed by \texttt{gender\_nw}~(majority vote of \texttt{N}~and \texttt{W}) and \texttt{gender\_n}~(the categorical label from \texttt{N}), as shown in \Cref{app:fig:class}; under equal-width discretization of the mean annotation scores (\Cref{app:fig:discretization:cut1}), the classifier attained a macro AUC of $0.97$ and Cohen's $\kappa$~of $0.81$. The consistency of \texttt{mean\_ngender\_rn}~as the leading feature across tasks and discretization choices confirms that fusing complementary tool signals---one commercial, one population-based---produces a more robust gender estimate than any individual tool alone.

\subsection{Pinyin-based gender score}
\label{app:sec:py-results}

\subsubsection{Comparing \texorpdfstring{$\hat{y}_p$}{ŷₚ} with \texorpdfstring{$y_h$}{yₕ}}
\label{app:sec:reg:gt-py}

Since the gender associated with a pinyin name is ambiguous without knowing the underlying hanzi, we first evaluate our aggregation strategies on the held-out set, where the mapping is known. \Cref{app:tbl:pinyin_stats_test_set} lists the $34$~held-out pinyin names, each with its gold-standard score ($y_h$), aggregated predictions ($\hat{y}_p$), and dispersion measures---spread ($\Delta$), standard deviation, and ambiguity (\Cref{sec:pinyin-model})---of the hanzi-level predictions ($\hat{y}_h$). The key question is whether the aggregated scores recover the gold-standard signal despite this heterogeneity---and, when they cannot, whether the dispersion measures signal the unreliability.

\begin{table}[hb!]
\centering
\caption{\textbf{Pinyin-based gender score inference on the held-out set.}
For each of the 34 pinyin names in the test set, we report the number of associated hanzi names in our corpus ($n_H$), the mean gold-standard gender score ($y_h$), and whether the name is single-character (\textsc{sc}). 
For the hanzi-level predictions ($\hat{y}_h$), we include the range, standard deviation, spread ($\Delta$), and ambiguity (Amb.). 
We also report our four pinyin-level aggregation strategies ($\hat{y}_p$): unweighted averages over CNKI and CNKI+, and weighted averages using CNKI and Population frequencies (\Cref{sec:pinyin-model}).
}
\label{app:tbl:pinyin_stats_test_set}
\small
\begin{tabular}{lrccrrrrrrrrr}
\toprule
 & & & & \multicolumn{5}{c}{$\hat{y}_h$} & \multicolumn{4}{c}{$\hat{y}_p$} \\
\cmidrule(lr){5-9} \cmidrule(lr){10-13}
Pinyin & $n_H$ & $y_h$ & \textsc{sc} & min & max & std & $\Delta$ & Amb. & M\textsubscript{CNKI} & M\textsubscript{CNKI+} & W\textsubscript{CNKI} & W\textsubscript{Pop.} \\
\midrule
fang & 19 & 0.40 & \checkmark & 0.28 & 0.65 & 0.11 & 0.37 & 0.21 & 0.42 & 0.41 & 0.56 & 0.61 \\
jie & 52 & 0.46 & \checkmark & 0.27 & 0.67 & 0.09 & 0.40 & 0.23 & 0.43 & 0.38 & 0.40 & 0.36 \\
jin & 43 & 0.46 & \checkmark & 0.26 & 0.62 & 0.10 & 0.36 & 0.27 & 0.37 & 0.40 & 0.36 & 0.40 \\
jun & 22 & 0.27 & \checkmark & 0.25 & 0.60 & 0.09 & 0.35 & 0.22 & 0.35 & 0.36 & 0.33 & 0.34 \\
meng & 25 & 0.79 & \checkmark & 0.28 & 0.62 & 0.10 & 0.34 & 0.25 & 0.47 & 0.40 & 0.55 & 0.56 \\
qi & 84 & 0.48 & \checkmark & 0.24 & 0.64 & 0.08 & 0.40 & 0.22 & 0.36 & 0.36 & 0.43 & 0.38 \\
shen & 35 & 0.25 & \checkmark & 0.28 & 0.54 & 0.06 & 0.26 & 0.11 & 0.33 & 0.35 & 0.31 & 0.29 \\
xi & 101 & 0.38 & \checkmark & 0.25 & 0.67 & 0.08 & 0.42 & 0.18 & 0.40 & 0.38 & 0.46 & 0.39 \\
xuan & 40 & 0.31 & \checkmark & 0.29 & 0.66 & 0.10 & 0.37 & 0.30 & 0.43 & 0.40 & 0.49 & 0.43 \\
zhen & 42 & 0.60 & \checkmark & 0.26 & 0.64 & 0.11 & 0.38 & 0.32 & 0.41 & 0.40 & 0.37 & 0.53 \\
\midrule
bingchen & 2 & 0.17 & \ding{53} & 0.26 & 0.27 & 0.01 & 0.01 & 0.01 & 0.26 & 0.26 & 0.26 & 0.26 \\
bohua & 2 & 0.17 & \ding{53} & 0.28 & 0.29 & 0.01 & 0.01 & 0.01 & 0.28 & 0.28 & 0.28 & 0.28 \\
chunfang & 2 & 0.88 & \ding{53} & 0.27 & 0.80 & 0.37 & 0.53 & 0.42 & 0.53 & 0.53 & 0.77 & 0.76 \\
dongxia & 3 & 0.81 & \ding{53} & 0.55 & 0.82 & 0.14 & 0.27 & 0.22 & 0.70 & 0.70 & 0.76 & 0.73 \\
dongyan & 6 & 0.83 & \ding{53} & 0.26 & 0.82 & 0.21 & 0.56 & 0.39 & 0.63 & 0.63 & 0.60 & 0.69 \\
feinan & 2 & 0.58 & \ding{53} & 0.36 & 0.56 & 0.14 & 0.20 & 0.16 & 0.46 & 0.46 & 0.46 & 0.49 \\
haian & 1 & 0.19 & \ding{53} & 0.17 & 0.17 & -- & 0.00 & 0.00 & 0.17 & 0.17 & 0.17 & 0.17 \\
huazhen & 2 & 0.10 & \ding{53} & 0.19 & 0.77 & 0.41 & 0.58 & 0.47 & 0.48 & 0.48 & 0.58 & 0.54 \\
jianju & 1 & 0.12 & \ding{53} & 0.19 & 0.19 & -- & 0.00 & 0.00 & 0.19 & 0.19 & 0.19 & 0.19 \\
nianman & 1 & 0.33 & \ding{53} & 0.39 & 0.39 & -- & 0.00 & 0.00 & 0.39 & 0.39 & 0.39 & 0.39 \\
qingdong & 3 & 0.06 & \ding{53} & 0.16 & 0.20 & 0.03 & 0.04 & 0.04 & 0.17 & 0.17 & 0.16 & 0.18 \\
rongwen & 2 & 0.52 & \ding{53} & 0.32 & 0.51 & 0.13 & 0.19 & 0.15 & 0.41 & 0.41 & 0.41 & 0.37 \\
sisi & 2 & 0.83 & \ding{53} & 0.44 & 0.64 & 0.14 & 0.20 & 0.16 & 0.54 & 0.54 & 0.62 & 0.60 \\
taimeng & 1 & 0.38 & \ding{53} & 0.45 & 0.45 & -- & 0.00 & 0.00 & 0.45 & 0.45 & 0.45 & 0.45 \\
wangmin & 3 & 0.73 & \ding{53} & 0.19 & 0.72 & 0.30 & 0.53 & 0.43 & 0.38 & 0.38 & 0.31 & 0.36 \\
wenhan & 4 & 0.56 & \ding{53} & 0.15 & 0.47 & 0.16 & 0.32 & 0.23 & 0.23 & 0.23 & 0.22 & 0.27 \\
xuetian & 1 & 0.54 & \ding{53} & 0.51 & 0.51 & -- & 0.00 & 0.00 & 0.51 & 0.51 & 0.51 & 0.51 \\
yannan & 10 & 0.38 & \ding{53} & 0.22 & 0.80 & 0.19 & 0.58 & 0.44 & 0.56 & 0.56 & 0.53 & 0.63 \\
yingzhen & 1 & 0.85 & \ding{53} & 0.84 & 0.84 & -- & 0.00 & 0.00 & 0.84 & 0.84 & 0.84 & 0.84 \\
yiwei & 24 & 0.54 & \ding{53} & 0.15 & 0.70 & 0.16 & 0.55 & 0.35 & 0.28 & 0.28 & 0.29 & 0.24 \\
yuewen & 5 & 0.29 & \ding{53} & 0.17 & 0.83 & 0.27 & 0.66 & 0.54 & 0.45 & 0.45 & 0.45 & 0.43 \\
zehui & 5 & 0.81 & \ding{53} & 0.16 & 0.65 & 0.21 & 0.50 & 0.32 & 0.27 & 0.27 & 0.28 & 0.35 \\
zhangqi & 3 & 0.56 & \ding{53} & 0.15 & 0.44 & 0.16 & 0.29 & 0.23 & 0.25 & 0.25 & 0.33 & 0.19 \\
zuying & 2 & 0.21 & \ding{53} & 0.28 & 0.62 & 0.24 & 0.34 & 0.27 & 0.45 & 0.45 & 0.45 & 0.55 \\
\bottomrule
\end{tabular}
\end{table}

Single-character pinyin names, which map to dozens of hanzi variants (up to 101 for \textit{xi}), pose the greatest challenge, as all four strategies converge toward the center of the scale (${\sim}0.3$--$0.4$), diluting strong gender signals. For instance, \textit{meng} ($y_h = 0.79$) is predicted at only $M_\text{CNKI} = 0.47$ and $M_\text{CNKI+} = 0.40$, because averaging over 25 hanzi variants with mixed gender associations ($\text{std} = 0.10$, $\Delta = 0.34$, Amb.$= 0.25$) attenuates the score toward a middle value. Weighted strategies partially recover this signal, raising the prediction to $W_\text{CNKI} = 0.55$ and $W_\text{Pop.} = 0.56$. However, weighting alone cannot fully compensate when when the ambiguity remains high.

Two-character pinyin names with few hanzi variants are comparatively straightforward. Names mapping to a single hanzi (e.g., \textit{yingzhen}, \textit{haian}) produce identical predictions across strategies with zero ambiguity, and even names with 2--3 variants show minimal divergence. Weighted strategies also show clearer gains. For \textit{chunfang} ($y_h = 0.88$), population-weighted CNKI ($W_\text{CNKI} = 0.77$) outperforms the unweighted mean ($M_\text{CNKI} = 0.53$), and similar patterns appear for \textit{dongxia} and \textit{fang}. Yet even among two-character names, difficult cases persist. \textit{Zehui} ($y_h = 0.81$) is predicted at only $0.27$--$0.35$ across all strategies, as its five hanzi variants span a wide range of gender associations ($\hat{y}_h \in [0.16, 0.65]$, $\text{std} = 0.21$, $\Delta = 0.50$, Amb.$= 0.32$), attenuating the strongly female signal.

These results show that aggregating hanzi-based predictions ($\hat{y}_h$) into a pinyin-based gender score ($\hat{y}_p$) is effective in many cases, yet a point estimate alone is insufficient. %
We recommend complementing the aggregated score with a measure of how concentrated or dispersed the underlying hanzi-based scores are. The standard deviation and spread ($\Delta$) quantify dispersion but are sensitive to outliers and do not account for the shape of the distribution. 
The ambiguity score we propose (\Cref{sec:pinyin-model}) addresses these limitations by focusing on the central 80\% of hanzi-based predictions ($q_{90} - q_{10}$), discounting outliers while preserving the distributional shape. Low ambiguity indicates agreement across hanzi variants, making the aggregated score trustworthy. High ambiguity reveals fundamental disagreement that no aggregation strategy can fully resolve.

\subsubsection{Comparing \texorpdfstring{$\hat{y}_p$}{ŷₚ} with \texorpdfstring{$\hat{y}_h$}{ŷₕ}}
\label{app:sec:reg:hz-py}

\begin{figure}[t]
    \centering
    
    \begin{subfigure}[t]{0.24\textwidth}
        \centering
        \includegraphics[height=0.17\textheight]{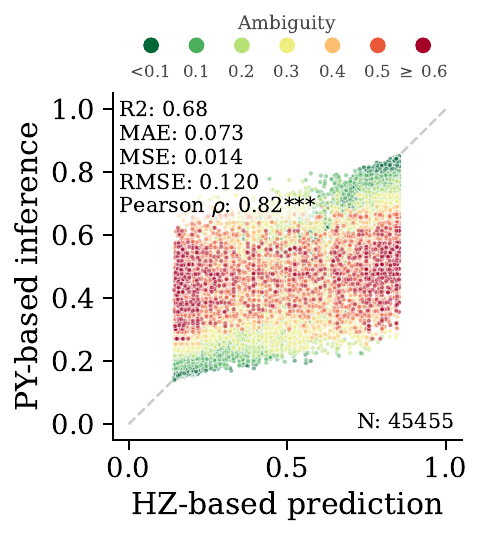}
        \caption{Mean CNKI}
        \label{fig:py-avg-cnki}
    \end{subfigure}
    \hfill
    \begin{subfigure}[t]{0.24\textwidth}
        \centering
        \includegraphics[height=0.17\textheight]{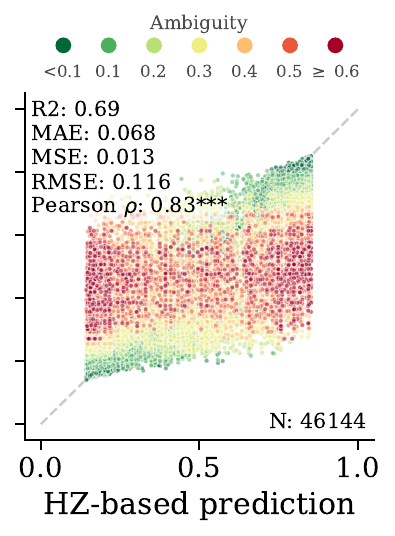}
        \caption{Mean CNKI+}
        \label{fig:py-avg-cnkiplus}
    \end{subfigure}
    \hfill
    \begin{subfigure}[t]{0.24\textwidth}
        \centering
        \includegraphics[height=0.17\textheight]{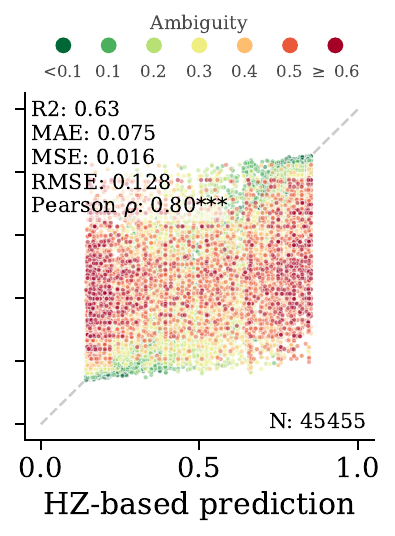}
        \caption{Weighted CNKI}
        \label{fig:py-weighted-cnki}
    \end{subfigure}
    \hfill\begin{subfigure}[t]{0.24\textwidth}
        \centering
        \includegraphics[height=0.17\textheight]{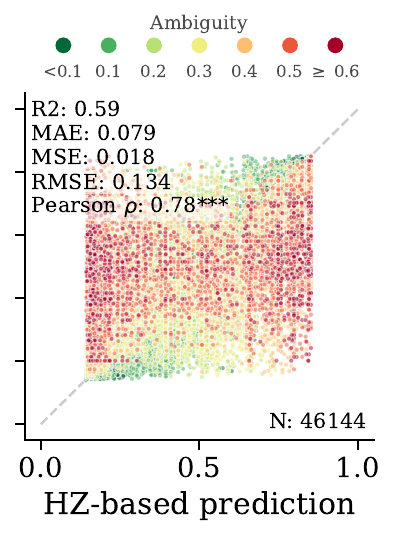}
        \caption{Weighted pop.}
        \label{fig:py-weighted-all}
    \end{subfigure}
    \caption{\textbf{Hanzi- vs.\ pinyin-based gender score inference.} 
    Hanzi-based predictions $\hat{y}_h$ (x-axis) versus pinyin-based predictions $\hat{y}_p$ (y-axis) for each aggregation strategy: (a) Mean CNKI, (b) Mean CNKI+, (c) Weighted CNKI by author counts, and (d) Weighted by population counts. 
    Each point represents one unique hanzi given name in the CNKI corpus. 
    Color encodes the ambiguity of the aggregated pinyin-level gender score, defined as the difference between the 90th and 10th percetiles ($q_{90} - q_{10}$) of the hanzi-level predictions for a given pinyin name, binned into intervals (${<}0.1$, $[0.1, 0.2)$, $[0.2, 0.3)$, $[0.3, 0.4)$, $[0.4, 0.5)$, $[0.5, 0.6)$, ${\geq}0.6$).
    The dashed line indicates perfect agreement. 
    Each panel reports goodness of fit ($R^2$) and error metrics (MAE, MSE, RMSE). 
    Mean CNKI+ (b) yields the highest agreement among all strategies (lowest RMSE). 
    Panels (a) and (c) show fewer name pairs because of the single-character augmentation (\Cref{sec:chinese-given-names}).
    }
    \label{app:fig:pinyin_inference_all}
\end{figure}

\Cref{app:fig:pinyin_inference_all} extends the evaluation beyond the held-out set by comparing, for every unique hanzi name in the CNKI corpus, its hanzi-based prediction ($\hat{y}_h$) against the pinyin-based score ($\hat{y}_p$) assigned by each aggregation strategy. While the previous analysis (\Cref{app:sec:reg:gt-py}) asked whether $\hat{y}_p$ recovers a known gold-standard label ($y_h$), this comparison asks how much information is lost when the hanzi identity is unknown and only the pinyin-level estimate $\hat{y}_p$ is available.

Across all four strategies, predictions are compressed toward the center of the scale. In the mid-range ($0.25 \lesssim \hat{y}_h \lesssim 0.6$), we observe both $\hat{y}_p > \hat{y}_h$ and $\hat{y}_p < \hat{y}_h$, meaning that some hanzi names associated with male are inferred as female at the pinyin level, and vice versa. This is expected for gender-neutral names, where the hanzi variants sharing a single pinyin form span a wide range of gender associations---some female, some male, some neutral---introducing substantial noise into the aggregated score. At the extremes ($\hat{y}_h < 0.25$ or $\hat{y}_h > 0.65$), the gender association at the hanzi level is stronger, and the pinyin-level estimates align more closely with the hanzi-based predictions, particularly for the unweighted means. Our ambiguity score (\Cref{sec:pinyin-model}), which measures how much the hanzi-based predictions for a given pinyin disagree, captures this pattern directly. As the color encoding in \Cref{app:fig:pinyin_inference_all} shows, low ambiguity (green) concentrates at the extremes where hanzi variants agree, while high ambiguity (red) dominates the mid-range where they diverge.

Among the four strategies, Mean CNKI+ aligns most closely with the hanzi-based predictions ($R^2 = 0.69$, $\text{RMSE} = 0.116$, $\rho = 0.83$), marginally outperforming Mean CNKI. The weighted strategies perform worse ($R^2 = 0.63$ and $0.59$ for Weighted CNKI and Weighted population, respectively), likely because frequency-based weighting amplifies the dominance of a few high-frequency hanzi names rather than producing a more representative gender estimate, particularly when the frequency distribution is skewed.

\section{Gender label classification}
\label{app:sec:class-models}

\begin{figure}[t]
    \centering
    \begin{subfigure}[t]{0.32\textwidth}
        \centering
        \includegraphics[width=\textwidth]{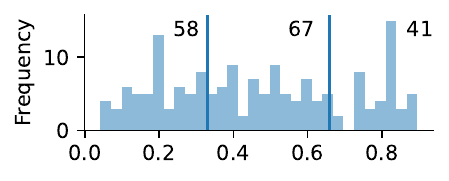}
        \caption{Cut 1: mean, (0.33, 0.66)}
        \label{app:fig:discretization:cut1}
    \end{subfigure}
    \hfill
    \begin{subfigure}[t]{0.32\textwidth}
        \centering
        \includegraphics[width=\textwidth]{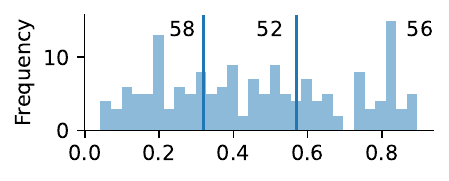}
        \caption{Cut 2: mean, (0.32, 0.57)}
        \label{app:fig:discretization:cut2}
    \end{subfigure}
    \hfill
    \begin{subfigure}[t]{0.32\textwidth}
        \centering
        \includegraphics[width=\textwidth]{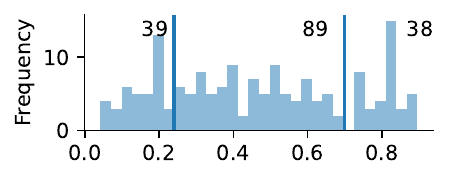}
        \caption{Cut 3: mean, (0.24, 0.7)}
        \label{app:fig:discretization:cut3}
    \end{subfigure}

    \begin{subfigure}[t]{0.32\textwidth}
        \centering
        \includegraphics[width=\textwidth]{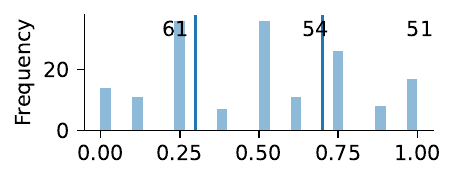}
        \caption{Cut 4: median, (0.3, 0.7)}
        \label{app:fig:discretization:cut4}
    \end{subfigure}
    \hfill
    \begin{subfigure}[t]{0.32\textwidth}
        \centering
        \includegraphics[width=\textwidth]{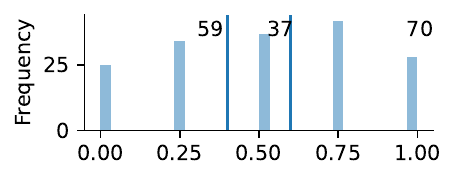}
        \caption{Cut 5: mode, (0.4, 0.6)}
        \label{app:fig:discretization:cut5}
    \end{subfigure}
    \hfill
    \begin{subfigure}[t]{0.32\textwidth}
        \centering
        \includegraphics[width=\textwidth]{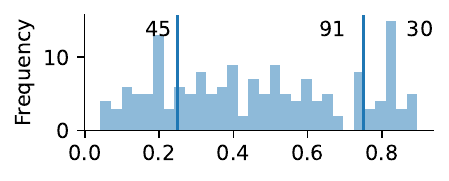}
        \caption{Cut 6: mean, (0.25, 0.75)}
        \label{app:fig:discretization:cut6}
    \end{subfigure}

    \begin{subfigure}[t]{0.32\textwidth}
        \centering
        \includegraphics[width=\textwidth]{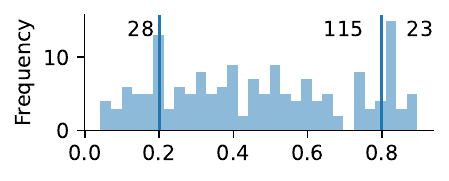}
        \caption{Cut 7: mean, (0.2, 0.8)}
        \label{app:fig:discretization:cut7}
    \end{subfigure}
    \begin{subfigure}[t]{0.32\textwidth}
        \centering
        \includegraphics[width=\textwidth]{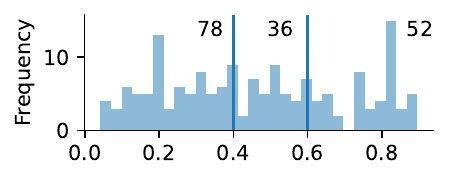}
        \caption{Cut 8: mean, (0.4, 0.6)}
        \label{app:fig:discretization:cut8}
    \end{subfigure}

    \caption{\textbf{Discretization of gender scores.} Aggregated human annotations were mapped into the classes \textit{male}, \neutral, and \textit{female}~using different discretization strategies. (a–c) Equal-width, equal-size, and visually derived cuts based on the mean. (d–e) Visual cuts using the median and mode, respectively. (f) Fuzzy cut following annotation instructions ($0$ = definitely male, $0.25$ = more likely male, $0.5$ = neutral, $0.75$ = more likely female, $1$ = definitely female). (g) Conservative fuzzy cut assigning 20\% to male and female extremes. (h) Conservative fuzzy cut assigning 20\% to the neutral middle. These discretizations were later used as target variables in model training.}
    \label{app:fig:discretization}
\end{figure}

\subsection{Gender score discretization}
\label{app:sec:class:hz:discretization}

When collecting the ground truth (\Cref{{sec:gold-standards}}), we opted for numeric gender scores rather than categorical labels, as gender associations in names---particularly in Pinyin---are not binary but exist on a spectrum. However, converting these scores into discrete classes for classification is itself non-trivial. Retaining all five ordinal levels would introduce evaluation ambiguity, as distinctions such as \textit{likely female} vs.\ \textit{most likely female} are neither semantically clear nor practically useful. We therefore discretize into three classes: \textit{male}, \textit{neutral}, and \textit{female}. The neutral class is paramount: genuinely gender-neutral names exist, and the inherent ambiguity of Pinyin---where a single romanization may map to characters spanning the full gender spectrum---makes a binary scheme inadequate, as it would force borderline names into \textit{male} or \textit{female} and obscure a meaningful signal.

The core challenge is selecting appropriate bin boundaries. We explore eight discretization strategies (\Cref{app:fig:discretization}), varying the boundary-selection criterion and resulting class balance. Cuts~1--3 use equal-width bins based on the mean of the annotated scores $y_h$, with progressively asymmetric thresholds that shift more samples toward the neutral class. Cut~4 uses the median and Cut~5 the mode, producing different placements that reflect the skewed, multimodal score distribution. Cut~6 places boundaries at $0.25$ and $0.75$, which correspond directly to the original annotation values of 2 and 4 after normalizing the 1--5 scale to $[0, 1]$. Cuts~7 and~8 adopt conservative fuzzy strategies: Cut~7 assigns only the most extreme 20\% to the male and female classes, yielding a dominant neutral class, while Cut~8 reverses this by reserving 20\% for the neutral middle.

The discretization choice strongly affects class proportions. Narrower neutral regions (e.g., Cuts~5 and~8) risk misclassifying borderline cases, while wider ones (e.g., Cuts~3 and~7) absorb borderline cases into neutral, leaving fewer samples in the male and female classes. Rather than selecting a single binning \textit{a priori}, we retain all eight and use each separately as the target variable when training our classifiers on the Hanzi-derived feature set $\mathbf{X}$ (\Cref{app:sec:tools_performance}), allowing us to assess model robustness across discretization choices.

\subsection{Hanzi-based inference (gender label \texorpdfstring{$\hat{l}_h$}{l̂ₕ})}
\label{app:sec:hanzi:class}

Using the eight discretization schemes described in \Cref{app:sec:class:hz:discretization}, we train three-class \texttt{CatBoost} classifiers to predict the perceived gender label $\hat{l}_c \in \{\textit{male}, \textit{neutral}, \textit{female}\}$. 
As with regression, we use an 80\%--20\% stratified split, grid search over the same hyperparameters, 3-fold cross-validation, and early stopping, minimizing the \texttt{MultiClass} objective (three-class cross-entropy). 
We repeat training for each feature combination, discretization scheme, and random seed (three seeds total), then select as the final model the run with the highest macro ROC AUC on the held-out test set---jointly identifying the best discretization and feature set. 
We report macro-averaged ROC AUC, confusion matrix, and Cohen's $\kappa$ (agreement between predicted and true labels corrected for chance).

\subsubsection{Performance}
\label{app:sec:classeval}

\begin{figure}[t]
    \centering
    \begin{subfigure}[t]{0.24\textwidth}
        \centering
        \includegraphics[width=\textwidth]{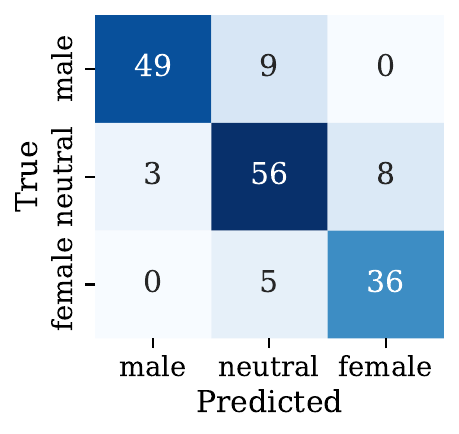}
        \caption{All Data}
        \label{app:fig:class-all}
    \end{subfigure}
    \begin{subfigure}[t]{0.24\textwidth}
        \centering
        \includegraphics[width=\textwidth]{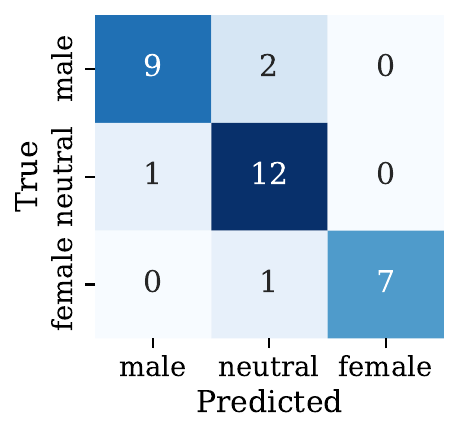}
        \caption{Test Set}
        \label{app:fig:class-val}
    \end{subfigure}
    \begin{subfigure}[t]{0.5\textwidth}
        \centering
        \includegraphics[width=\textwidth]{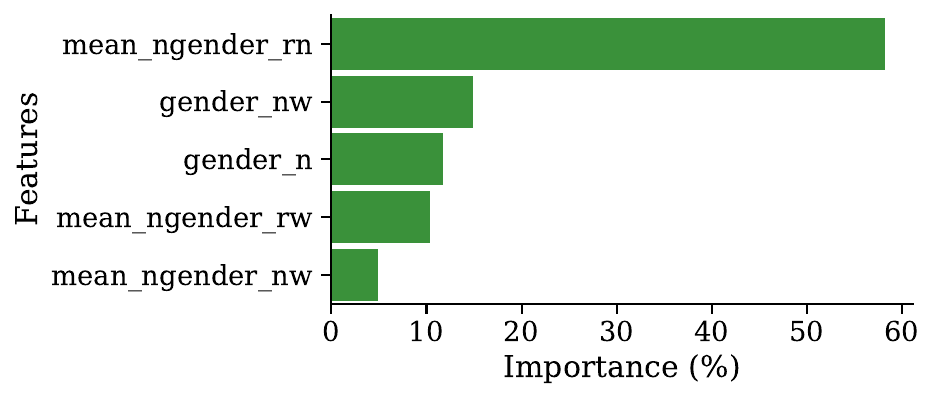}
        \caption{Feature Importance (CatBoost)}
        \label{app:fig:class-fi}
    \end{subfigure}
    \hfill
    \begin{subfigure}[t]{0.24\textwidth}
        \centering
        \includegraphics[width=\textwidth]{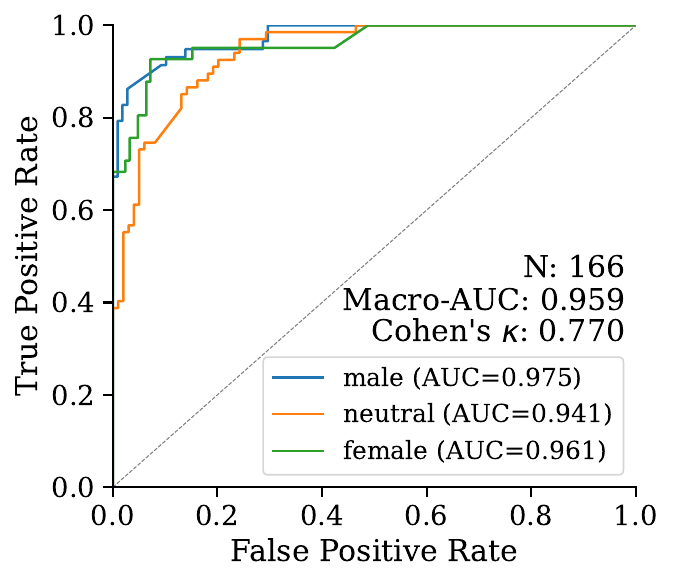}
        \caption{ROC-AUC (all data)}
        \label{app:fig:class-auc-all}
    \end{subfigure}
    \begin{subfigure}[t]{0.24\textwidth}
        \centering
        \includegraphics[width=\textwidth]{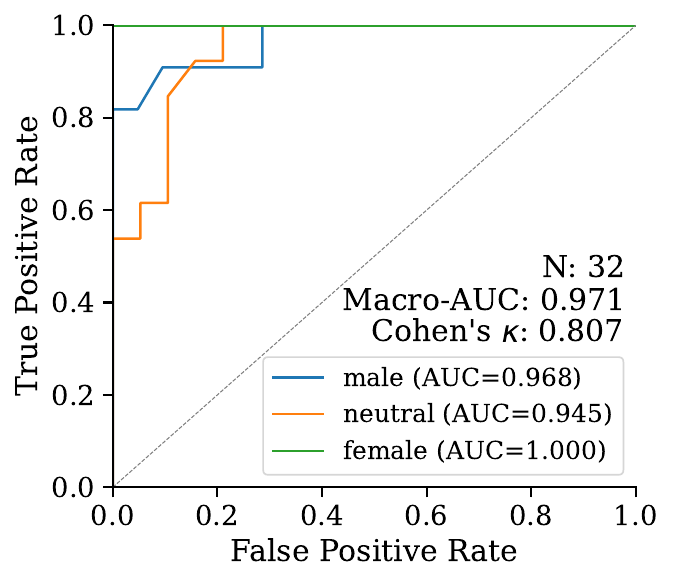}
        \caption{ROC-AUC (test set)}
        \label{app:fig:class-auc-val}
    \end{subfigure}
    \begin{subfigure}[t]{0.5\textwidth}
        \centering
        \includegraphics[width=\textwidth]{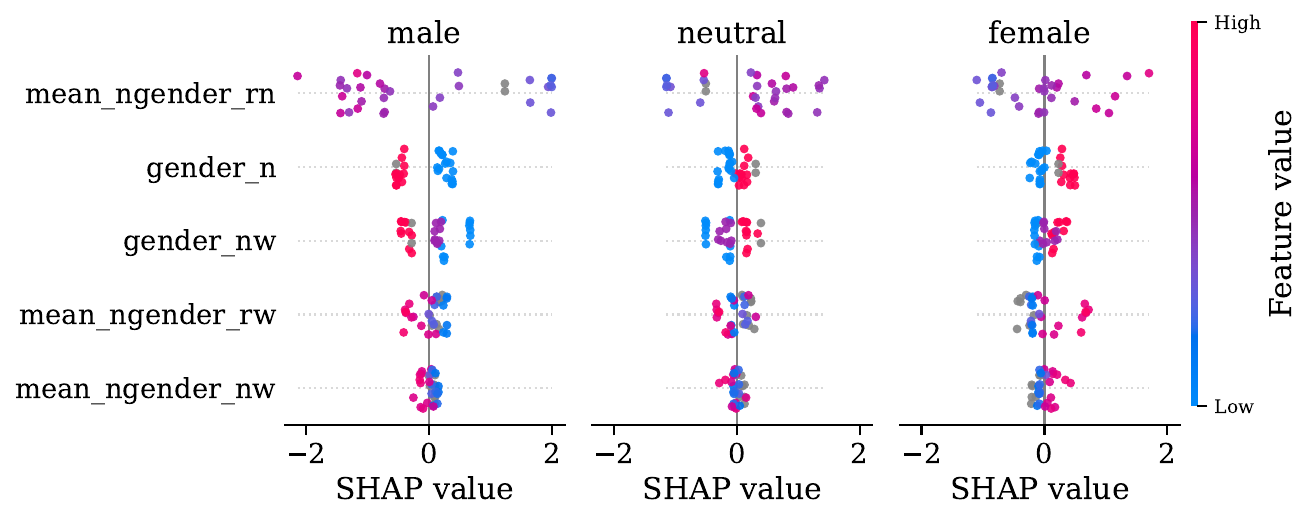}
        \caption{Feature Importance (SHAP)}
        \label{app:fig:class-shap}
    \end{subfigure}
    \caption{
    \textbf{Performance of hanzi-based gender label classification ($\hat{l}_h$).} 
    The best results were obtained with Cut 1 (mean-based equal-width discretization; see Figure~\ref{app:fig:discretization}a). 
    (a,b) Confusion matrices for the full ground-truth set (train + test) and the test set, respectively.
    (c) CatBoost feature importance, showing the aggregated mean predictions from
    \texttt{R}~and \texttt{N}~(\texttt{mean\_ngender\_rn}) is the strongest predictor.
    (d,e) One-vs-rest ROC curves for the entire ground-truth set and the test set, respectively.
    (f) SHAP summary plot indicating consistent class-discriminative effects of the main aggregated predictors.
    Overall, the classifier achieves high and stable performance across classes, with predictive power primarily driven by features from tools \texttt{R}~and \texttt{N}.
    }
    \label{app:fig:class}
\end{figure}

Among all 8 discretization cuts and 1,023 possible feature subsets, the best-performing configuration uses Cut~1 (\Cref{app:fig:discretization:cut1}) with five features: three numeric (\texttt{mean\_ngender\_rn}, \texttt{mean\_ngender\_rw}, \texttt{mean\_ngender\_nw}) and two categorical (\texttt{gender\_nw}~and \texttt{gender\_n}).

The confusion matrices (\Cref{app:fig:class-all,app:fig:class-val}) indicate near-perfect classification across all three gender-label classes. On the held-out test set, only two samples are misclassified out of 32, confirming that the model generalizes well beyond the training data.
This is further supported by the ROC curves (\Cref{app:fig:class-auc-all,app:fig:class-auc-val}), where per-class AUC values range from 0.945 to 1.000 on the test set, yielding a macro-averaged AUC of 0.971. A Cohen's $\kappa$ of 0.807 indicates substantial agreement between predicted and true labels well beyond chance, reinforcing that the classifier captures meaningful structure rather than exploiting class imbalance or artifacts.

The CatBoost feature importance (\Cref{app:fig:class-fi}) reveals that \texttt{mean\_ngender\_rn}{} alone accounts for roughly 60\% of the model's predictive power, with the remaining features contributing complementary but smaller shares. The SHAP summary plot (\Cref{app:fig:class-shap}) offers a more granular, class-resolved view: for each class, the direction and magnitude of SHAP values align consistently with the expected semantics of the features, indicating that the model relies on interpretable, linguistically grounded cues rather than opaque interactions.

\subsubsection{Comparison with off-the-shelf tools}
\label{sec:eval-character}

For the gender label classification task, we compare our hanzi-based method (HZ) against \texttt{Namsor Genderize Name Geo}~(\texttt{N}) and two large language models (LLMs), \texttt{Claude Opus 4.8}~and \texttt{GPT 5.4 Nano}~, both proprietary, prompted with extended thinking enabled to return the perceived gender of each character name in our held-out test set (see \Cref{app:sec:prompt} for prompting details). All tools support at minimum the labels \textit{male}, \textit{female}, and \textit{unknown}\ (or an equivalent that we map to \textit{unknown}\ when not explicitly provided). \texttt{N}~does not predict a \textit{neutral}\ category, whereas the two LLMs do, as we explicitly instructed them to distinguish between \textit{neutral}\ (names commonly used for either gender~\cite{you2024beyond}) and \textit{unknown}\ (names with no clear gender association). We additionally include LLMs as baselines because they represent the current state of the art in general-purpose language understanding and are increasingly used for annotation tasks~\cite{you2024beyond}. This comparison thus also highlights the trade-off between large, resource-intensive models and our lightweight, lexicon-based approach.

\begin{figure}[htbp]
    \centering
    \begin{subfigure}[t]{0.3\textwidth}
        \centering
        \includegraphics[width=\textwidth]{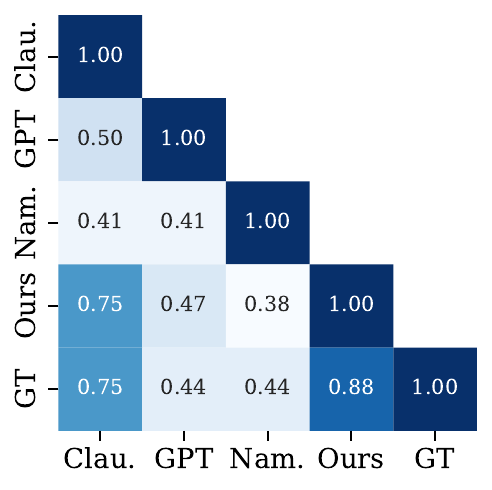}
        \caption{Accuracy}
        \label{app:fig:hz_class_comp_raw}
    \end{subfigure}
    \begin{subfigure}[t]{0.3\textwidth}
        \centering
        \includegraphics[width=\textwidth]{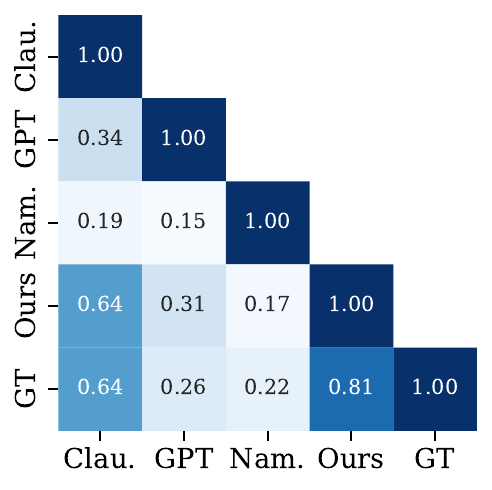}
        \caption{Cohen $\kappa$}
        \label{app:fig:hz_class_comp_kappa}
    \end{subfigure}
    \caption{
    \textbf{Pairwise agreement between hanzi-based gender classification methods.} 
    a) Accuracy and (b) Cohen's $\kappa$ for all method pairs, including ground truth (GT). Our method achieves the highest agreement with GT (accuracy $= 0.88$, $\kappa = 0.81$), followed by \texttt{Claude}~($0.75$, $0.64$), while \texttt{GPT}~($0.44$, $0.26$) and \texttt{Namsor}~($0.44$, $0.22$) show the lowest agreement across all comparisons.
    }
    \label{app:fig:hz_class_comparison}
\end{figure}

We report pairwise agreement (accuracy) and Cohen's $\kappa$ with respect to ground-truth (GT) annotations (\Cref{app:fig:hz_class_comparison}). While raw accuracy captures overall label overlap, $\kappa$ corrects for agreement expected by chance, making it more informative under class imbalance. Our method achieves the highest agreement with GT ($\kappa = 0.81$, accuracy $= 0.88$), followed by \texttt{Claude}\ ($\kappa = 0.64$, accuracy $= 0.75$). \texttt{GPT}\ and \texttt{N}\ both show low agreement ($\kappa = 0.26$, accuracy $= 0.44$ and $\kappa = 0.22$, accuracy $= 0.44$, respectively). Our HZ method and \texttt{Claude}\ exhibit moderate mutual agreement ($\kappa = 0.64$, accuracy $= 0.75$), suggesting both approaches capture some shared underlying patterns despite their fundamentally different architectures---one lexicon-based, the other neural.

\begin{figure}[htbp]
    \centering
    \begin{subfigure}[t]{1.\textwidth}
        \centering
        \includegraphics[width=\textwidth]{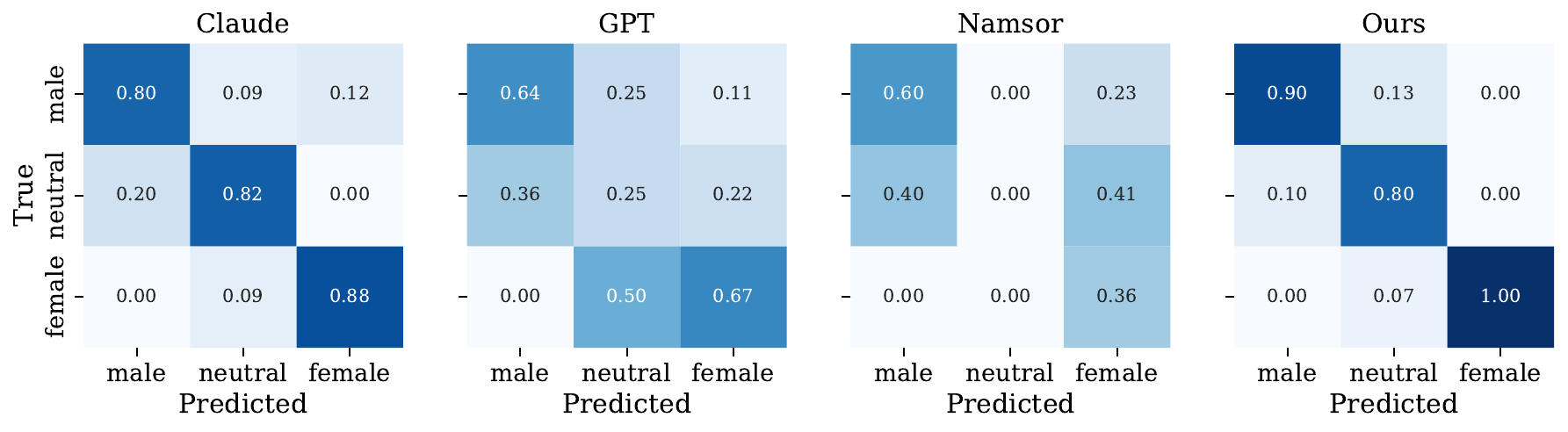}
        \caption{Precision}
        \label{app:fig:hz_pr:precision}
    \end{subfigure}
    \begin{subfigure}[t]{1.\textwidth}
        \centering
        \includegraphics[width=\textwidth]{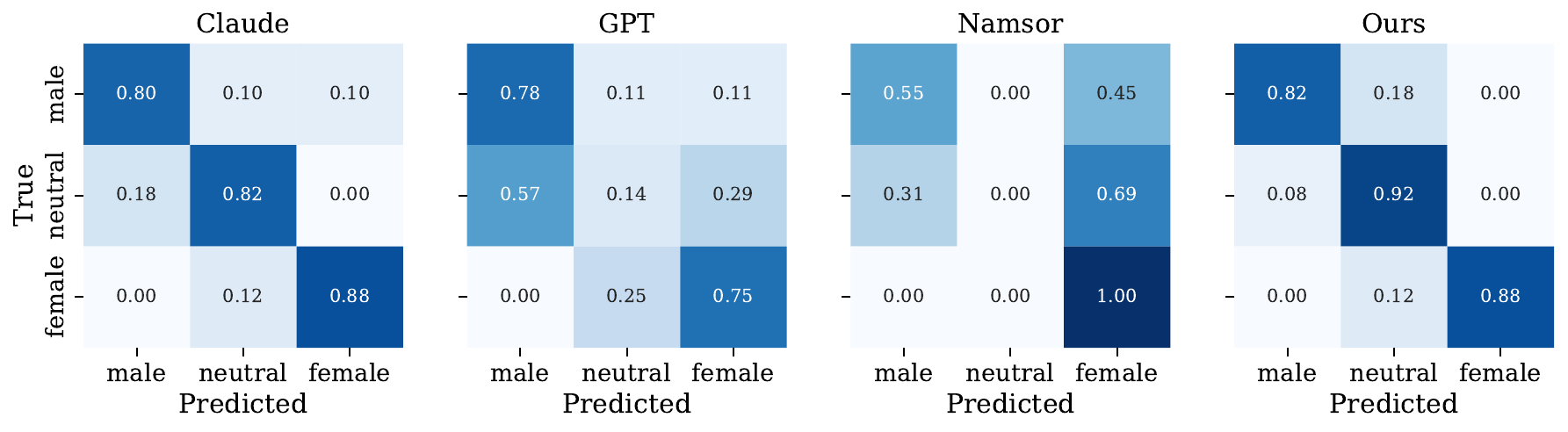}
        \caption{Recall}
        \label{app:fig:hz_pr:recall}
    \end{subfigure}
    \caption{\textbf{Per-class precision and recall of hanzi-based gender classification ($N{=}32$).} 
    Confusion matrices comparing our method (Ours) against two LLMs (\texttt{Claude}, \texttt{GPT}) and \texttt{Namsor}\ on the held-out test set. 
    (a) Precision: our approach leads for \textit{male}\ ($0.90$) and \textit{female}\ ($1.00$), and ranks second for \textit{neutral}\ ($0.80$, behind \texttt{Claude}'s $0.82$).
    (b) Recall: our method achieves the highest \textit{male}\ recall ($0.82$) and \textit{neutral}\ recall ($0.92$), 
    and second highest for \textit{female} ($0.88$), matching \texttt{Claude}; \texttt{Namsor}\ reaches $1.00$ only by over-predicting female (precision $0.36$).
    }
    \label{app:fig:hz_pr}
\end{figure}

To disentangle these scores, we examine per-class precision (how rarely a method mislabels names \emph{as} a class) and recall (how rarely it misses names \emph{of} that class) in \Cref{app:fig:hz_pr}. For precision, our approach leads on \textit{male}\ (0.90), outperforming \texttt{Claude}\ (0.80), \texttt{GPT}\ (0.64), and \texttt{N}\ (0.60). \texttt{Claude}\ leads on \textit{neutral}\ (0.82 vs.\ 0.80 for HZ), while our method achieves perfect \textit{female}\ precision (1.00), followed by \texttt{Claude}\ (0.88) and \texttt{GPT}\ (0.67). For recall, our method ranks highest on both \textit{male}\ (0.82) and \textit{neutral}\ (0.92), outperforming all competitors including \texttt{Claude}\ (0.80 and 0.82, respectively). \texttt{GPT}\ shows particularly weak \textit{neutral}\ recall (0.14), indicating a strong tendency to over-assign gendered labels. For \textit{female}, \texttt{N}\ achieves perfect recall (1.00), though this is offset by poor precision across other classes, while \texttt{Claude}\ and our method both achieve 0.88. Overall, \texttt{N}\ exhibits the weakest performance in both precision and recall, largely due to its inability to predict \textit{neutral}\ labels, which leads to systematic misclassification of gender-neutral names.

\subsection{Pinyin-based inference (\texorpdfstring{$\hat{l}_p$}{l̂ₚ})}
\label{app:sec:class:piyin-inference}

\subsubsection{Aggregation}
\label{app:sec:class:py:aggregation}

We aggregate hanzi-based labels to pinyin-based labels using a supermajority rule. %
Let $\hat{l}_{h} \in \{\textit{male}, \textit{neutral}, \textit{female}\}$ denote the predicted gender label for hanzi character~$h$.
For each pinyin~$p$, we compute the proportion of each gender label across all hanzi that map to~$p$:
\begin{equation}
  r_{l}(p)
  = \frac{\displaystyle\sum_{h \in \mathcal{H}_{\text{cnki+}}(p)}
          \mathbb{1}[\hat{l}_{h}=l]}
         {\big|\mathcal{H}_{\text{cnki+}}(p)\big|}
  \,,\quad l \in \{\text{m},\text{n},\text{f}\}
\end{equation}
A pinyin is assigned \emph{male} if $r_{\text{m}}(p)>\tau$,
\emph{female} if $r_{\text{f}}(p)>\tau$, and \emph{neutral} otherwise, where $\tau=\tfrac{2}{3}$ is a supermajority threshold. We denote the resulting pinyin-based label as $\hat{l}_{p} \in \{\textit{male}, \textit{neutral}, \textit{female}\}$.

\subsubsection{Ambiguity}
\label{app:sec:class:py_ambiguity}

As with the continuous gender score (\Cref{sec:pinyin-model}), a single pinyin can correspond to multiple hanzi with different gender labels. We quantify this variation using two measures.

\para{Entropy.}
We measure hanzi-based label heterogeneity using the normalized Shannon entropy,
\begin{equation}
  E(p)
  = -\frac{1}{\log 3}
    \sum_{l \in \{\text{m},\text{n},\text{f}\}}
    r_{l}(p)\,\log\, r_{l}(p)\,,
\end{equation}
where $E(p) \in [0,1]$ and larger values indicate greater heterogeneity.

\para{Disagreement.}
We measure the inconsistency between the pinyin-based label $\hat{l}_p$
and the underlying hanzi-based labels using the disagreement rate,
\begin{equation}
  D(p)
  = \frac{1}{\big|\mathcal{H}_{\text{cnki}}(p)\big|}
    \sum_{h \in \mathcal{H}_{\text{cnki}}(p)}
    \mathbb{1}\!\left[\hat{l}_{h} \neq \hat{l}_{p}\right],
\end{equation}
where $D(p) \in [0,1]$ and larger values indicate greater disagreement.

\begin{figure}[t]
\centering
\begin{subfigure}[t]{0.24\textwidth}
    \centering
    \includegraphics[width=\linewidth]{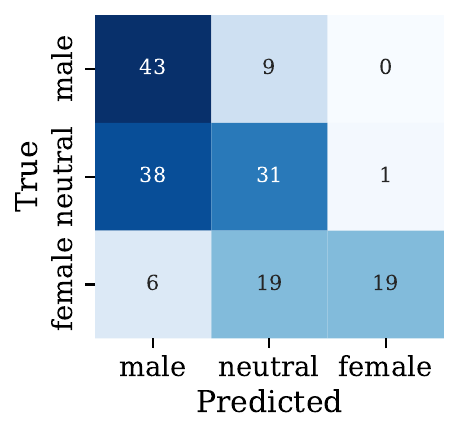}
    \caption{All Data}
    \label{app:fig:py_class_performance:cmall}
\end{subfigure}
\begin{subfigure}[t]{0.24\textwidth}
    \centering
    \includegraphics[width=\linewidth]{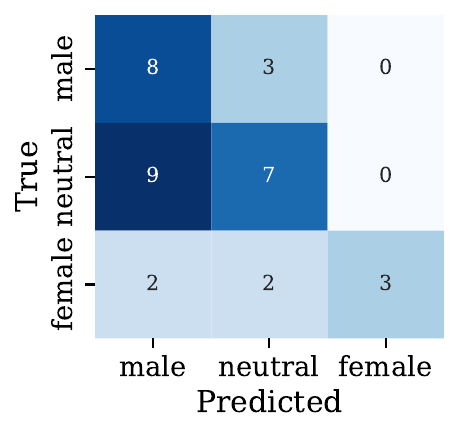}
    \caption{Test Set}
    \label{app:fig:py_class_performance:cmtest}
\end{subfigure}
\begin{subfigure}[t]{0.24\textwidth}
    \centering
    \includegraphics[width=\linewidth]{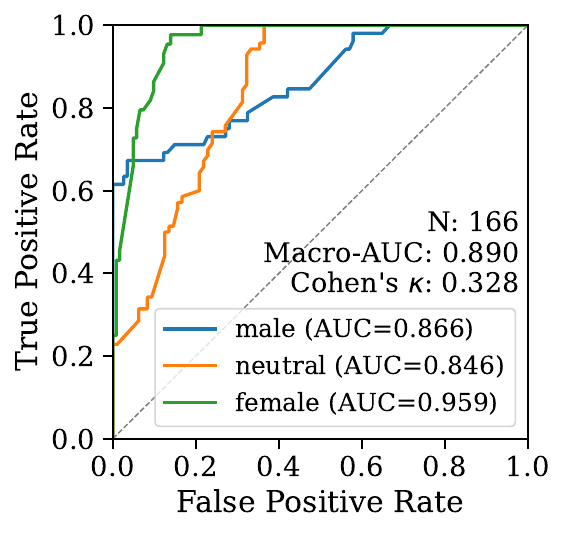}
    \caption{ROC-AUC (all data)}
    \label{app:fig:py_class_performance:rocaucall}
\end{subfigure}
\begin{subfigure}[t]{0.24\textwidth}
    \centering
    \includegraphics[width=\linewidth]{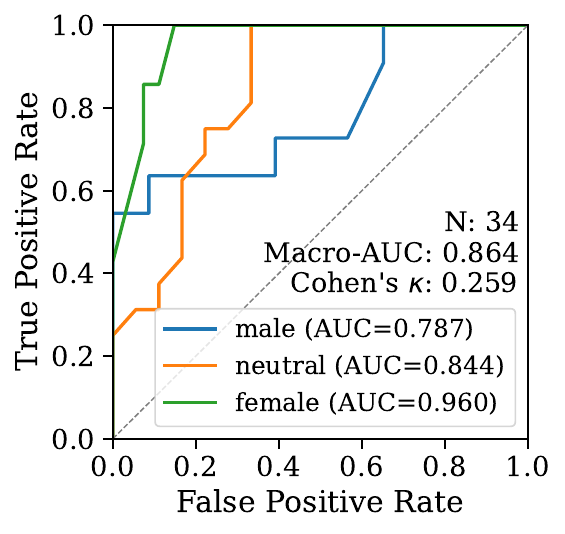}
    \caption{ROC-AUC (test set)}
    \label{app:fig:py_class_performance:rocauctest}
\end{subfigure}
\caption{
\textbf{Performance of pinyin-based gender classification ($\hat{l}_p$).}
(a,b) Confusion matrices on the full ground-truth set (train + test) and the test set, evaluated against the ground-truth labels of each pinyin's associated hanzi.
(c,d) One-vs-rest ROC curves on the same splits, using the hanzi-label proportions $r_l(p)$ as class probabilities.
Our supermajority classifier achieves high macro-AUC ($0.890$/$0.864$) but low Cohen's $\kappa$ ($0.328$/$0.259$), reflecting poor performance on the minority class. %
}
\label{app:fig:py_class_performance}
\end{figure}

\begin{figure}[tbp!]
    \centering
    \includegraphics[width=1.0\linewidth]{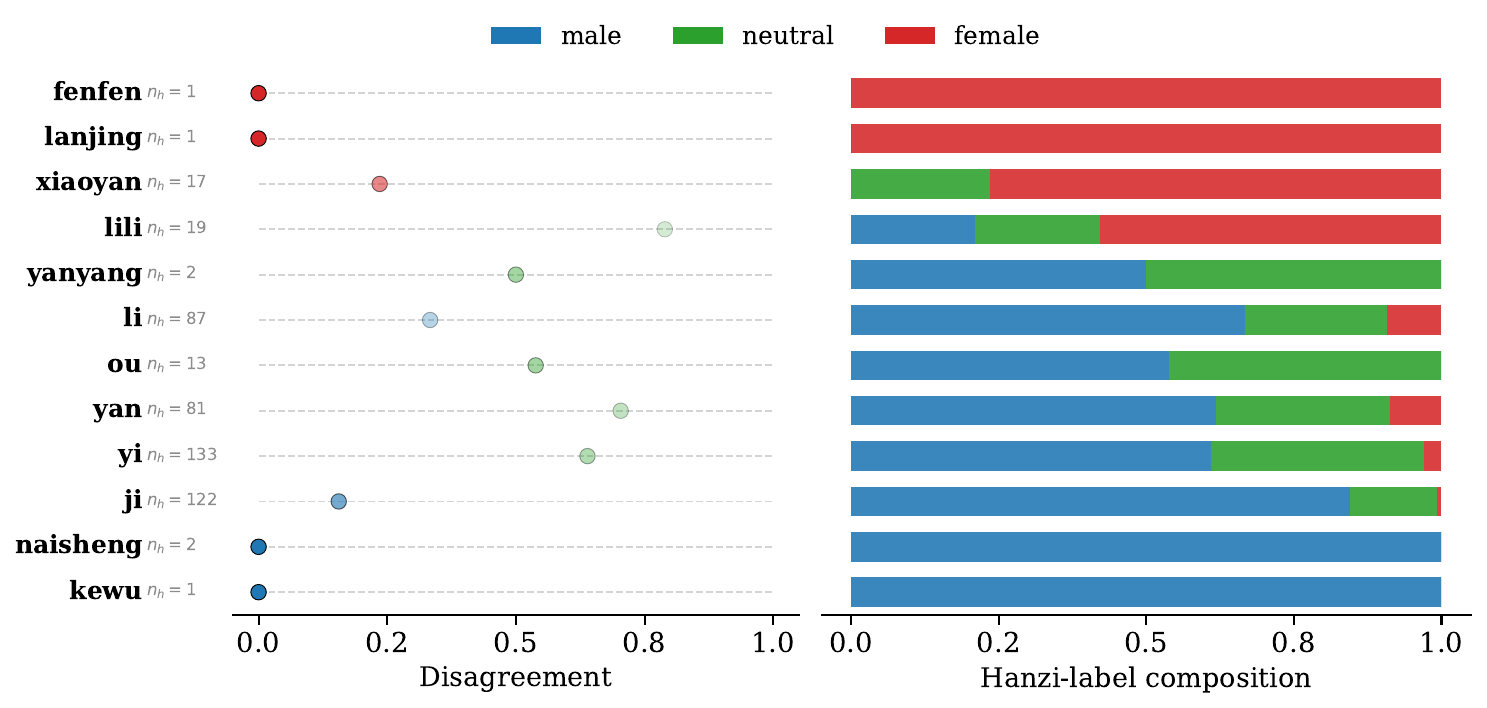}
    \caption{\textbf{Disagreement between pinyin- and hanzi-based gender labels.}
    Left: each dot represents a pinyin name, colored by the inferred majority label $\hat{l}_p$ and with transparency inversely proportional to the entropy of its underlying hanzi-label distribution---low entropy (opaque) indicates strong consensus among hanzi mappings, while high entropy (transparent) indicates mixed labels. The $x$-axis measures the fraction of hanzi labels that disagree with the final pinyin label. 
    Right: hanzi-label composition for each name, showing the proportion of \textit{male}, \textit{neutral}, and \textit{female}\ labels across all hanzi mappings. 
    Names shown are the same as in \Cref{fig:sample_py_ambiguity} for direct comparison.}
    \label{app:fig:py_disagreement}
\end{figure}

\subsubsection{Performance}
\label{app:sec:class:py_performance}

We evaluate the pinyin-based gender label $\hat{l}_p$ by comparing the supermajority prediction for each pinyin against the ground-truth labels of its associated hanzi names. \Cref{app:fig:py_class_performance} summarizes the results via confusion matrices and one-vs-rest ROC curves on both the full ground-truth set (train + test; \Cref{app:fig:py_class_performance:cmall,app:fig:py_class_performance:rocaucall}) and the held-out test set (\Cref{app:fig:py_class_performance:cmtest,app:fig:py_class_performance:rocauctest}).
On the full set ($N{=}166$), the classifier achieves a macro-AUC of $0.890$, with per-class AUCs of $0.866$ (\textit{male}), $0.846$ (\textit{neutral}), and $0.959$ (\textit{female}). Performance on the held-out test set ($N{=}34$) remains comparable, with a macro-AUC of $0.864$ and per-class AUCs of $0.787$, $0.844$, and $0.960$ for \textit{male}, \textit{neutral}, and \textit{female}, respectively. The high AUC values indicate that the proportion-based class probabilities $r_l(p)$ rank pinyin names effectively across all three gender categories.
However, Cohen's $\kappa$ is very low---$0.328$ on both the full set and $0.259$ on the test set---indicating only fair agreement beyond chance when the supermajority rule is applied as a hard classifier. This discrepancy reflects the inherent ambiguity of pinyin-based inference: a single pinyin may map to hanzi spanning multiple gender categories, making the discrete supermajority label a lossy summary of the underlying distribution---a limitation absent in hanzi-based classification (\Cref{app:fig:class}).

\begin{table}[ht!]
\centering
\caption{\textbf{Pinyin-based gender label inference on the held-out set.}
For each pinyin given name, we report the number of associated hanzi names ($n_h$), whether the name is single-character (SC), the hanzi-based gender label counts, the pinyin-based supermajority gender label ($\hat{y}_p$), normalized entropy, and disagreement. Entropy measures how mixed the hanzi-based labels are within a pinyin name. Disagreement measures the share of hanzi-based labels that differ from $\hat{y}_p$.}
\label{app:tbl:pinyin_class_inference_testset}
\begin{tabular}{lrrrrrlrr}
\toprule
 & & & \multicolumn{3}{c}{Hanzi-based labels $\hat{l}_h$} & & \multicolumn{2}{c}{Ambiguity} \\
\cmidrule(lr){4-6} \cmidrule(lr){8-9}
 Pinyin & $n_h$ & SC & Male & Neutral & Female & $\hat{l}_p$ & Entropy & Disagreement \\
\midrule
shen & 35 & \checkmark & 31 & 4 & 0 & male & 0.32 & 0.11 \\
jun & 22 & \checkmark & 18 & 4 & 0 & male & 0.43 & 0.18 \\
qi & 84 & \checkmark & 69 & 14 & 1 & male & 0.47 & 0.18 \\
jie & 52 & \checkmark & 41 & 8 & 3 & male & 0.58 & 0.21 \\
zhen & 42 & \checkmark & 29 & 12 & 1 & male & 0.64 & 0.31 \\
meng & 25 & \checkmark & 17 & 7 & 1 & male & 0.68 & 0.32 \\
jin & 43 & \checkmark & 26 & 16 & 1 & neutral & 0.69 & 0.63 \\
xuan & 40 & \checkmark & 27 & 10 & 3 & male & 0.73 & 0.33 \\
fang & 19 & \checkmark & 11 & 6 & 2 & neutral & 0.84 & 0.68 \\
\midrule
bingchen & 2 & \ding{53} & 2 & 0 & 0 & male & 0.00 & 0.00 \\
bohua & 2 & \ding{53} & 2 & 0 & 0 & male & 0.00 & 0.00 \\
haian & 1 & \ding{53} & 1 & 0 & 0 & male & 0.00 & 0.00 \\
jianju & 1 & \ding{53} & 1 & 0 & 0 & male & 0.00 & 0.00 \\
nianman & 1 & \ding{53} & 1 & 0 & 0 & male & 0.00 & 0.00 \\
qingdong & 3 & \ding{53} & 3 & 0 & 0 & male & 0.00 & 0.00 \\
rongwen & 2 & \ding{53} & 0 & 2 & 0 & neutral & 0.00 & 0.00 \\
taimeng & 1 & \ding{53} & 0 & 1 & 0 & neutral & 0.00 & 0.00 \\
xuetian & 1 & \ding{53} & 0 & 1 & 0 & neutral & 0.00 & 0.00 \\
yingzhen & 1 & \ding{53} & 0 & 0 & 1 & female & 0.00 & 0.00 \\
zehui & 5 & \ding{53} & 4 & 0 & 1 & male & 0.46 & 0.20 \\
wenhan & 4 & \ding{53} & 3 & 1 & 0 & male & 0.51 & 0.25 \\
dongxia & 3 & \ding{53} & 0 & 1 & 2 & female & 0.58 & 0.33 \\
wangmin & 3 & \ding{53} & 2 & 0 & 1 & male & 0.58 & 0.33 \\
zhangqi & 3 & \ding{53} & 2 & 1 & 0 & male & 0.58 & 0.33 \\
yiwei & 24 & \ding{53} & 19 & 3 & 2 & male & 0.59 & 0.21 \\
chunfang & 2 & \ding{53} & 1 & 0 & 1 & neutral & 0.63 & 1.00 \\
feinan & 2 & \ding{53} & 1 & 1 & 0 & neutral & 0.63 & 0.50 \\
huazhen & 2 & \ding{53} & 1 & 0 & 1 & neutral & 0.63 & 1.00 \\
sisi & 2 & \ding{53} & 0 & 1 & 1 & neutral & 0.63 & 0.50 \\
zuying & 2 & \ding{53} & 1 & 0 & 1 & neutral & 0.63 & 1.00 \\
dongyan & 6 & \ding{53} & 1 & 1 & 4 & female & 0.79 & 0.33 \\
yannan & 10 & \ding{53} & 1 & 6 & 3 & neutral & 0.82 & 0.40 \\
yuewen & 5 & \ding{53} & 2 & 2 & 1 & neutral & 0.96 & 0.60 \\
\bottomrule
\end{tabular}
\end{table}

To quantify this ambiguity, we report the normalized Shannon entropy $E(p)$ and disagreement rate $D(p)$ for each pinyin in the held-out set (\Cref{app:tbl:pinyin_class_inference_testset}). Consider, for instance, \textit{qi} ($n_h{=}84$): although its supermajority label is \textit{male} (69 male, 14 neutral, 1 female), it still exhibits moderate entropy ($E{=}0.47$) and non-trivial disagreement ($D{=}0.18$), reflecting the presence of neutral hanzi mappings. In contrast, \textit{fang} ($n_H{=}19$; 11 male, 6 neutral, 2 female) receives a \textit{neutral} supermajority label with high entropy ($E{=}0.84$) and disagreement ($D{=}0.68$), indicating that no single gender label dominates its hanzi set. At the other extreme, pinyin names with few associated hanzi---such as \textit{bingchen} or \textit{haian} ($n_H \leq 2$)---yield zero entropy and disagreement, but this apparent certainty may reflect the coverage of $\mathcal{H}_{\text{cnki+}}$ rather than genuine consensus, as additional hanzi mappings beyond those attested in the corpus could introduce further label variation.

\Cref{app:fig:py_disagreement} illustrates this tension for a representative set of names (the same as in \Cref{fig:sample_py_ambiguity}). For instance, \textit{yan} and \textit{yi} both have \textit{male}\ as their most frequent hanzi label (around $62\%$), yet this falls short of the $\tau{=}\frac{2}{3}$ supermajority threshold, and the combined \textit{neutral}\ and \textit{female}\ share suffices to yield a \textit{neutral}\ pinyin label $\hat{l}_p$. Conversely, names such as \textit{fenfen} and \textit{lanjing} exhibit zero disagreement but map to only a single hanzi each, making their apparent consensus trivial. These cases highlight that the supermajority vote, while informative of the dominant gender tendency, is insufficient to capture the degree of variation among the associated hanzi. Reporting entropy and disagreement alongside $\hat{l}_p$ is essential for a complete characterization of pinyin-level gender label ambiguity.

\subsubsection{Comparison with off-the-shelf tools}
\label{app:sec:py_comparison}

\begin{figure}[t]
    \centering
    \begin{subfigure}[t]{1.\textwidth}
        \centering
        \includegraphics[width=\textwidth]{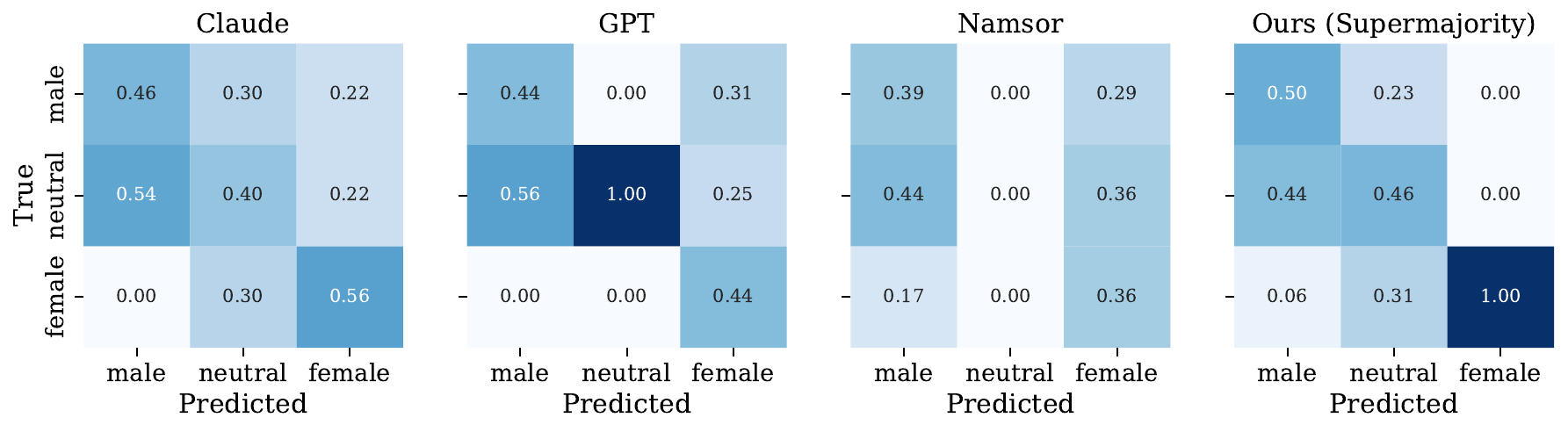}
        \caption{Precision}
        \label{app:fig:py_pr:precision}
    \end{subfigure}
    \begin{subfigure}[t]{1.\textwidth}
        \centering
        \includegraphics[width=\textwidth]{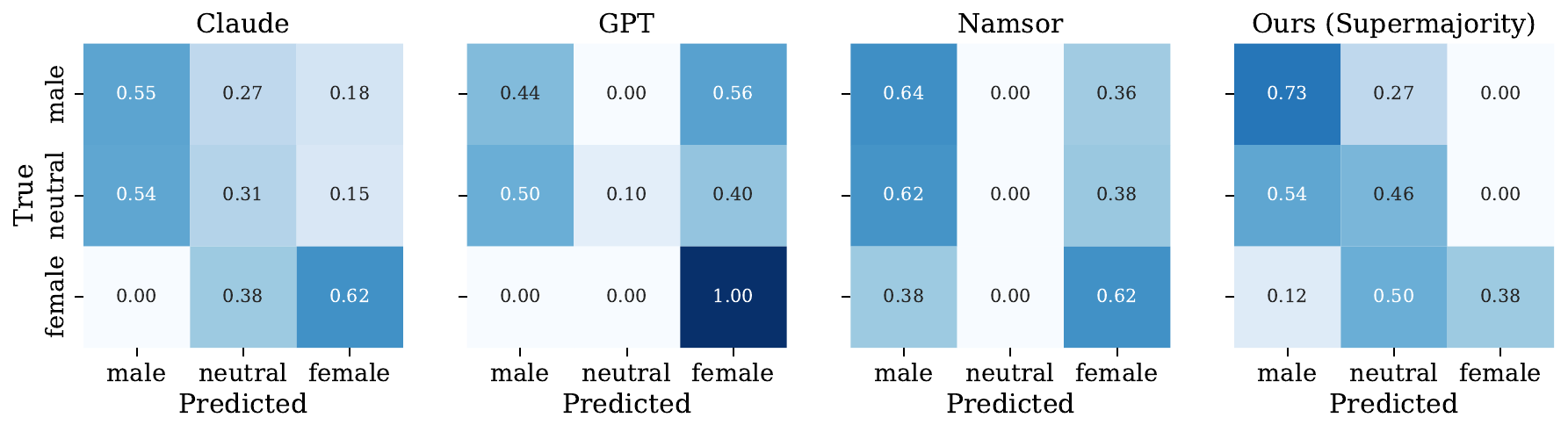}
        \caption{Recall}
        \label{app:fig:py_pr:recall}
    \end{subfigure}
    \caption{\textbf{Per-class precision and recall of pinyin-based gender classification ($N{=}32$).} 
    Confusion matrices comparing our supermajority method against two LLMs (\texttt{Claude}, \texttt{GPT}) and \texttt{Namsor}\ on the held-out test set. 
    (a) Precision: our approach leads for \textit{male}\ ($0.50$) and \textit{female}\ ($1.00$), and ranks second for \textit{neutral}\ ($0.46$, behind \texttt{GPT}'s $1.00$).
    (b) Recall: our method achieves the highest \textit{male}\ recall ($0.73$) and \textit{neutral}\ recall ($0.46$), though all methods show substantially weaker performance than their hanzi-based counterparts (\Cref{app:fig:hz_pr}).
    }
    \label{app:fig:py_pr}
\end{figure}

\begin{figure}[ht!]
    \centering
    \begin{subfigure}[t]{0.3\textwidth}
        \centering
        \includegraphics[width=\textwidth]{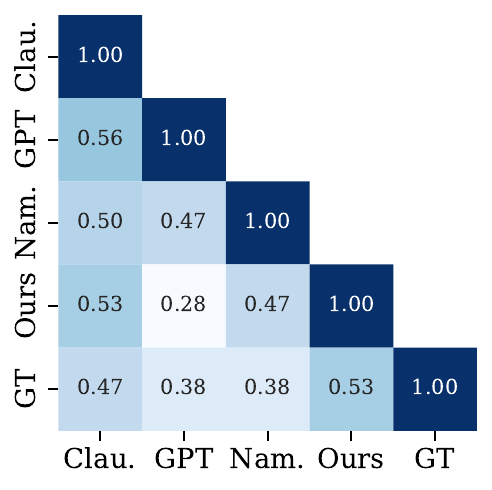}
        \caption{Accuracy}
        \label{app:fig:py_class_comp_raw}
    \end{subfigure}
    \begin{subfigure}[t]{0.3\textwidth}
        \centering
        \includegraphics[width=\textwidth]{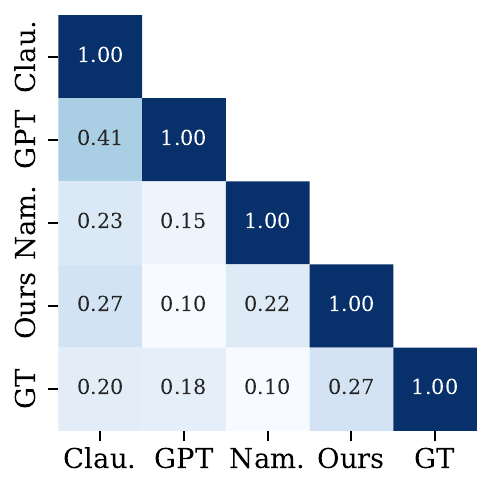}
        \caption{Cohen $\kappa$}
        \label{app:fig:py_class_comp_kappa}
    \end{subfigure}
    \caption{\textbf{Pairwise agreement between pinyin-based gender classification methods ($N{=}32$).} 
    (a) Accuracy and (b) Cohen's $\kappa$ for all method pairs, evaluated against the ground-truth labels of each pinyin's associated hanzi on the held-out test set. 
    Our method achieves the highest agreement with GT (accuracy $= 0.53$, $\kappa = 0.27$), followed by \texttt{Claude}~($0.47$, $0.20$), with \texttt{GPT}~($0.38$, $0.18$) and \texttt{Namsor}~lowest ($0.38$, $0.10$). All values are lower than their hanzi-based counterparts (\Cref{app:fig:hz_class_comparison}), reflecting the information loss inherent in pinyin-based inference.
    }
    \label{app:fig:py_class_comparison}
\end{figure}

We evaluate our pinyin-based gender inference (i.e., supermajority aggregation, see \Cref{app:sec:class:py:aggregation}) and compare its performance against three commercial tools: \texttt{Claude Opus 4.8}, \texttt{GPT 5.4 Nano}, and \texttt{Namsor Genderize Name Geo}. The evaluation consists of inferring gender from the romanized form alone for each pinyin, then comparing these predictions against the ground-truth label of the pinyin's corresponding hanzi, thereby quantifying how much information each method recovers despite the lossy romanization. The per-class confusion matrices (\Cref{app:fig:py_pr}) reveal distinct trade-offs: our approach achieves the highest precision for \textit{female}~(1.00) and the highest recall for \textit{male}~(0.73), yielding the most balanced classification profile. \texttt{Claude}\ shows moderate diagonal dominance but suffers from considerable confusion between \textit{neutral}~and adjacent classes. \texttt{GPT}\ achieves perfect \textit{neutral}~precision (1.00) but with near-zero recall (0.10), indicating it rarely predicts the label at all. \texttt{Namsor}\ collapses the task into a binary decision by never predicting \textit{neutral}, which inflates its \textit{female}~recall (0.62) at the cost of misclassifying all \textit{neutral}~names. Turning to pairwise agreement (\Cref{app:fig:py_class_comparison}), our method attains the highest accuracy (0.53) and Cohen's $\kappa$ $(0.27)$ with ground truth. The uniformly low $\kappa$ across all methods reflects the fact that accuracy is largely driven by correct classification of the majority class, a baseline that $\kappa$ discounts by construction. All values are substantially lower than their hanzi-based counterparts (\Cref{app:fig:hz_class_comparison}), confirming the inherent information loss in romanized input and underscoring the difficulty of the pinyin setting even for our ensemble approach.

\subsection{Implications for Scientometrics and Science-of-Science research}
\label{app:sec:implications}

To test the generalizability of our methods and compare our inference with existing approaches, we use publication data from the American Physical Society (APS)~\cite{APS2026Datasets}, which contains metadata for physics publications including author names.
A subset of Chinese authors in APS have given names recorded in both hanzi and pinyin, providing a natural evaluation setting: the hanzi form yields a more reliable gender prediction, serving as a reference for assessing the corresponding pinyin-based inference.
We compare against three open-source baselines---Karimi et al.~\cite{Karimi2016}, van Buskirk et al.~\cite{van2023open}, and Saeta~\cite{saeta2016genderguesser}---whose predictions we obtain from \cite{bachmann2026cumulative}; we additionally apply Saeta to pinyin names. 
Each baseline infers gender from pinyin alone and produces either a binary gender label or no prediction when gender is unknown (hereafter \emph{unknowns}).
We also evaluate three commercial tools (\texttt{Namsor Genderize Name Geo}, \texttt{Claude Opus 4.8}, and \texttt{GPT 5.4 Nano}), which, like our method, infer gender independently for each script; the prompt used for the LLMs is provided in \Cref{app:sec:prompt}.
To ensure a fair comparison, we restrict evaluation to the $54\%${} of Chinese-named APS authors whose pinyin given names are fully spelled out, excluding the remaining 46\% whose names appear only as initials (e.g., ``J.'').
Because our method grounds inference in the underlying hanzi, it can produce predictions for initialed names as well; the baselines cannot. 
The results below therefore represent a conservative estimate of the coverage differences between methods.

\begin{figure}[ht!]
    \centering
    \begin{subfigure}[t]{1.0\textwidth}
        \centering
        \includegraphics[width=\textwidth]{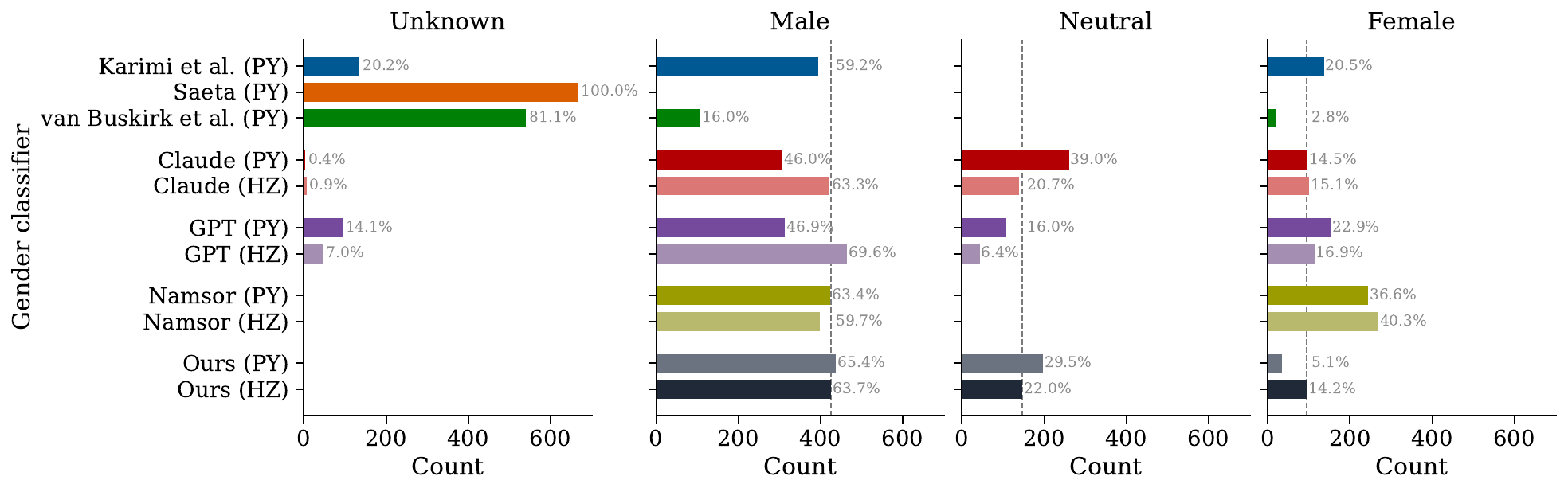}
        \caption{Gender distribution}
        \label{fig:py_baselines:dist}
    \end{subfigure}
    ~
    \begin{subfigure}[t]{0.48\textwidth}
        \centering
        \includegraphics[width=\textwidth]{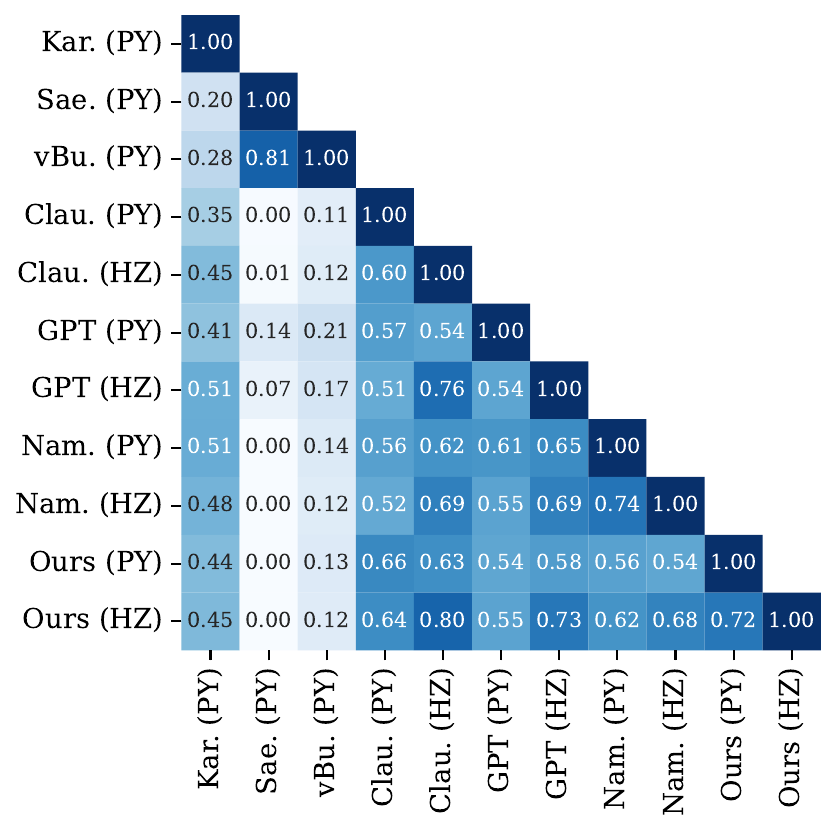}
        \caption{Agreement}
        \label{fig:py_baselines:agreement}
    \end{subfigure}
    \hfill
    \begin{subfigure}[t]{0.48\textwidth}
        \centering
        \includegraphics[width=\textwidth]{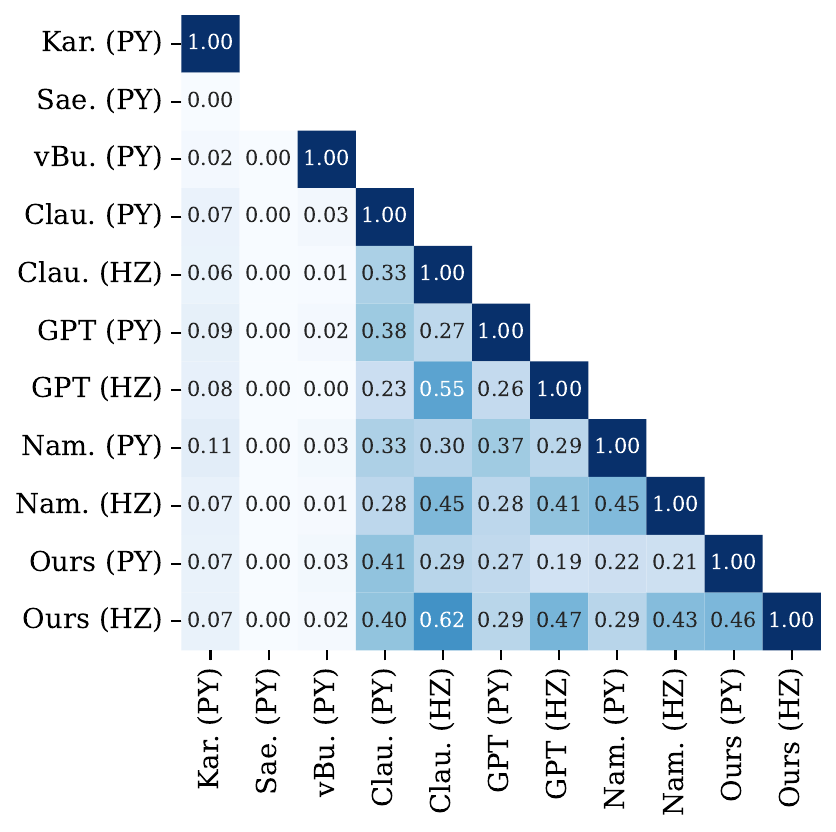}
        \caption{Cohen's $\kappa$}
        \label{fig:py_baselines:kappa}
    \end{subfigure}
    \caption{\textbf{Gender label distribution and agreement across eleven inference methods ($\hat{l}_p$, $N=666$).}
    Comparison of gender label inference for $666$ unique Chinese author given names in APS across three baseline methods (Karimi et al.~\cite{Karimi2016}, Saeta~\cite{saeta2016genderguesser}, and van Buskirk et al.~\cite{van2023open}), three commercial tools (\texttt{Claude}, \texttt{GPT}, and \texttt{Namsor}), and our pinyin supermajority (PY) and hanzi-based (HZ) classification models.
    The baselines infer gender from available pinyin, while the commercial tools and our models each independently classify from both scripts.
    (a)~Distribution of predicted gender labels by method.
    (b)~Pairwise classification agreement.
    (c)~Pairwise Cohen's $\kappa$, measuring agreement beyond chance.
    \texttt{Claude}~(HZ) achieves the highest agreement with our HZ model in both raw accuracy~(b) and chance-corrected $\kappa$~(c), followed by \texttt{GPT}~(HZ) and our PY method.
    }
    \label{fig:py_baselines}
\end{figure}

\Cref{fig:py_baselines:dist} shows the distribution of predicted gender labels across all eleven inference methods. 
Given the strong performance of our hanzi-based classifier, which outperforms the three commercial tools even after controlling for chance agreement and class imbalance (\Cref{app:fig:hz_class_comparison}), we use Ours~(HZ) as the reference distribution throughout this comparison. 
The most immediate pattern is the share of ``\textbf{unknown}'' labels among the open-source baselines: Saeta assigns no label to 100\% of names, van Buskirk et al.\ to $\approx$ 81\%, and Karimi et al.\ to 20\%, rendering these methods largely unusable for Chinese given names written in pinyin. 
\texttt{GPT}\ also leaves a small share of names unlabeled (14\% for PY, 7\% for HZ), while our models, \texttt{Claude}, and \texttt{Namsor}\ always produce a definite label.
For the \textbf{male} category, Ours~(HZ) predicts $\approx$ 64\% of names as male.
Among the remaining methods, \texttt{Namsor}~(PY) and \texttt{Claude}~(HZ) are closest, followed by Ours~(PY) and Karimi et al.
A consistent pattern emerges for our method and \texttt{Namsor}: pinyin-based inference tends to assign a slightly higher proportion of male labels than its hanzi-based counterpart; for \texttt{Claude}\ and \texttt{GPT}, however, the pattern reverses, with hanzi-based inference predicting more male names.
For the \textbf{neutral} category, Ours~(HZ) predicts $\approx$ 22\% of names as neutral.
\texttt{Claude}~(HZ) provides the closest estimate, followed by our pinyin-based supermajority approach.
\texttt{Namsor} does not produce neutral labels at all, as it outputs only binary predictions.
The pinyin-based predictions of \texttt{Claude}\ and \texttt{GPT}\ overestimate neutral names relative to their own hanzi-based predictions.
For the \textbf{female} category, Ours~(HZ) predicts $\approx$ 14\% of names as female.
\texttt{Claude}\ approximates this share in both scripts (14.5\% for PY, 15.1\% for HZ), followed by \texttt{GPT}~(HZ) (16.9\%).
The remaining commercial tools and baselines overestimate this group, while our pinyin-based approach underestimates it (5.1\%).

To quantify how robust these distributions are relative to our hanzi-based approach, we compute pairwise agreement and Cohen's $\kappa$ (\Cref{fig:py_baselines:agreement,fig:py_baselines:kappa}).
The method with the highest raw agreement with Ours~(HZ) is \texttt{Claude}~(HZ) (80\%), followed by \texttt{GPT}~(HZ) (73\%), our pinyin model (72\%), and \texttt{Namsor}~(HZ).
After controlling for chance agreement and class imbalance, overall $\kappa$ values are modest across all methods, but \texttt{Claude}~(HZ) achieves the highest score, followed by \texttt{GPT}~(HZ), our pinyin approach, and \texttt{Namsor}~(HZ).

\section{LLM prompt design}
\label{app:sec:prompt}

Designing a reliable prompt for large-scale gender classification required iterative refinement driven by both qualitative evaluation and practical deployment constraints. This section documents the development process and the rationale behind the final prompt structure.

\para{Initial prototyping.}
We began with two script-specific prompt templates---one for given names in Hanzi, another for given names in Pinyin---each describing the classification task and providing few-shot examples of names deemed male, female, or neutral. 
We tested these templates on the $166$~ground-truth names described in~\Cref{sec:gold-standards} using three models: \texttt{Claude Opus~4.6} and~\texttt{4.7} (via Claude Code\footnote{\url{https://code.claude.com/docs/en/vs-code}} chat in Visual Studio Code) and \texttt{GPT-5.4} (via GitHub Copilot chat in Visual Studio Code\footnote{\url{https://code.visualstudio.com/docs/chat/copilot-chat}}).
All three models produced plausible outputs on this small sample, confirming the feasibility of LLM-based gender inference for Chinese names.

\para{Scaling and content-filtering constraints.}
We then attempted to apply the same prompts at scale (18{,}739~Pinyin names and 46{,}144~Hanzi names) using \texttt{GPT 5.4 Nano}, OpenAI's model for high-volume classification tasks, and \texttt{Claude Opus 4.8}\footnote{Opus~4.8 was the only Opus version available in Claude Code at the time of our experiments.} %
via Claude Code. While this succeeded for \texttt{GPT}, \texttt{Claude}\ rejected every request for the full pinyin set with an \emph{``Output blocked by content filtering policy''} error. Attempts to circumvent this limitation by splitting input files into smaller batches or delegating the task to agentic workflows proved unsuccessful; the filter appeared to be triggered by the classification task itself rather than by batch size.
We briefly circumvented the issue by submitting requests through the Claude.ai\footnote{\url{https://claude.ai/}} chat interface, but this workaround stopped functioning after a short period. 
We therefore switched to the Anthropic API,\footnote{\url{https://platform.claude.com/docs/en/home}} which provided direct programmatic access without the chat-interface content filter, and submitted names in batches of $100$ per request to stay within output-token limits while minimizing API calls.
This transition also motivated a more systematic prompt engineering effort: we used \texttt{Claude}~to draft an initial structured prompt, then iteratively revised the wording, output schema, and scoring definitions through multiple rounds of manual evaluation.
To ensure consistency across all methods, we re-ran both \texttt{GPT}\ and \texttt{Claude}~(HZ) under the same revised prompt and batching setup.
It is worth noting that the intermediate results from the initial single-pass configuration for both \texttt{GPT}\ and \texttt{Claude}\ were worse than those reported in the current version of this manuscript, suggesting that submitting long name lists in a single request introduces noise into the inference.

\begin{figure}[htbp]
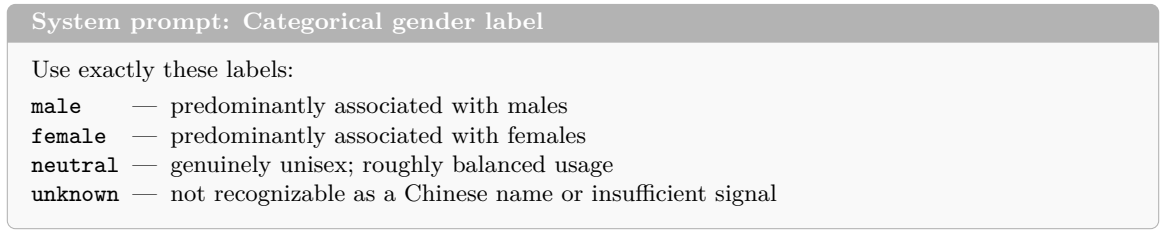

\centering
\small
\begin{tcolorbox}[
    colback=gray!5,
    colframe=gray!60,
    title={\textbf{System prompt: Categorical gender label}},
    fonttitle=\small,
    boxrule=0.4pt,
    left=6pt, right=6pt, top=4pt, bottom=4pt
]
Use exactly these labels:

\smallskip
\begin{tabular}{@{}l@{\;\;---\;\;}p{0.75\linewidth}@{}}
  \texttt{male}    & predominantly associated with males \\
  \texttt{female}  & predominantly associated with females \\
  \texttt{neutral} & genuinely unisex; roughly balanced usage \\
  \texttt{unknown} & not recognizable as a Chinese name or
                     insufficient signal \\
\end{tabular}
\end{tcolorbox}
\caption{\textbf{Categorical labeling block.} 
Discrete gender labels appended to the preamble in the \emph{categorical} inference mode.}
\label{fig:prompt-categorical}
\end{figure}

\begin{figure}[htbp]
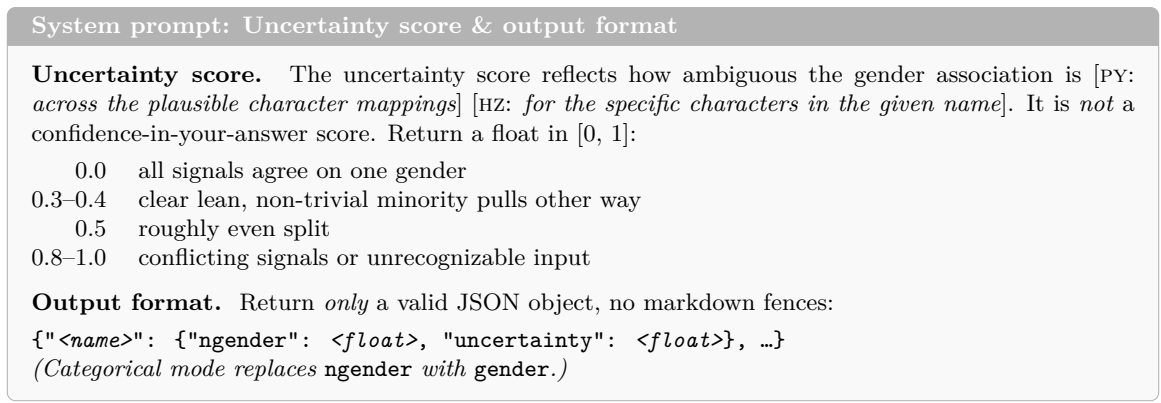

\centering
\small
\begin{tcolorbox}[
    colback=gray!5,
    colframe=gray!60,
    title={\textbf{System prompt: Uncertainty score \& output format}},
    fonttitle=\small,
    boxrule=0.4pt,
    left=6pt, right=6pt, top=4pt, bottom=4pt
]
\textbf{Uncertainty score.}\;
The uncertainty score reflects how ambiguous the gender association is
[\textsc{py}: \textit{across the plausible character mappings}]
[\textsc{hz}: \textit{for the specific characters in the given name}].
It is \emph{not} a confidence-in-your-answer score.
Return a float in $[0,\,1]$:

\smallskip
\begin{tabular}{@{}rl@{}}
  0.0        & all signals agree on one gender \\
  0.3--0.4   & clear lean, non-trivial minority pulls other way \\
  0.5        & roughly even split \\
  0.8--1.0   & conflicting signals or unrecognizable input \\
\end{tabular}

\medskip\noindent
\textbf{Output format.}\;
Return \emph{only} a valid JSON object, no markdown fences:

\smallskip\noindent
\texttt{\{"\textit{<name>}": \{"ngender": \textit{<float>},
"uncertainty": \textit{<float>}\}, \ldots\}}

\noindent
\textit{(Categorical mode replaces} \texttt{ngender} \textit{with}
\texttt{gender}\textit{.)}
\end{tcolorbox}
\caption{\textbf{Uncertainty and output schema.} 
Uncertainty is defined as intrinsic name ambiguity, explicitly distinguished from model confidence. 
The JSON schema ensures lossless round-tripping across batches of up to 100 names. 
This block is shared by both inference modes.}
\label{fig:prompt-uncertainty}
\end{figure}

\para{Final prompt architecture.}
The resulting prompt is assembled from four modular blocks, each addressing a distinct concern, so that inference mode and input script can be varied by swapping a single block while all other instructions remain constant:

\begin{enumerate}[leftmargin=1.5em, itemsep=2pt]
  \item \textbf{Preamble} (Figure~\ref{fig:prompt-preamble}). Establishes
    the domain---Chinese onomastics---and provides script-specific
    instructions. The Pinyin variant directs the model to reason over
    the distribution of plausible character mappings for each syllable,
    whereas the Hanzi variant operates directly on character-level
    gender signals, reflecting the difference in information available
    from each script.

  \item \textbf{Scoring mode} (Figures~\ref{fig:prompt-numeric}
    and~\ref{fig:prompt-categorical}). Defines either a continuous
    score in $[0,1]$ or four discrete labels. Both endpoints of the
    numeric scale are anchored with concrete name examples to reduce
    calibration drift across batches.

  \item \textbf{Uncertainty definition}
    (Figure~\ref{fig:prompt-uncertainty}). Quantifies intrinsic
    ambiguity in the name itself. The prompt explicitly distinguishes
    this from model confidence to prevent the model from conflating
    classification difficulty with name-level gender ambiguity.

  \item \textbf{Output schema}
    (Figure~\ref{fig:prompt-uncertainty}). Specifies a flat JSON
    object keyed by the original input string, ensuring lossless
    round-tripping between input and output across batches of up to
    100 names.
\end{enumerate}

\begin{figure}[htbp]
\centering
\small
\begin{tcolorbox}[
    colback=gray!5,
    colframe=gray!60,
    title={\textbf{System prompt: Preamble}},
    fonttitle=\small,
    boxrule=0.4pt,
    left=6pt, right=6pt, top=4pt, bottom=4pt
]
You are an expert sociolinguist specializing in Chinese onomastics.
Your task is to infer the most likely gender association of Chinese
personal names provided in \textbf{Pinyin (without tone marks)}
\textit{[Hanzi: Chinese characters]}.

\smallskip\noindent\textbf{Domain knowledge you must apply:}

\begin{enumerate}[leftmargin=1.2em, itemsep=2pt, topsep=2pt, parsep=0pt]
  \item \textbf{Name-level gender signal.}
    \begin{itemize}[leftmargin=1em, itemsep=1pt, topsep=1pt]
      \item[\textsc{py}:] A single Pinyin syllable maps to many possible characters.
            Reason across the most common character candidates and weigh
            their aggregate gender association.
      \item[\textsc{hz}:] Each character carries gender connotations shaped by
            meaning, radical composition, and naming-convention frequency.
            Evaluate the gender signal of each character based on its
            established usage in Chinese personal names.
    \end{itemize}
  \item \textbf{Given name vs.\ full name.}
    When a likely surname is present, focus gender inference on the
    remaining given-name portion.
  \item \textbf{Compound given names.}
    Evaluate the combined gender connotation of multi-character/syllable
    given names, not each unit in isolation.
\end{enumerate}
\end{tcolorbox}
\caption{\textbf{Prompt preamble.} Role assignment and script-specific domain knowledge. 
Instructions unique to the Pinyin (\textsc{py}) and Hanzi (\textsc{hz}) variants are marked inline; all other text is shared.}
\label{fig:prompt-preamble}
\end{figure}

\begin{figure}[htbp]
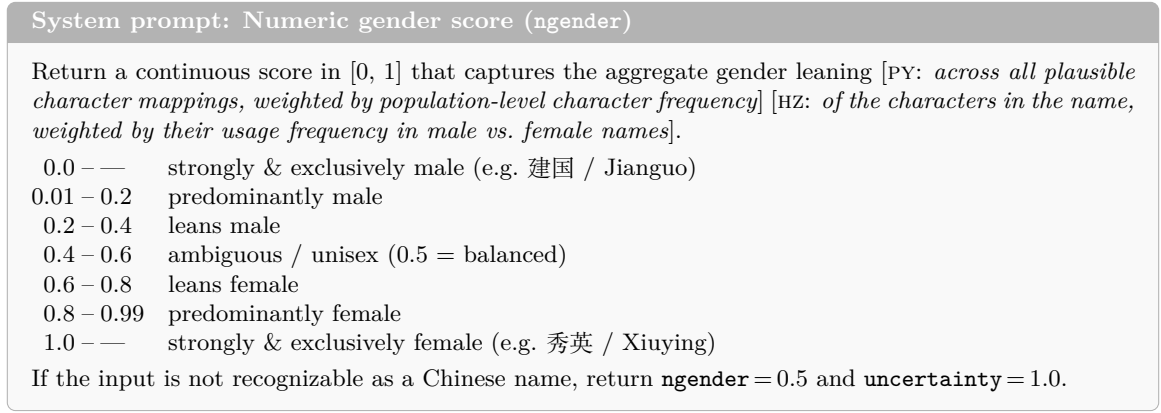

\centering
\small
\begin{tcolorbox}[
    colback=gray!5,
    colframe=gray!60,
    title={\textbf{System prompt: Numeric gender score (\texttt{ngender})}},
    fonttitle=\small,
    boxrule=0.4pt,
    left=6pt, right=6pt, top=4pt, bottom=4pt
]
Return a continuous score in $[0,\,1]$ that captures the aggregate
gender leaning
[\textsc{py}: \textit{across all plausible character mappings, weighted
by population-level character frequency}]
[\textsc{hz}: \textit{of the characters in the name, weighted by their
usage frequency in male vs.\ female names}].

\smallskip
\begin{tabular}{@{}r@{\;--\;}l@{\quad}l@{}}
  0.0  & ---  & strongly \& exclusively male (e.g.\ 建国 / Jianguo) \\
  0.01 & 0.2  & predominantly male \\
  0.2  & 0.4  & leans male \\
  0.4  & 0.6  & ambiguous / unisex (0.5 = balanced) \\
  0.6  & 0.8  & leans female \\
  0.8  & 0.99 & predominantly female \\
  1.0  & ---  & strongly \& exclusively female (e.g.\ 秀英 / Xiuying) \\
\end{tabular}

\smallskip\noindent
If the input is not recognizable as a Chinese name, return
\texttt{ngender}\,=\,0.5 and \texttt{uncertainty}\,=\,1.0.
\end{tcolorbox}
\caption{\textbf{Numeric scoring block.} 
Continuous gender score definition appended to the preamble (Figure~\ref{fig:prompt-preamble}) in the \emph{numeric} inference mode. 
}
\label{fig:prompt-numeric}
\end{figure}

\end{document}